\pdfoutput=1

\documentclass[11pt]{article}

\usepackage[]{emnlp2021}

\usepackage{times}
\usepackage{latexsym}

\usepackage[T1]{fontenc}

\usepackage[utf8]{inputenc}

\usepackage{microtype}      
\usepackage{graphicx}       
\usepackage[T1]{fontenc}    
\usepackage{multirow}       
\usepackage{subcaption}     
\usepackage{amssymb}        
\usepackage{bold-extra}     
\usepackage{bm}             
\usepackage[hang,flushmargin]{footmisc}  
\newcommand{\LN}{\linebreak\noindent}    

\newcommand{\POS}{\texttt{POS}}
\newcommand{\NER}{\texttt{NER}}
\newcommand{\DEP}{\texttt{DEP}}
\newcommand{\CON}{\texttt{CON}}
\newcommand{\SRL}{\texttt{SRL}}
\newcommand{\mask}{\mathbf{z}}

\usepackage{xcolor,colortbl}
\usepackage{amsmath}
\usepackage{nicefrac}
\usepackage{bbm}
\usepackage[ruled,vlined]{algorithm2e}

\newcommand{\dG}[1]{\textcolor[RGB]{0,102,0}{#1}}
\newcommand{\dR}[1]{\textcolor[RGB]{153,0,0}{#1}}
\newcommand{\posdiff}[1]{{\small(\dG{+#1})}}
\newcommand{\negdiff}[1]{{\small(\dR{-#1})}}
\newcommand{\textsec}[1]{\textsection\ref{#1}}
\title{The Stem Cell Hypothesis:\\ Dilemma behind Multi-Task Learning with Transformer Encoders}

\author{Han He \\
  Department of Computer Science \\
  Emory University \\
  Atlanta, GA 30322 USA \\
  \texttt{han.he@emory.edu} \\\And
  Jinho D. Choi \\
  Department of Computer Science \\
  Emory University \\
  Atlanta, GA 30322 USA \\
  \texttt{jinho.choi@emory.edu} \\}

\begin{document}
\maketitle

\begin{abstract}

Multi-task learning with transformer encoders (MTL) has emerged as a powerful technique to improve performance on closely-related tasks for both accuracy and efficiency while a question still remains whether or not it would perform as well on tasks that are distinct in nature.
We first present MTL results on five NLP tasks, \POS, \NER, \DEP, \CON, and \SRL, and depict its deficiency over single-task learning.
We then conduct an extensive pruning analysis to show that a certain set of attention heads get claimed by most tasks during MTL, who interfere with one another to fine-tune those heads for their own objectives.
Based on this finding, we propose the \textit{Stem Cell Hypothesis} to reveal the existence of attention heads naturally talented for many tasks that cannot be jointly trained to create adequate embeddings for all of those tasks.
Finally, we design novel parameter-free probes to justify our hypothesis and demonstrate how attention heads are transformed across the five tasks during MTL through label analysis.



\end{abstract}
\section{Introduction}
\label{sec:introduction}

Transformer encoders (TEs) have established recent state-of-the-art results on many core NLP tasks \cite{he-choi-2019, yu-etal-2020-named, ijcai2020-560}. 
However, their architectures can be viewed ``over-parameterized'' as downstream tasks may not need all those parameters, prone to cause an overhead in computation.
One promising approach to mitigate this overhead is multi-task learning (MTL) where a TE is shared across multiple tasks; thus, it needs to be run only once to generate final embeddings for all tasks \cite{clark-etal-2019-bam}.


Despite the success in MTL on closely-related tasks such as language understanding \cite{wang-etal-2018-glue} or relation extraction \cite{chen-etal-2020-relabel, lin-etal-2020-joint}, MTL on core NLP tasks (e.g., tagging, parsing, labeling) whose decoders are very distinct has not been well-studied.
This work employs the state-of-the-art decoders on five core tasks for MTL and thoroughly analyzes interactions among those tasks to explore a possibility of reducing the computation overhead from TEs.
Surprisingly, our experiments depict that models jointly trained by MTL give lower accuracy than ones trained individually, that is against findings from previous work.
In fact,\LN models jointly trained with all five tasks perform the worst among any other combination (Section~\ref{sec:mtl}).

These experimental results urge us to figure out why MTL on core tasks with a shared TE leads to worse performance than its single-task counterparts. 
Our exploration begins by detecting essential heads for each task by forcing the TE to use as few attention heads as possible while maintaining accuracy similar to a fully-utilized encoder.
Our experiments reveal that all five tasks rely on almost the same set of attention heads. Hence, they compete for those heads during MTL, causing to blur out features extracted by individual tasks.
Thus, we propose the \textit{Stem Cell Hypothesis}, likening these talented attention heads to \textit{stem cells}, which cannot be fine-tuned for multiple tasks that are very distinct (Section~\ref{sec:pruning-anlaysis}).

To validate this hypothesis, many parameter-free probes are designed to observe how every attention head is updated while trained individually or jointly.
Intriguingly, we find that heads not fine-tuned for any task can still give remarkably high performance to predict certain linguistic structures, confirming the existence of \textit{stem cells} inherently more talented;
it is consistent with previous work stating that TEs carry on a good amount of syntactic and semantic knowledge \cite{tenney-etal-2019-bert,liu-etal-2019-linguistic,jawahar-etal-2019-bert,hewitt-manning-2019-structural}.
After single-task learning, probing results typically improve along with the task performance, illustrating that the \textit{stem cells} are developed into more task-specific experts.
On the contrary, MTL often drops\LN both probing and task performance, supporting our hypothesis that attention heads lose expertise when exposed to multiple teaching signals that may conflict to one another (Section~\ref{sec:probing-ana}).

\noindent The \textit{Stem Cell Hypothesis} is proposed to shed light on a possible direction to MTL research using TEs, comprising an unbearable amount of parameters, by wisely assigning attention heads to downstream tasks.
Although most analysis in this study is based on BERT, we also provide extensive experimental results and visualization of other recent TEs including RoBERTa \cite{liu2019roberta}, ELECTRA \cite{clark2020electra} and DeBERTa \cite{he2020deberta} in \textsec{ssec:other-transformers} to further demonstrate the generality of our hypothesis.
To the best of knowledge, this is the first time that a comprehensive analysis of attention heads is made for MTL on those core tasks by introducing novel parameter-free probing methods.\footnote{All our resources including source codes and models are public available at \url{https://github.com/emorynlp/stem-cell-hypothesis}.}

\section{Related Work}
\label{sec:related-work}

A small portion of our work overlaps with multi-task learning. MTL with pre-trained transformers specifically in NLP \cite{wang-etal-2018-glue,clark-etal-2019-bam,liu-etal-2019-multi,kondratyuk-straka-2019-75,chen-etal-2020-relabel,lin-etal-2020-joint} has been widely studied. Most work focus on neural architecture design to encourage beneficial message passing across tasks. Our MTL framework adopts conventional architecture and applies tricks of batch sampling \cite{wang-etal-2019-tell} and loss balancing.

Most of our work falls into the analysis of BERT, especially from a linguistic view. Since BERT was introduced, studies on explaining why BERT works have never stopped. The most related studies are those trying to study the linguistic structures learnt by BERT. Among them, \citet{tenney-etal-2019-bert} and \citet{liu-etal-2019-linguistic} showed part-of-speech, syntactic chunks and roles can be discovered from BERT embeddings. Using a supervised probe, \citet{hewitt-manning-2019-structural} successfully discover full dependency parse trees. The encoded dependency structure is also supported by \citet{jawahar-etal-2019-bert} using probes on embeddings. Apart from these parameterized probes, parameter-free approaches \cite{clark-etal-2019-bert, wu-etal-2020-perturbed} also agree with the existence of rich linguistic knowledge in BERT, which is closely related to our probing methods.

What remains unclear is the impact of fine-turning on TEs. Using supervised probes, \citet{peters-etal-2019-tune} claim that fine-tuning adapts BERT embeddings to downstream tasks, which is later challenged by \citet{hewitt-liang-2019-designing} since supervised probe itself can encode knowledge. Then, \citet{zhao-bethard-2020-berts} propose a methodology to test such encoding of a linguistic phenomenon by comparing the probing performance before and after fine-tuning. Our probing methods align with these unsupervised probes while focus more on explaining the impact of multi-task learning. 
\section{Multi-Task Learning}
\label{sec:mtl}

Our goal of MTL is to build a joint model sharing the same encoder but using a distinct decoder for each task that outperforms its single-task counterparts while being faster and more memory efficient.
Our model adapts hard parameter sharing \cite{caruana:93a} such that all decoders take the same hidden states generated by the shared encoder as input and make task-specific predictions in parallel.

\begin{table*}[htbp!]
\centering\small
\begin{tabular}{c||c|c|c|c|c||c}
    & \textbf{\POS}          & \textbf{\NER}          & \textbf{\DEP}          & \textbf{\CON}          & \textbf{\SRL}          & \bf MTL-5          \\ \hline \hline
\textbf{\POS} & \cellcolor{gray!32}\textbf{98.32 $\pm$ 0.02} & 98.28 $\pm$ 0.01 & 98.28 $\pm$ 0.02 & 98.30 $\pm$ 0.02 & 98.27 $\pm$ 0.02 & 98.25 $\pm$ 0.01 \\ 
\textbf{\NER} & 89.34 $\pm$ 0.24 & \cellcolor{gray!32}89.04 $\pm$ 0.14 & \textbf{89.43 $\pm$ 0.14} & 88.38 $\pm$ 0.06 & 89.18 $\pm$ 0.25 & 88.94 $\pm$ 0.10 \\ 
\textbf{\DEP} & 94.04 $\pm$ 0.02 & 94.06 $\pm$ 0.07 & \cellcolor{gray!32}\textbf{94.24 $\pm$ 0.03} & 94.12 $\pm$ 0.04 & 94.12 $\pm$ 0.03 & 93.84 $\pm$ 0.08 \\ 
\textbf{\CON} & 94.23 $\pm$ 0.03 & 94.38 $\pm$ 0.04 & 94.33 $\pm$ 0.02 & \cellcolor{gray!32}\textbf{94.43 $\pm$ 0.03} & 94.25 $\pm$ 0.05 & 94.10 $\pm$ 0.05 \\ 
\textbf{\SRL} & 82.92 $\pm$ 0.10 & 82.39 $\pm$ 0.07 & 82.05 $\pm$ 0.03 & \textbf{83.17 $\pm$ 0.02} & \cellcolor{gray!32}82.93 $\pm$ 0.08 & 82.30 $\pm$ 0.13 \\ 
\end{tabular}
\caption{Performance of single-task learning (main diagonal highlighted in gray), multi-task learning on all 5 tasks (MTL-5), and multi-task learning on every pair of the tasks (non-diagonal cells; e.g., \DEP'th row in \NER'th column is the \DEP\ result of the joint model between \DEP\ and \NER). See also Table~\ref{tab:mtl-results-more} for similar results of other TEs.
}
\vspace{-1.0ex}
\label{tab:mtl-results}
\end{table*}

\subsection{Shared Encoder}

For main experiments, BERT \cite{devlin-etal-2019-bert} is used as the shared encoder although our approach can be adapted to any transformer encoders (\textsec{ssec:other-transformers}).
Every token gets split into subtokens by BERT; eventually, the average of the last layer's hidden states generated for those subtokens is used as the final embedding of that token.
Additionally, word dropout is applied for generalization by replacing random subtokens with \texttt{[MASK]} during training.

\subsection{Task-Specific Decoders}
\label{ssec:task-specific-decoders}

Five tasks are experimented, part-of-speech tagging (\POS), named entity recognition (\NER), dependency parsing (\DEP), constituency parsing (\CON), and semantic role labeling (\SRL).
For each task, a state-of-the-art decoder is adopted (except for \POS) to provide a modern benchmark for MTL on these tasks, and simplified to build an efficient model.

\vspace{-0.1ex}
\paragraph{\POS} 

A linear layer is used as a \POS\ decoder that takes the final embedding of each token from BERT and generates the output vector where each dimension gives the score of a particular POS tag.

\vspace{-0.2ex}
\paragraph{\NER} 

The biaffine decoder \cite{yu-etal-2020-named} is used for \NER.
For simplification, document context, fastText and character-level embeddings as well as variational BiLSTM encoding from the original approach are removed.

\vspace{-0.2ex}
\paragraph{\DEP} 

The biaffine decoder \cite{dozat:17a} is used for \DEP\ as well. 
For simplification, part-of-speech tags, character-level embeddings and the variational BiLSTM are removed in our approach.
Also, the final embedding of \texttt{[CLS]} from BERT is used to represent the root node.

\vspace{-0.2ex}
\paragraph{\CON} 

The two-stage CRF decoder is used for \CON\ \cite{ijcai2020-560}. 
The unlabeled bracket scorer is optimized using a tree-structure CRF objective on unlabeled constituents. 
The encoding layer from the original approach is substituted by BERT.
Also, \texttt{[CLS]} and \texttt{[SEP]} in BERT are used to represent \texttt{[BOS]} and \texttt{[EOS]}, respectively.

\vspace{-0.2ex}
\paragraph{\SRL} 

The end-to-end span ranking decoder \cite{he-etal-2018-jointly} is used for \SRL.
The attention-based span representations are replaced by the averaged embeddings as suggested by \citet{xia-etal-2019-syntax}. 
For simplification, a linear layer is used as the ranker instead of the biaffine one since they have shown similar performance in our preliminary experiments.

\subsection{Data and Loss Balancing}

During multi-task training, batches from different tasks are shuffled together and randomly sampled to optimize the shared encoder and the corresponding decoder. 
Following \citet{wang-etal-2019-tell}, a task is sampled based on a probability proportional to its dataset size raised to the power of $0.8$.
To balance the losses of all tasks, a running average of every task is monitored and its loss is updated as follow:

\begin{equation*}
\mathcal{L}_t^{\prime} = \frac{\sum_{\forall i}{\bar{\mathcal{L}_i}}}{\bar{\mathcal{L}_t}} \cdot \mathcal{L}_t
\end{equation*}
$\mathcal{L}_t$ is the current loss of the task $t$, $\bar{\mathcal{L}_t}$ is the running average of the most recent 5 losses of $t$, and $\mathcal{L}_t^{\prime}$ is the updated loss of $t$.
This balancing method normalizes the loss of each task to the same magnitude and has shown to prevent MTL from being biased to specific tasks in our preliminary experiments.

\subsection{MTL Experiments}
\label{ssec:mtl-experiments}

Our models are experimented on the OntoNotes 5 \cite{weischedel2013ontonotes} using the data split suggested by \citet{pradhan-etal-2013-towards}.
Table~\ref{tab:mtl-results} illustrates performance of all models using the following evaluation metrics - \POS: accuracy, \NER: span-level labeled F1, \DEP: labeled attachment score, \CON: constituent-level labeled F1, \SRL: micro-averaged F1 of (predicate, argument, label).
Every model is trained 3 times and their average score and standard deviation on the test set are reported.
For \DEP,\LN the gold trees from \CON\ are converted into the Stanford dependencies v3.3.0 \cite{de2008stanford}.
Detailed descriptions about the experimental settings are provided in Appendix~\ref{ssec:experimental-settings}.

\noindent Single-task learning models are first trained then compared to the MTL model trained on all 5 tasks (MTL-5).
Interestingly, MTL-5 is outperformed by its single-task counterparts for all tasks.
Due to the high complexity of MTL-5, it is hard to tell which combinations of tasks introduce negative transfer.
Thus, we conduct MTL on every pair of the tasks to observe if there is any task combination that yields a positive result (non-diagonal cells in Table~\ref{tab:mtl-results}).

Among the 10 pairwise task combinations, none of them derives a win-win situation. 
\NER\ results are generally improved with MTL although results on the other tasks are degraded, implying that \NER\ takes advantage of the other tasks by hurting their performance.
\SRL\ is also benefited from \CON\ although it is not the case for the other way around. Results of other recent TEs reveal similar patterns as shown in Appendix~\ref{ssec:other-transformers}.

\section{Pruning Analysis}
\label{sec:pruning-anlaysis}

To answer why MTL leads to suboptimal results in Section~\ref{sec:mtl}, pruning strategies are applied to BERT such that only attention heads absolutely necessary to get the best performance are kept for every task.
This allows us to see if there exists a common set of heads that multiple tasks want to claim and train for\LN only their objectives, which can cause conflicts for those heads to be shared across all tasks.

\begin{table*}[htbp]
\centering\resizebox{\textwidth}{!}{
\begin{tabular}{c||c|c|c|c||c|c|c||r|r}
    & \multicolumn{4}{c||}{\bf Performance} & \multicolumn{3}{c||}{\bf \% of Attention Heads Kept} & \multicolumn{2}{c}{\bf PS/S}                    \\ \cline{2-10}
    & \bf STL & \bf STL-SP & \bf STL-DP & \bf MTL-DP & \bf STL-SP & \bf STL-DP & \bf MTL-DP & \multicolumn{1}{c|}{\bf STL} & \multicolumn{1}{c}{\bf STL-DP}  \\ \hline\hline
\bf \POS & 98.32$\pm$0.02          & 98.22$\pm$0.03 & \textbf{98.35$\pm$0.02} & 98.28$\pm$0.01 & 53.24$\pm$4.07 & \textbf{40.51$\pm$1.61} & 50.70$\pm$1.20          & 405        & \textbf{1,245} \\ 
\bf \NER & 89.04$\pm$0.14          & 88.87$\pm$0.10 & \textbf{89.05$\pm$0.08} & 88.78$\pm$0.13 & 57.87$\pm$2.63 & \textbf{49.77$\pm$7.13} & 50.70$\pm$1.20          & 661        & \textbf{700}  \\ 
\bf \DEP & \textbf{94.24$\pm$0.03} & 94.08$\pm$0.10 & 94.22$\pm$0.06          & 93.92$\pm$0.06 & 63.66$\pm$5.02 & \textbf{50.00$\pm$2.08} & 50.70$\pm$1.20          & 241        & \textbf{601}  \\ 
\bf \CON & \textbf{94.43$\pm$0.03} & 94.16$\pm$0.08 & 94.24$\pm$0.03          & 94.16$\pm$0.05 & 45.37$\pm$0.40 & \textbf{44.91$\pm$0.40} & 50.70$\pm$1.20          & 191       & \textbf{397}  \\ 
\bf \SRL & 82.93$\pm$0.08          & 83.01$\pm$0.05 & \textbf{83.11$\pm$0.16} & 82.77$\pm$0.10 & 82.41$\pm$5.99 & 53.24$\pm$4.07          & \textbf{50.70$\pm$1.20} & 299       & \textbf{326}   \\ 
\end{tabular}
}
\caption{Results of single-task learning (STL), STL with static pruning (STL-SP), STL with dynamic pruning (STL-DP), and multi-task learning on the 5 tasks with dynamic pruning (MTL-DP). PS/S: processed samples per second for speed comparison. The STL Performance column is equivalent to the main diagonal in Table~\ref{tab:mtl-results}. See also Table~\ref{tab:pruning-results-more} for pruning results of other recent TEs.}
\label{tab:pruning-results}
\end{table*}

\subsection{Pruning based on \bm{$L_0$}-Regularization}
\label{ssec:pruning-regularization}

BERT is essentially a stack of multi-head attention layers and there is a wide consensus that different layers learn distinct knowledge \cite{lin-etal-2019-open,hewitt-manning-2019-structural,jawahar-etal-2019-bert,liu-etal-2019-linguistic,tenney-etal-2019-bert}.
Inspired by this, we analyze if each head learns features unique to process different tasks.
First, $L_0$-regularization is applied \cite{louizos2018learning} to encourage BERT to use as few heads as possible during training.
In particular, a binary variable $z_j$ is assigned to the $j$'th head and multiplied to the output of that head (see \citet{vaswani2017attention} for $\mathbf{Q}, \mathbf{K}, \mathbf{V}, d_k$):
\begin{equation*}
	\text{Attention}^{(j)}(\mathbf{Q}, \mathbf{K}, \mathbf{V}) = z_j \cdot \text{softmax}(\frac{\mathbf{Q}\mathbf{K}^\top}{\sqrt{d_k}})\mathbf{V}
\end{equation*}
Unfortunately, these binary variables $\mask=\{z_j : \forall_{j = [1, \ell]}\}$ ($\ell$: total \# of heads) are discrete and non-differentiable so cannot be directly learnt using gradient based optimization.
To allow for efficient continuous optimization, each $\mask_j$ is then relaxed as a random variable drawn
independently from a continuous random distribution. Specifically, the relaxed $\mask$ is re-parameterized by its inverse of the cumulative density function (CDF) as $G_{\bm{\alpha}}(\mathbf{u})$. It is sampled as follows, where $\bm{\alpha}$ is a learnable parameter of the inverse CDF, $U$ is the uniform distribution over the interval $[0,1]$ and $\mathbf{u} = \{u_j : \forall_{j = [1, \ell]}\}$ denotes the iid samples from it:
\[
\mathbf{u} \sim U(0,1) \quad\Rightarrow\quad \mask = G_{\bm{\alpha}}(\mathbf{u}) \\
\]
Then, the \textit{Hard Concrete Distribution} \cite{louizos2018learning} is chosen for $\mathbf{z}$, which gives the following form of $G_{\bm{\alpha}}(\mathbf{u})$ that is differentiable, where $(l, r)$ defines the interval that $g_{\bm{\alpha}}(\mathbf{u})$ can be stretched into ($l < 0$, $r > 1$):
\begin{align*}
g_{\bm{\alpha}}(\mathbf{u}) &=\text{sigmoid}(\log \mathbf{u} - \log(\mathbf{1}-\mathbf{u}) + \bm{\alpha})  \\
G_{\bm{\alpha}}(\mathbf{u}) &= \min(\mathbf{1}, \max(\mathbf{0}, g_{\bm{\alpha}}(\mathbf{u}) \times (r - l) + l))
\end{align*}
By sampling $\mathbf{u}$ and applying the Monte Carlo approximation, the learnable $L_0$-objective is obtained in a closed form, which gets jointly optimized with a task specific loss or the balanced MTL loss:
\begin{align}
\label{eq:l0-loss}
\begin{split}
\mathbb{E}_{\,\mathbf{u} \sim U(0,1)}\left[\mask\right]&=\text{sigmoid}\left( \bm{\alpha} - \log\frac{-l}{r}\right) \\
	\mathbb{E}_{\,\mathbf{u} \sim U(0,1)}\left[ L_0 \right] &= \sum_{j=1}^n \mathbb{E}\left[ z_j\right]  \\
\end{split}
\end{align}

\subsection{Pruning Strategies}

Two types of pruning strategies, static and dynamic, are applied for the attention head analysis:

\paragraph{Static Pruning} 

We refer to the conventional two-stage train-then-prune as static pruning (SP) since it fine-tunes the encoder first then freezes the decoder for pruning \cite{voita-etal-2019-analyzing}.

\paragraph{Dynamic Pruning} 

Since SP requires twice the efforts to obtain a pruned model, we propose a new method that simultaneously fine-tunes and prunes. 
This strategy is referred to as dynamic pruning (DP) since the decoder dynamically adapts to the encoder that is being pruned during training, as opposed to SP which instead freezes the decoder. DP is found to be more effective in our experiments.


All pruning models are trained for 3 runs with different random seeds and the best checkpoints by scores on development sets are kept.
Once trained, $\mathbb{E}_{\,\mathbf{u} \sim U(0,1)}\left[\mask\right] \in (0, 1)$ is used as a measure of how much each head is being utilized. 

\begin{figure*}[ht!]
 \centering
 \begin{subfigure}[b]{0.4\columnwidth}
     \centering
     \includegraphics[width=\textwidth]{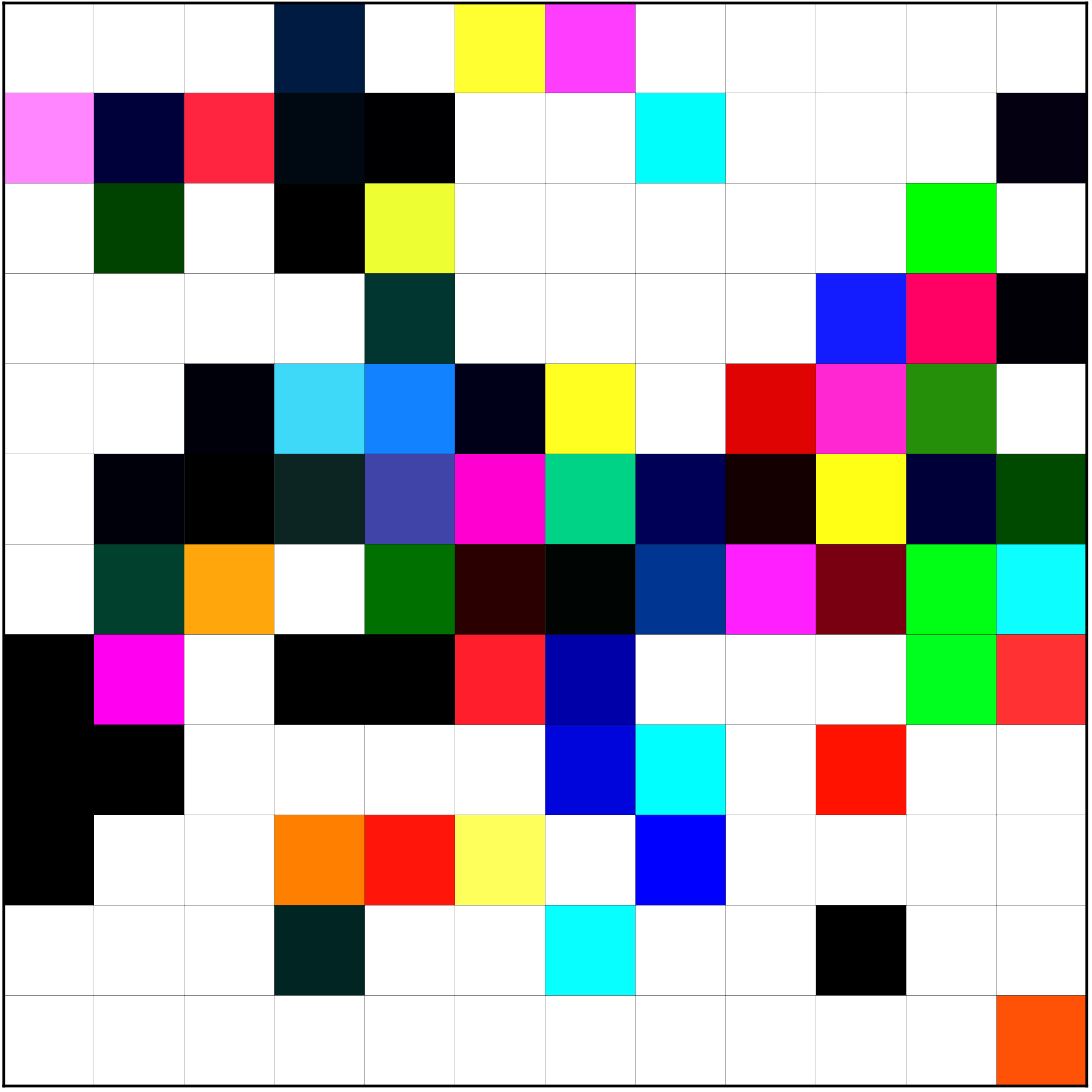}
     \caption{POS}
     \label{fig:probe-pos}
 \end{subfigure}
 \hfill
 \begin{subfigure}[b]{0.4\columnwidth}
     \centering
     \includegraphics[width=\textwidth]{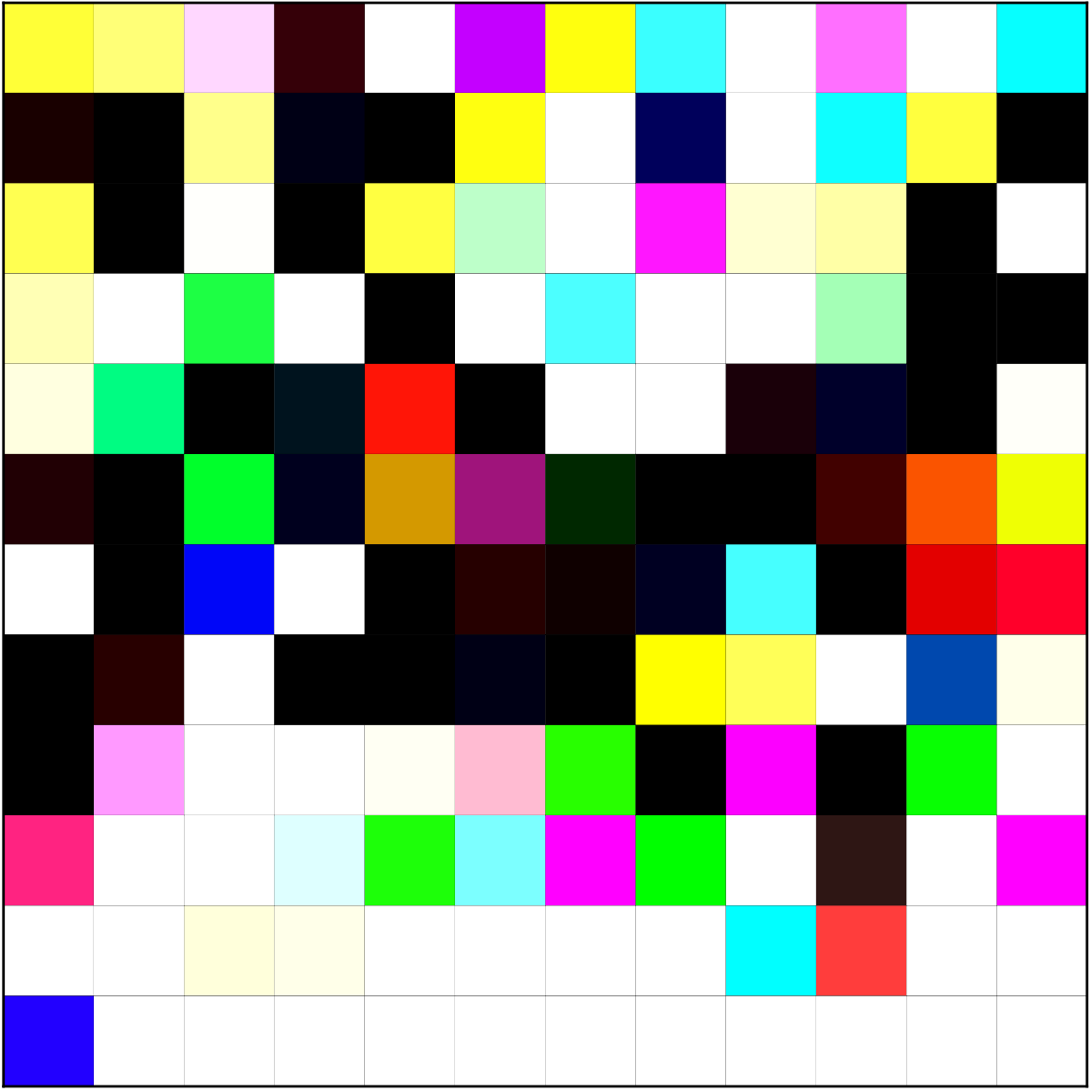}
     \caption{NER}
     \label{fig:probe-ner}
 \end{subfigure}
 \hfill
 \begin{subfigure}[b]{0.4\columnwidth}
     \centering
     \includegraphics[width=\textwidth]{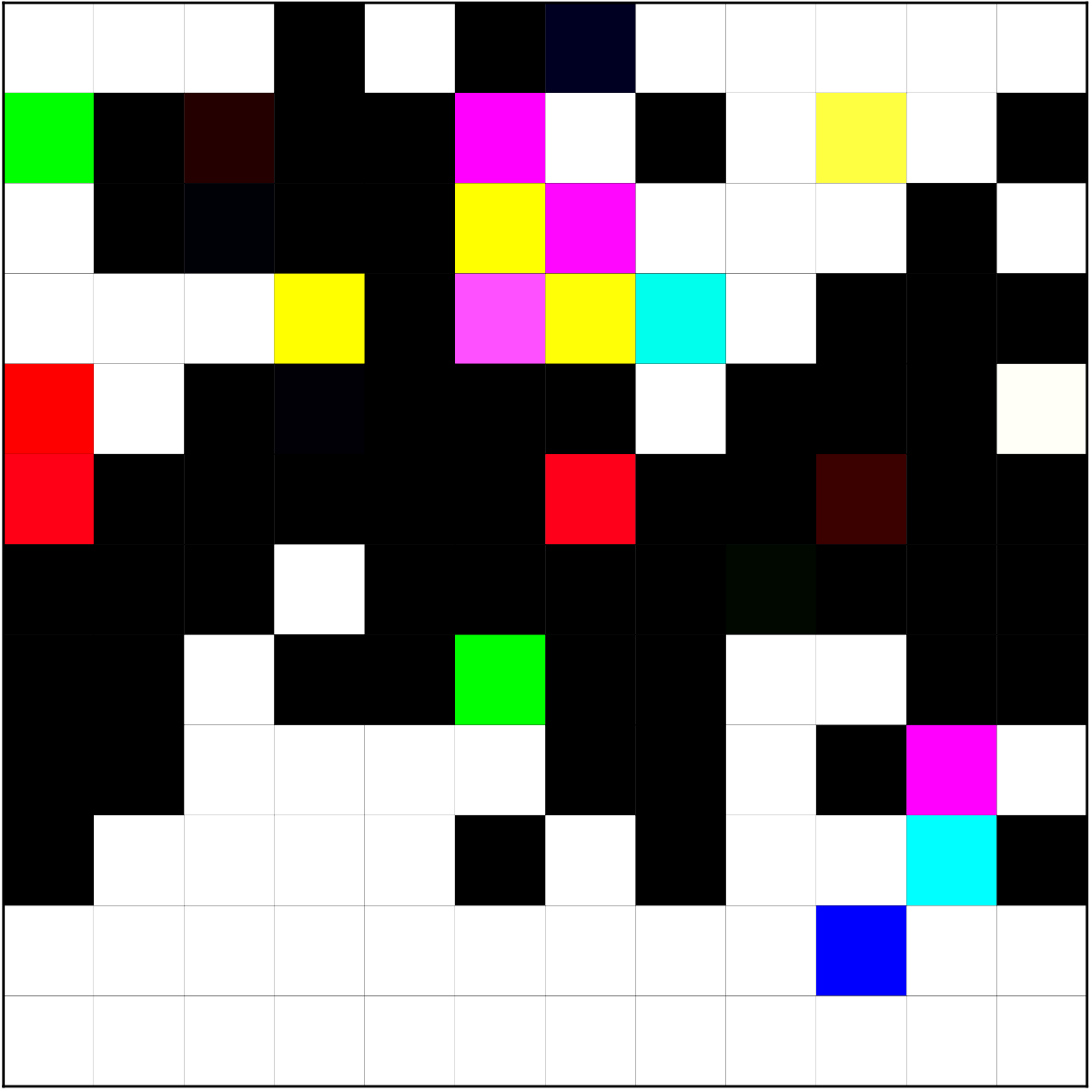}
     \caption{DEP}
     \label{fig:probe-dep}
 \end{subfigure}
 \hfill
 \begin{subfigure}[b]{0.4\columnwidth}
     \centering
     \includegraphics[width=\textwidth]{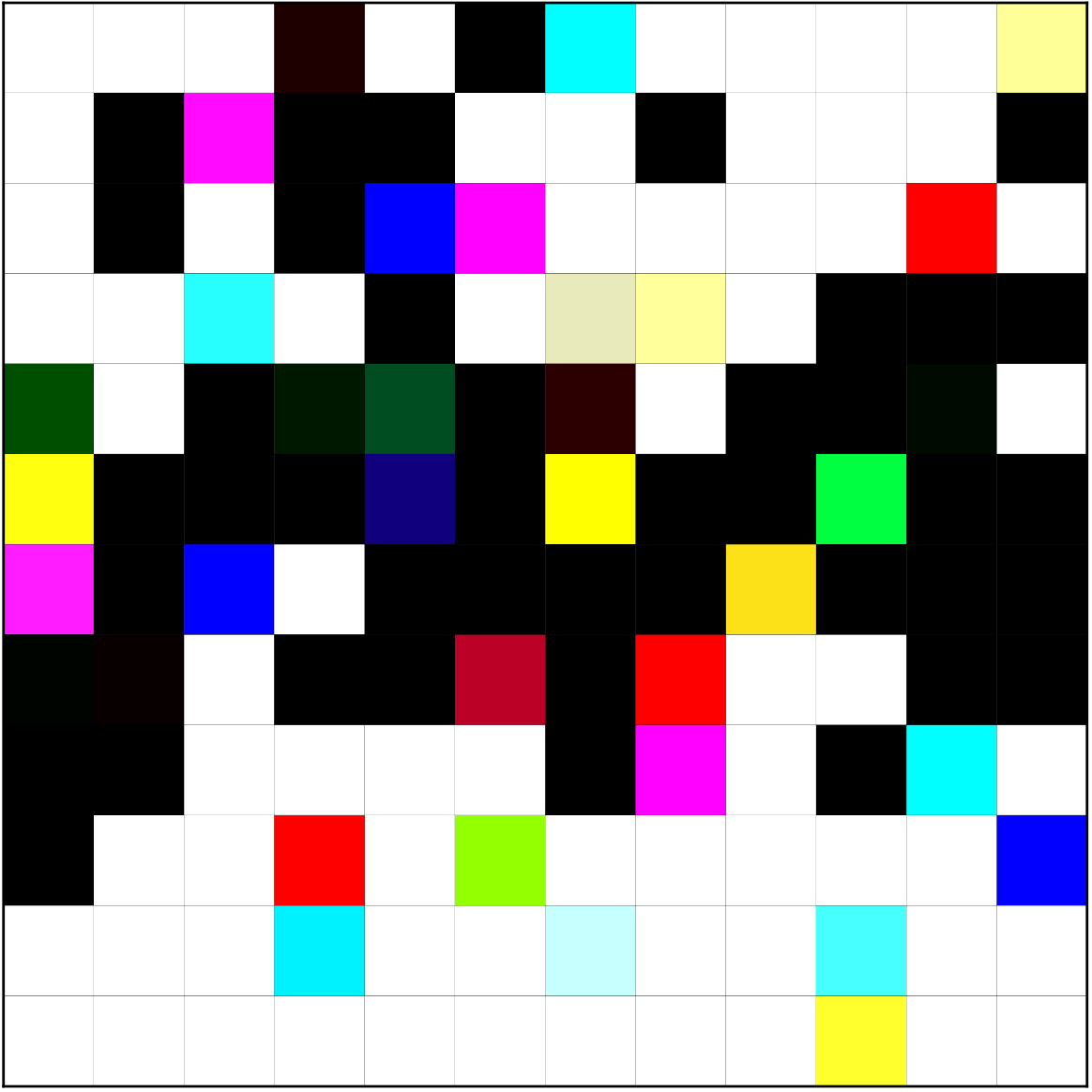}
     \caption{CON}
     \label{fig:probe-con}
 \end{subfigure}
 \hfill
 \begin{subfigure}[b]{0.4\columnwidth}
     \centering
     \includegraphics[width=\textwidth]{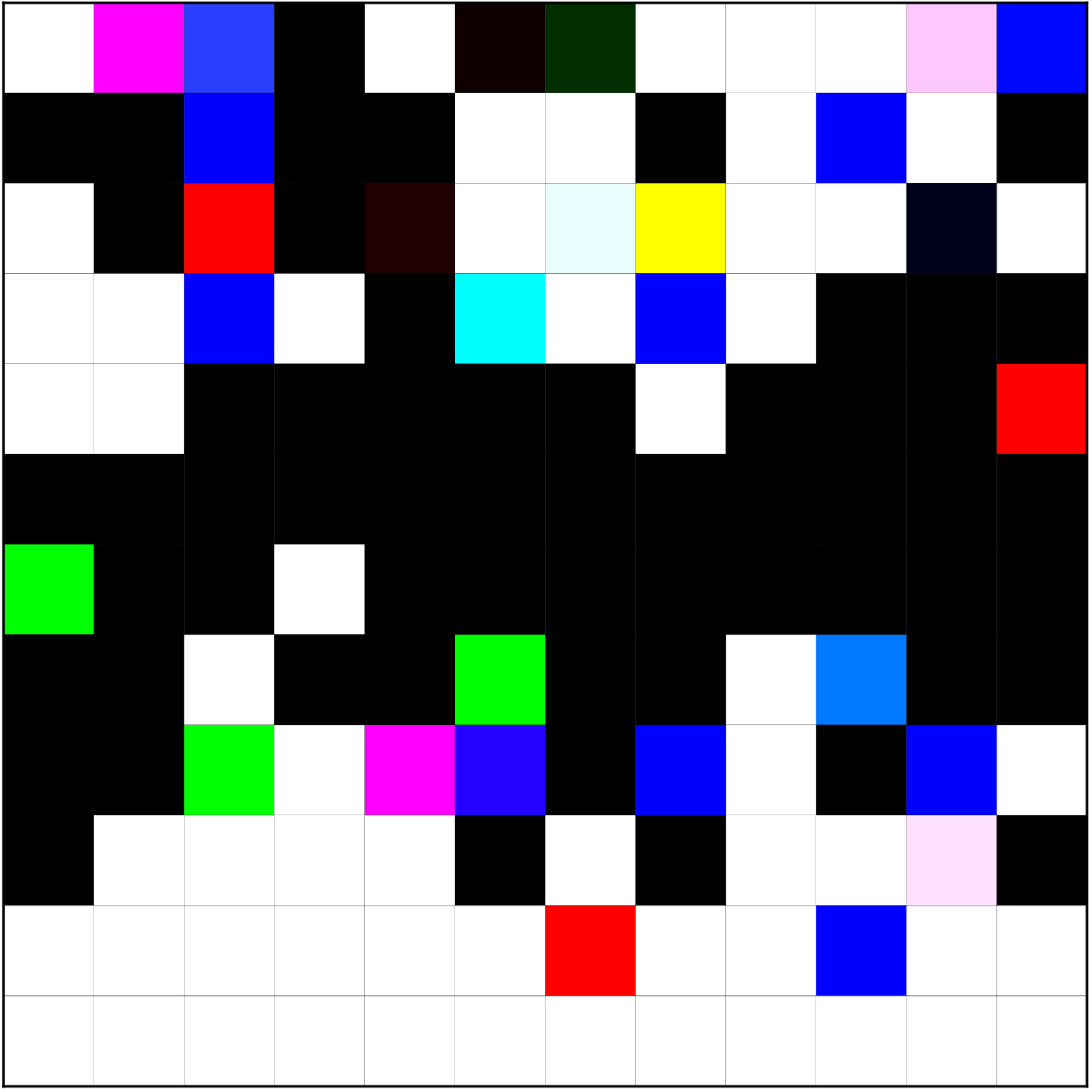}
     \caption{SRL}
     \label{fig:probe-srl}
 \end{subfigure}

 \begin{subfigure}[b]{\columnwidth}
     \centering
     \includegraphics[width=\textwidth]{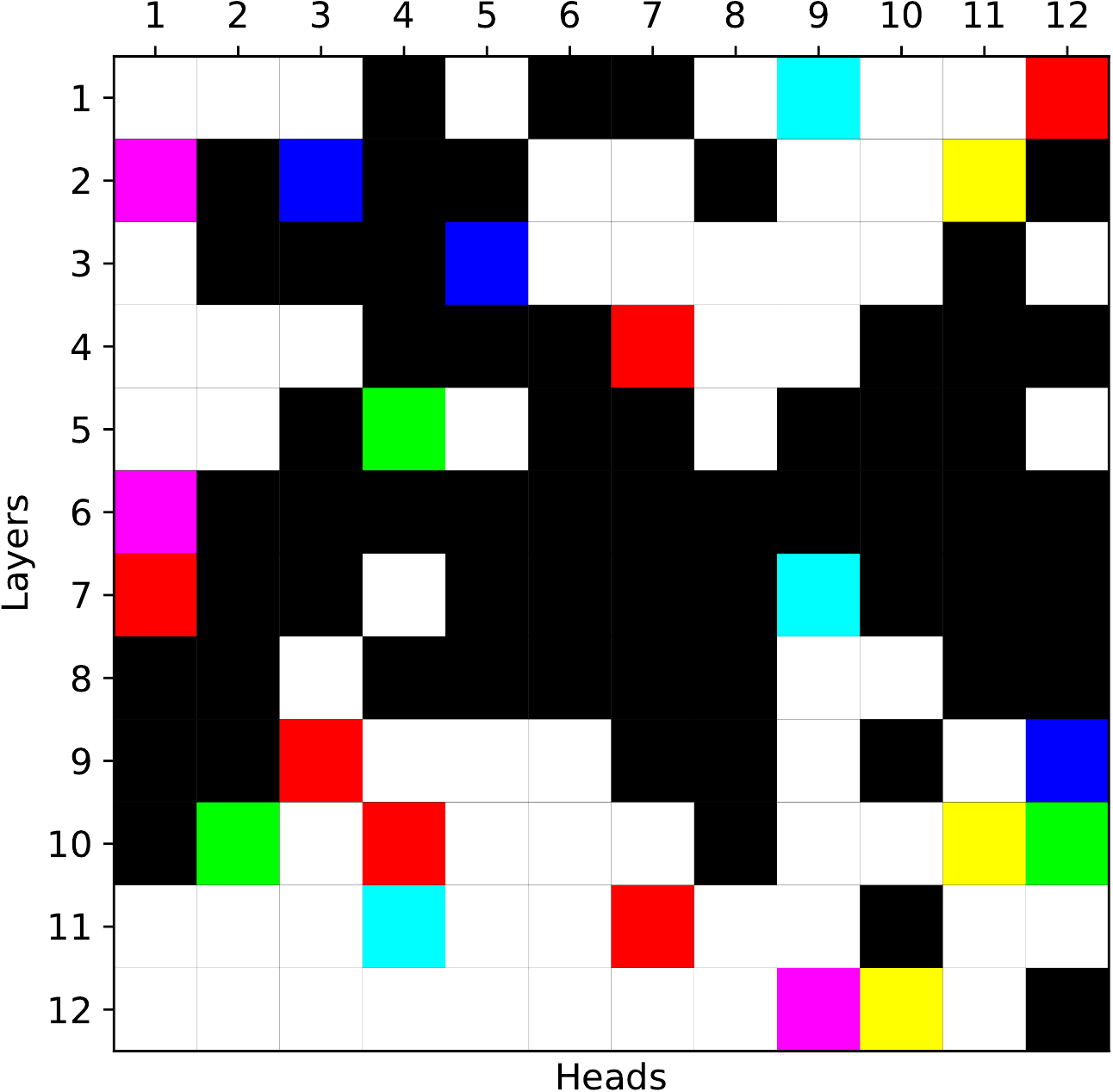}
     \caption{3-run utilization of the MTL-DP model, where each run is encoded in a RGB channel. Darker indicates higher utilization.}
     \label{fig:probe-joint}
 \end{subfigure}
 \hfill
 \begin{subfigure}[b]{\columnwidth}
     \centering
     \includegraphics[width=\textwidth]{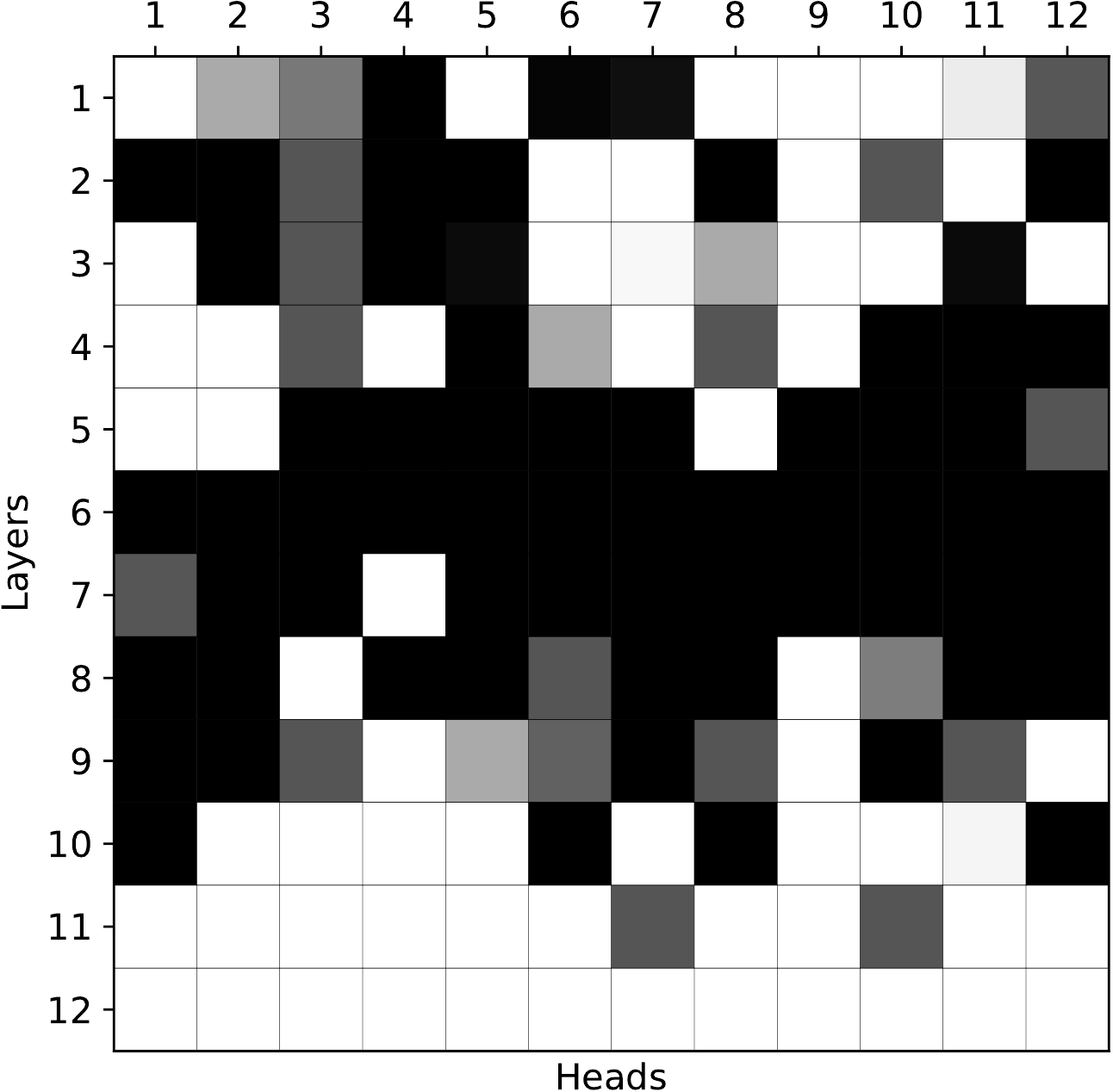}
     \caption{Average head utilization rates among the 5 tasks in 3 runs. Darker cells indicate higher utilization rates.}
     \label{fig:probe-overlay}
 \end{subfigure}
 \hfill
 
\caption{Head utilization of the STL-DP models (a - e, g) and the MTL-DP model (f). The detailed quantification of correlations is shown in Table~\ref{tbl:quantified-correlations}. See also Figure~\ref{fig:overlap-roberta},\ref{fig:overlap-electra} and \ref{fig:overlap-deberta} for similar visualization of other recent TEs.}
\label{fig:overlap}
\vspace{-1.5ex}
\end{figure*}

\subsection{Pruning Experiments}

Table~\ref{tab:pruning-results} shows single-task learning (STL) results using SP and DP on the 5 tasks.
Our DP strategy consistently performs better than the SP strategy as it shows higher accuracy on all tasks and prunes significantly greater numbers of heads except for \CON.
Compared to the STL models without any pruning, the STL-DP models perform well or slightly better\LN for \POS/\NER/\SRL\ due to the $L_0$-regularization, yet use $\approx$$50\%$ fewer numbers of heads.

Comparing the STL-DP models across different tasks, \SRL\ requires more heads than \DEP\ and \CON, which require more than \POS.
This aligns with the intuition behind their difficulty levels as \textit{semantic} > \textit{syntactic} > \textit{lexical} relations. 
On the other hand, \NER\ requires more heads than \CON\ because of the world knowledge it needs to capture from the data; thus, this knowledge is more scattered.

Finally, DP is applied to MTL-5 (MTL-DP), which shows slightly higher accuracy than MTL-5 in Table~\ref{tab:mtl-results} except for \NER\ by pruning $50\%$ of the heads.
This might imply that all tasks want to claim a similar set of heads even though about a half of the heads are underutilized during MTL training.

\subsection{Pruning Visualization}
\label{ssec:pruning-vis}

To visualize the utilization of heads across runs, the utilization rate ${z}^{(r)}_{j,t}$ of the $j$'th head in the $r$'th run for the task $t$ is encoded to the RGB channels:
\begin{equation*}
\mathrm{R/G/B}_{j,t} = 255 \times (1 - {z}^{(1/2/3)}_{j,t})
\end{equation*}
For instance, RGB of $(0, 0, 0)$ is \textit{black} indicating that the head is 100\% utilized in all 3 runs.
Based on this scheme, the head utilizations of all STL-DP models as well as the MTL-DP model are plotted in Figures~\ref{fig:probe-pos} $\sim$ \ref{fig:probe-joint}. 
To depict the overlaps of utilized heads across tasks, ${z}$'s are averaged over all STL-DP models for all runs then plotted as a grayscale heatmap (Figure~\ref{fig:probe-overlay}) using the following scheme:
\begin{equation*}
H_j = \frac{1}{15} \sum_{r=1}^3 \sum_{t=1}^5 z^{(r)}_{j,t}
\end{equation*}

\paragraph{Consistent head utilization by runs}
As shown in Figures~\ref{fig:probe-pos} $\sim$ \ref{fig:probe-srl} and the main diagonal in Table~\ref{tbl:quantified-correlations}, the head utilization per task seems quite similar across different runs, especially for syntactic/semantic tasks such as \DEP/\CON/\SRL.
The head utilization of \POS\ seems to be random because it is a simple task so that high performance can be achieved by a small set of the re-utilized heads.
This consistency across different runs is an essential prerequisite for the following analyses.

\begin{table}[htbp!]
\centering\resizebox{\columnwidth}{!}{
\begin{tabular}{c||c|c|c|c|c|c}
          & \bf \POS & \bf \NER & \bf \DEP & \bf \CON & \bf \SRL & \bf MTL-5 \\
\hline\hline
\bf \POS & \cellcolor{gray!32}75.51 & 74.77 & 76.45 & 83.54 & 73.32 & 78.34 \\
\bf \NER & 74.77 & \cellcolor{gray!32}73.28 & 73.57 & 75.23 & 74.44 & 66.99 \\
\bf \DEP & 76.45 & 73.57 & \cellcolor{gray!32}89.89 & 89.27 & 91.20 & 84.39 \\
\bf \CON & 83.54 & 75.23 & 89.27 & \cellcolor{gray!32}83.99 & 83.90 & 80.71 \\
\bf \SRL & 73.32 & 74.44 & 91.20 & 83.90 & \cellcolor{gray!32}81.01 & 80.07 \\
\bf MTL-5 & 78.34 & 66.99 & 84.39 & 80.71 & 80.07 & \cellcolor{gray!32}85.88 \\
\end{tabular}}
\caption{Adjusted R-squared of 3-run head utilization rates using the third run as the dependent variable (main diagonal highlighted in gray) and Pearson Correlation Coefficient of averaged head utilization rates between each pair of models (non-diagonal cells).}
\label{tbl:quantified-correlations}
\end{table}

\paragraph{Consistent head utilization across tasks}
Figures~\ref{fig:probe-pos} $\sim$ \ref{fig:probe-srl} show that all of these tasks are mostly utilizing heads from layers 5 to 8 (looking like \textbf{M}).
In contrast to \citet{jawahar-etal-2019-bert} and \citet{tenney-etal-2019-bert}, our findings suggest that the middle layers also provide rich surface and semantic features, which are aligned with \citet{liu-etal-2019-linguistic} showing that both \POS\ and chunking tasks perform the best when heads from the middle layers are utilized. 

\paragraph{Consistent head utilization by STL and MTL} 
Figures~\ref{fig:probe-joint} and \ref{fig:probe-overlay} illustrate almost the identical utilization patterns, implying that the MTL-DP model re-uses a very similar set of heads used by the STL-DP models.
According to \citet{vaswani2017attention}, the representation capacity of every head is limited by the design of multi-head attention.
Since (1) a similar set of heads are used across multiple tasks and (2) the limited representation capacity of individual heads confines them to only specific tasks, forcing them for MTL leads to worse results. 
Given this analogy, we propose the following hypothesis:

\noindent \textit{There exists a subset of attention heads in a transformer called ``stem cells'' that are commonly used by many tasks, which cannot be jointly trained for multiple tasks that are very different in nature.}

\noindent We refer to this claim as the \textit{Stem Cell Hypothesis} and seek to test it through the probing analysis.

\section{Probing Analysis}
\label{sec:probing-ana}

This paper hypothesizes the existence of \textit{stem cells}, which cannot be trained to create adequate embeddings to be shared by multiple tasks that are not so similar.
This section provides empirical evidence to this hypothesis by probing what roles each attention head plays once fine-tuned for end tasks.


\subsection{Probing Methods}
\label{ssec:proming-methods}

Previous studies on probing transformer encoders have focused on layer-level analysis limited to supervised probing \cite{hewitt-manning-2019-structural,tenney-etal-2019-bert, lin-etal-2019-open, jawahar-etal-2019-bert, zhao-bethard-2020-berts}.
This section introduces probes on the head-level instead to analyze the impact of fine-tuning on every individual head.
Since developing supervised probes on hundreds of heads requires extensive resource, parameter-free probing methods are used in this study.


\paragraph{Attention Probes} 

Attention between two words often matches a certain linguistic relation that gives a good indicator to knowledge encoded in the head.
Our decoders for \DEP\ and \SRL\ learn relationships between head/dependent words and predicate/argu-ment words respectively, which can be directly benefited from these attentions. 
Thus, the attention matrix from each head is used as the probe of that head. 
Following \citet{clark-etal-2019-bert}, an undirected edge is created between each word and its most attending word.
First, the subtoken-subtoken attention matrix is converted into a word-word matrix by averaging the attention probabilities of each multi-subtoken word.
The $\arg \max$ of each row $r$ in the attention\LN matrix is then calculated, denoted as $\mathbf{g}_r$, and evaluated on the basis of each task. 

For \DEP, directions of the gold arcs are removed and compared against the predicted arcs as follows ($h|d$: the index of a head$|$dependent word, $(h,d)$: an undirected arc from the gold tree, $n$: \# of arcs):

\begin{align*}
\frac{1}{n} \sum_{\forall (h,d)} \mathbbm{1}{ \left(\mathbf{g}_{h} =d \parallel \mathbf{g}_{d} =h \right) }
\end{align*}
For \SRL, we design a new probing method to evaluate how each word in the argument span is attended to the head word in its predicate ($p$: the index of a predicate head word, $T^p$: word indices in the span of $p$'s argument, $m$: \# of predicate-argument pairs):
\begin{equation*}
\frac{1}{m} \sum_{\forall p, \forall T^p} \mathbbm{1}{ \left(\mathbf{g}_{p} \in T^p \parallel \exists\: t \in T^p : \mathbf{g}_t = p \right) }
\end{equation*}

\noindent Only head words in the predicates are used for this analysis, which affects verb-particle constructions (e.g., only \textit{throw} is used for \textit{throw away}).
Moreover, not all words in an argument span are necessarily important to add meaning to its predicate.
We will explore these aspects in the future.

\paragraph{Attended-Value Probes} 

\POS/\NER/\CON\ can be viewed as tasks to find and label spans in a sentence, where the span is a word for \POS, a sequence of consecutive words for \NER\ and \CON, where a span can be overlapped with another span for \CON.
For these tasks, we again present a new probing method, depicted in Algorithm \ref{alg:probe}, that predicts the label of each span based on its representation:

\begin{algorithm}
  \DontPrintSemicolon
  \SetKwFunction{Train}{PseudoCluster}
  \SetKwFunction{Predict}{AVProbe}
  \SetKwProg{Fn}{Function}{:}{}
  \Fn{\Train{$\mathbf{H}, \mathcal{S}$}}{
        $\mathbf{C} \gets \mathbf{0} \in \mathbb{R}^{m \times d}$, $\mathbf{n} \gets \mathbf{0} \in \mathbb{R}^{m\times 1}$ \;
    \ForEach{$(b, e, \ell) \in \mathcal{S}$}{
      $\mathbf{C}_\ell \gets \mathbf{C}_\ell + \texttt{mean}(\mathbf{H}_b, \ldots, \mathbf{H}_e)$ \;
      $\mathbf{n}_\ell \gets \mathbf{n}_\ell + 1$ \;
      }
        \KwRet\ $\nicefrac{\mathbf{C}}{\mathbf{n}}$
  }
  \;
  \SetKwProg{Pn}{Function}{:}{\KwRet}
  \Pn{\Predict{$\mathbf{C}, \mathbf{H}$}}{
        \KwRet\ $\arg \max(\texttt{cossim}(\mathbf{C}, \mathbf{H}))$
  }  
 \caption{Attended-Value Probing}
 \label{alg:probe}
\end{algorithm}

\noindent The attended-value matrix $\mathbf{H}\in \mathbb{R}^{n \times d}$ is created by multiplying the attention matrix $\mathbf{A} \in \mathbb{R}^{n \times n}$ to the value matrix $\mathbf{V} \in \mathbb{R}^{n \times d}$ (Section~\ref{ssec:pruning-regularization}) such that $\mathbf{H} = \mathbf{A} \mathbf{V}$ ($n$: sentence length, $d$: embedding size, abbreviated $d_k$).
$\mathcal{S}$ is the set of gold spans, $m$ is the total number of labels, $(b, e, \ell)$ denotes the indices of the beginning word, the ending word, and the label respectively, and $\texttt{cossim}$ is a cosine similarity function with broadcasting enabled.

With Algorithm \ref{alg:probe}, the centroid of each label is obtained through \texttt{PseudoCluster} then used to predict labels of all spans.
Note that for \CON, only constituents on the height-3 (right above the POS-level) are used for this analysis.
We experimented with constituents on higher levels, which did not show good correlation with model performance as the spans got longer and noisy.
We plan to design  another probing method for deeper analysis in \CON.


\subsection{Probing Experiments}
\label{ssec:probing-experiments}

Probing experiments are conducted on all attention heads in the pre-trained BERT \cite{devlin-etal-2019-bert} and fine-tuned models trained by single-task learning (STL; diagonal in Table~\ref{tab:mtl-results}), pairwise multi-task learning (MTL; other cells in Table~\ref{tab:mtl-results}), and 5-task multitask learning (MTL-5; last column in Table~\ref{tab:mtl-results}) using the two probing methods, attention probes and attended-value probes (Section~\ref{ssec:proming-methods}).
For each model, the head with the highest probing accuracy among 144 heads (12 heads per layer, 12 layers) is selected per label.
Since every model is developed 3 times using different random seeds for better generalization (Section~\ref{ssec:mtl-experiments}), 3 heads are selected per label, which are averaged to get the final probing score for that label.
The full probing results with respect to all labels are described in Appendix~\ref{ssec:probing-results}.

Two important observations are found from these experiments.
(1) Even without fine-tuning, certain heads perform remarkably well on particular labels, confirming the existence of \textit{stem cells} (Sec.~\ref{sec:before-finetune}).\LN
(2) Most heads show higher performance once fine-tuned; nonetheless, MTL does not always enhance them for all tasks.
In fact, MTL models show improvement on only a few labels (Sec.~\ref{sec:after-finetune}), while MTL-5 models show no benefit for most labels.\footnote{The probing results for \DEP\ and \SRL\ are omitted here due to the space limit but explained in Appendix~\ref{ssec:probing-results}.}


\subsubsection{Pluripotent Stem Cells}
\label{sec:before-finetune}

The probing results of pre-trained attention heads from BERT (before fine-tuning) are visualized to verify the existence and pluripotency of \textit{stem cells}. 
These results show very high accuracies for many labels, confirming the existence of \textit{stem cells}. As they reside in the same pre-trained model, their pluripotency is therefore implied. Specifically, the number of probing tasks is 203 (\POS:49, \NER:19, \SRL:67, \DEP:45, \CON:23 as shown in Appendix~\ref{ssec:probing-results-detail}) which is larger than the number of heads (144) in BERT-base by itself. Not all of them provide task specific knowledge as shown in our pruning experiments (Section~\ref{sec:pruning-anlaysis}), so the number of utilized heads is even smaller. As a result, some heads must play multiple roles in different tasks.  

\paragraph{Dependency Parsing}

\begin{figure}[htbp!]
\centering
\includegraphics[width=\columnwidth]{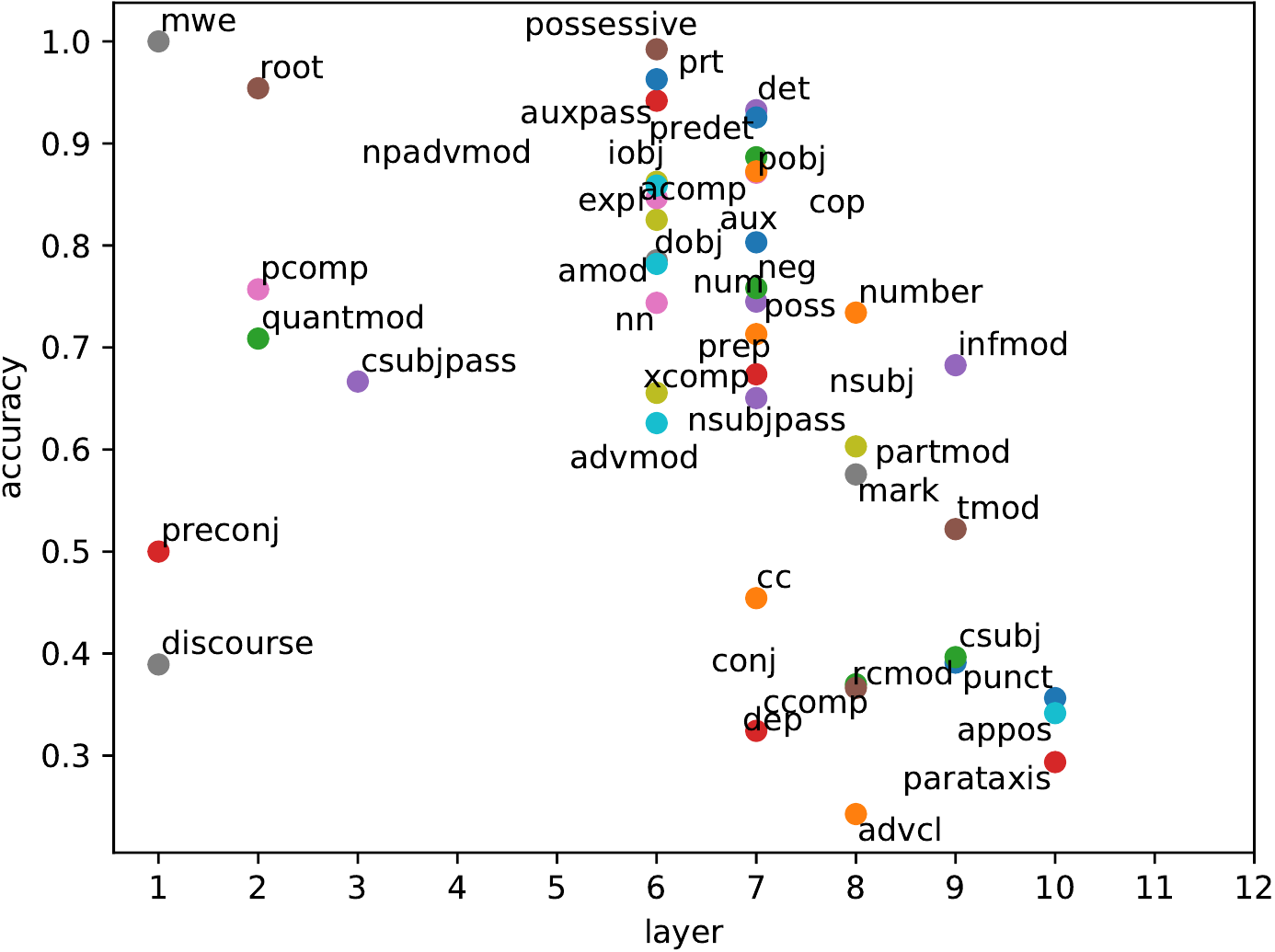}
\caption{\DEP\ layer analysis of pre-trained heads.}
\label{fig:head-dep}
\end{figure}

For \DEP, probing results from the best performing heads with respect to their layers for all labels are plotted in Figure~\ref{fig:head-dep}, some of which are even comparable to supervised results.

\noindent The best performing head of BERT finds the \texttt{ROOT} of a sentence with a $96.25\%$ accuracy without any supervision, demonstrating its ability to convey the concept.
Furthermore, the identification of \texttt{ROOT} happens mostly at the early stage of inference, i.e. in layers 2 and 3. This finding may conflict with the idea of syntactic features getting learned in middle layers \cite{jawahar-etal-2019-bert}. It takes the argument from \citet{tenney-etal-2019-bert} a step further suggesting that syntax can be encoded in early layers of TEs.

\paragraph{Semantic Role Labeling}

\begin{figure}[htbp!]
\centering
\includegraphics[width=\columnwidth]{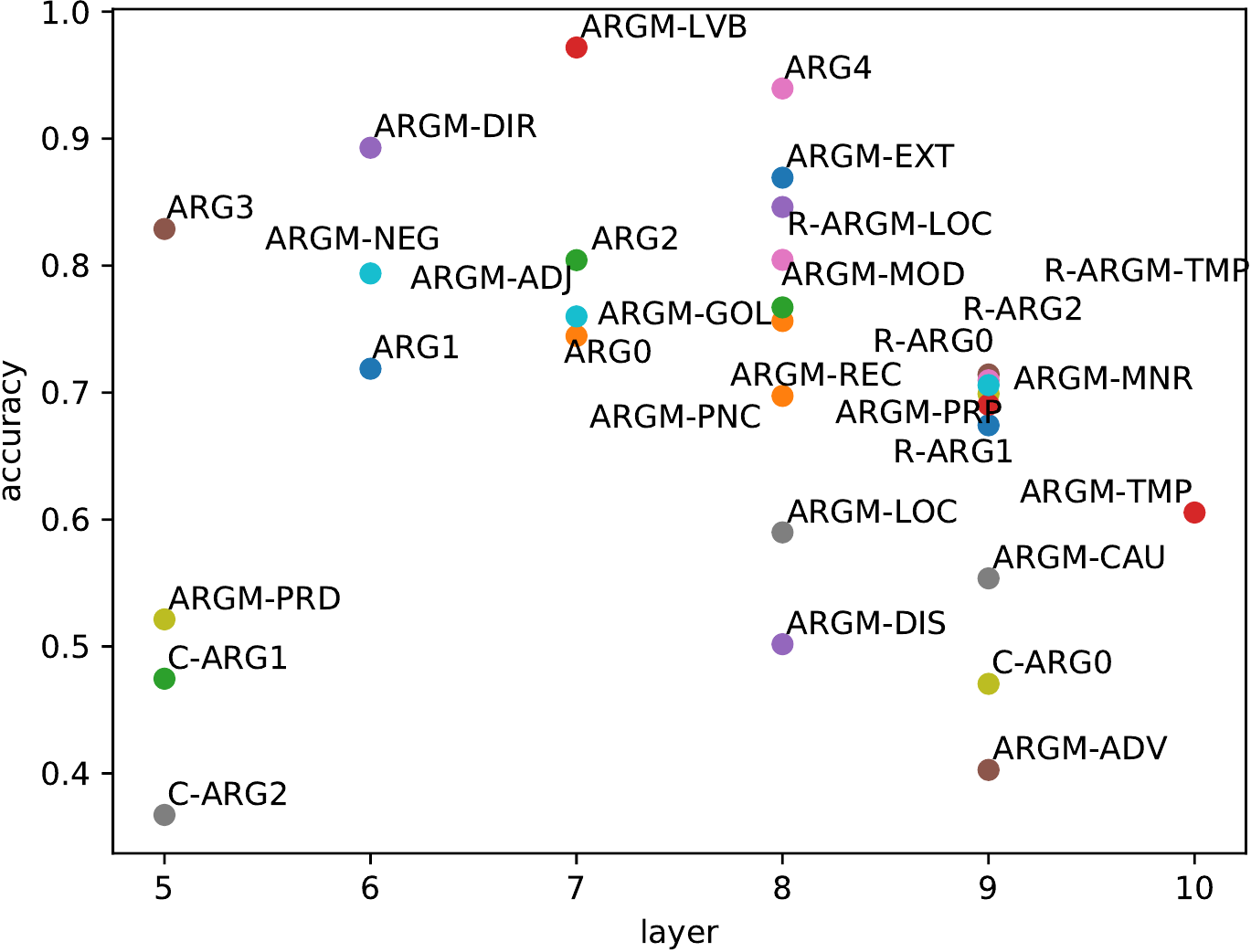}
\caption{\SRL\ layer analysis of pre-trained heads.}
\label{fig:head-srl}
\end{figure}

As shown in Figure~\ref{fig:head-srl}, probing shows promising results on many semantic roles.
Specifically, numbered arguments (\texttt{ARG0-4}) are recognized in layers 5 to 7, while modifiers are identified in layers 8 to 10 with $>80\%$ accuracies, including \texttt{ARGM-MOD} (modals), \texttt{ARGM-DIR} (directional), \texttt{ARGM-EXT} (extent), \texttt{ARGM-LVB} (light-verb), and \texttt{ARGM-COM} (comitative).
Unlike \DEP\ that most labels are learned within the first 7 layers,\LN \SRL\ requires 7+ layers to be learned such that no role reaches the peak before layer 5.
This implies that semantic roles take more efforts to be learned than syntactic dependencies.

%

\subsubsection{Stem Cells Specialization}
\label{sec:after-finetune}
Though \textit{stem cells} are pluripotent, they develop into specialized ones in STL and lose specialities in MTL according to the following comparisons of best performing heads across BERT, STL and MTL models.

\paragraph{Part-of-Speech Tagging}

\begin{figure}[htbp!]
\centering
\includegraphics[width=\columnwidth]{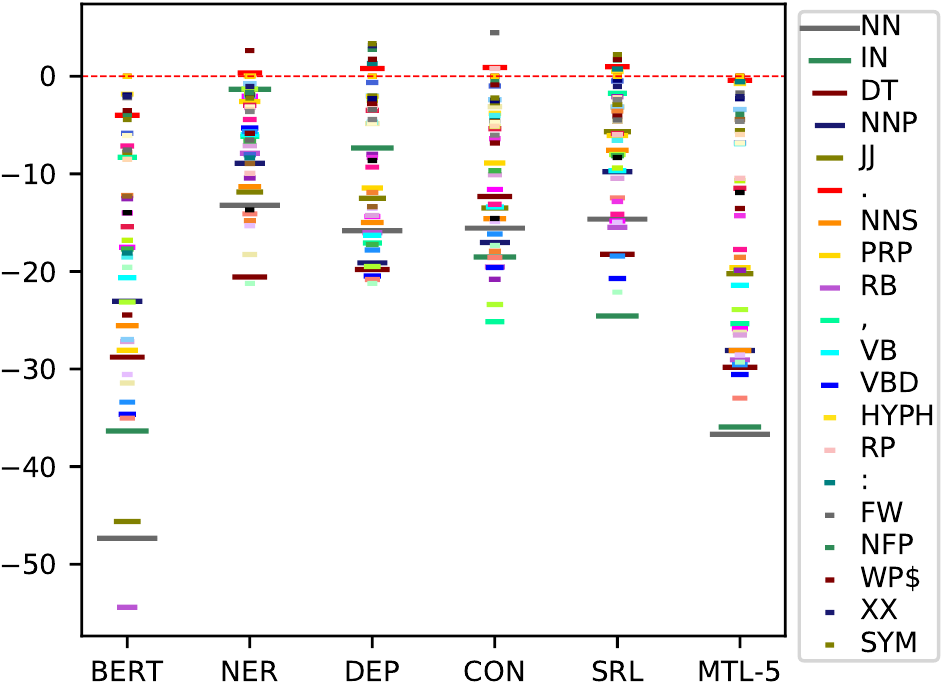}
\caption{\POS\ probing results comparison.}
\label{fig:diff-pos}
\end{figure}

Figure~\ref{fig:diff-pos} compares the heads in the STL model against the other models;\LN the $y$-axis shows the probing results from the model in the $x$-axis subtracted by the results of the STL model.
Labels (sorted by frequency) with negative scores for BERT imply that STL performs better on those labels than BERT (without getting fine-tuned), whereas negative labels with the other models (e.g., \NER, \DEP) imply that the joint models perform worse than the STL model on those labels.

\noindent For \POS, MTL degrades performance for most labels compared to STL.
Even without getting fine-tuned, the pre-trained BERT model performs very well on punctuation labels, which is expected.
The performance on \texttt{WP\$} (possessive wh-pronoun) is significantly improved with \NER, \DEP, and \SRL\ as a possessive wh-pronoun (e.g., \textit{whose}) often follows a name or is used in a relative clause that plays an important role in \DEP\ and \SRL.


\paragraph{Named Entity Recognition}

\begin{figure}[htbp!]
\centering
\includegraphics[width=\columnwidth]{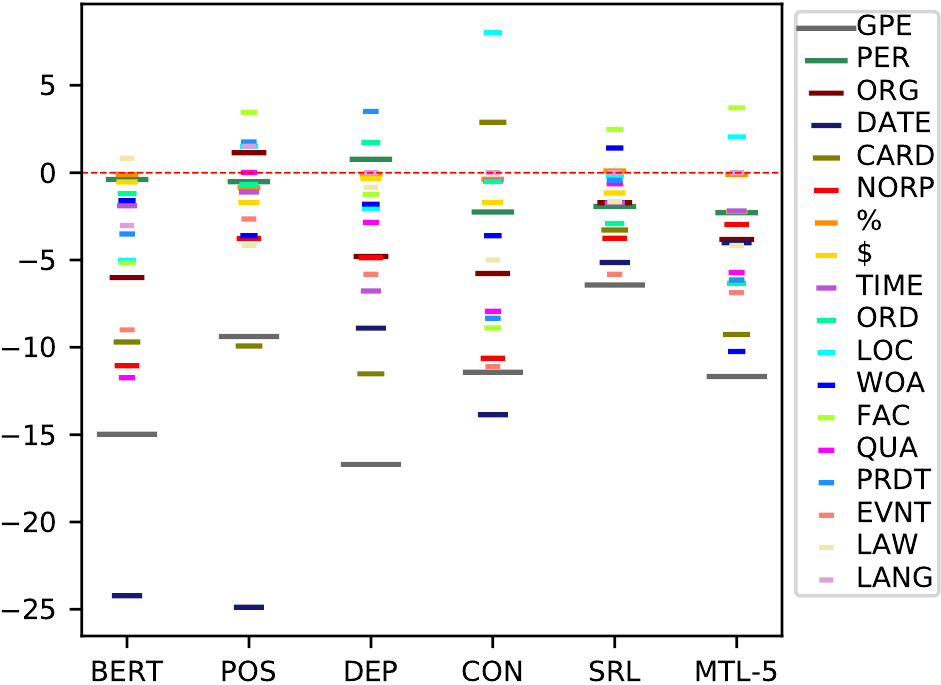}
\caption{\NER\ probing results comparison.}
\label{fig:diff-ner}
\end{figure}

For \NER, BERT detects \texttt{PERCENT}, \texttt{MONEY}, \texttt{LAW}, \texttt{LANGUAGE}, \texttt{NORP} (national$|$religious$|$political groups) and \texttt{PRODUCT} with over $90\%$ probing accuracy, probably due to the rich set of those entities present in pre-training data. 
Although most joint models degrade probing results for nearly every entity type, \POS\ and \DEP\ improve upon more entity types than the other tasks\LN (Figure~\ref{fig:diff-ner}), which is consistent with the results illustrated in Table~\ref{tab:mtl-results}. 

\paragraph{Constituency Parsing}

\begin{figure}[htbp!]
\centering
\includegraphics[width=\columnwidth]{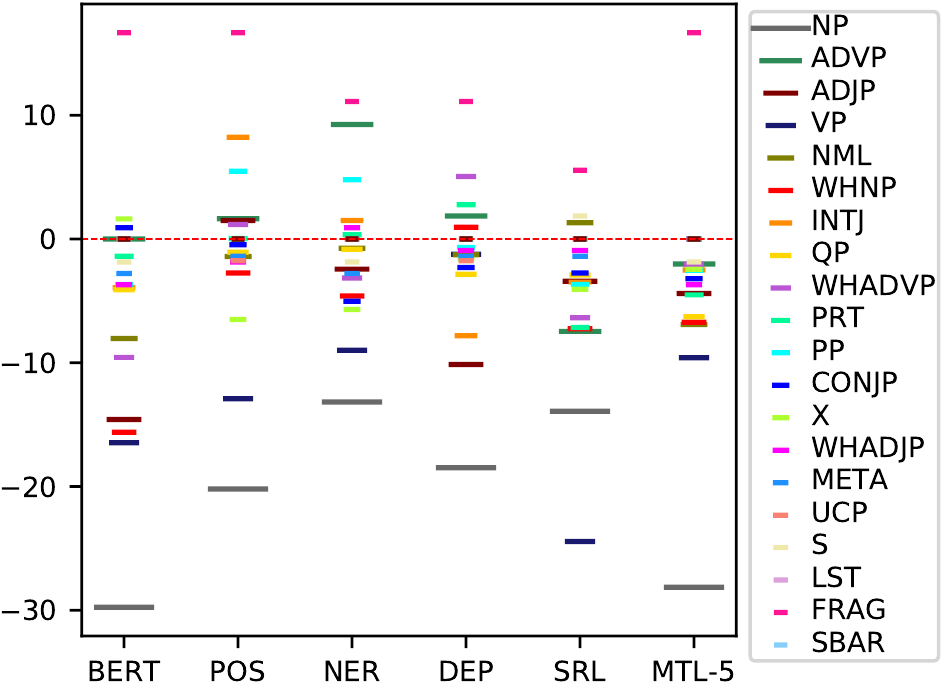}
\caption{\CON\ probing results comparison.}
\label{fig:diff-con}
\end{figure}

As shown in Figure \ref{fig:diff-con}, \POS\ improves the most number of constituent types but also causes the largest drop among the MTL models for the most frequent type, \texttt{NP} (noun phrase). 
This contributes to related constituent types such as \texttt{RB} (adverb) in \texttt{ADVP} (\texttt{RB} phrase), \texttt{WRB} (wh-adverb) in \texttt{WHADVP} (\texttt{WRB} phrase), and \texttt{UH} (interjection) in \texttt{INTJ} (\texttt{UH} phrase). 
Its dramatic decrease on the \texttt{NP} performance might be due to the internal lexical complexity in \texttt{NP}. 

\noindent Regarding \NER, its boost on \texttt{ADVP} can be due\LN to temporal entities (\texttt{TIME}, \texttt{DATE}) nested within \texttt{ADVP} such as \texttt{(}\texttt{ADVP} \texttt{(}\texttt{NP} \textit{one year}\texttt{)} \textit{ago}\texttt{)}, where \textit{one year ago} is a \texttt{DATE} entity.
As for \texttt{PP} (preposition phrase), it usually follows the induction rule of \texttt{PP} $\rightarrow$ \texttt{IN}/\texttt{TO} $+$ \texttt{NP}, where the \texttt{NP} is often an named entity (e.g., (\texttt{PP} (\texttt{TO} \textit{to}) (\texttt{NP} \textit{Mary}))). 
Regarding \DEP, it mainly improves wh-phrases like \texttt{WHNP} and \texttt{WHADVP}, which correspond to \texttt{nsubj} and \texttt{advmod} dependency relations, respectively. 
Regarding \SRL, it slightly improves \texttt{NML} (nominal modifiers) and \texttt{FRAG} (fragment), which may be ascribed to the strength of the span-based SRL not requiring constituency structures for decoding.

Note that in these probing analyses, we selected the best performing heads from each model independently as their locations are not regular anymore. Without the pruning objective (Equation~\ref{eq:l0-loss}), the locations of the best performing heads are non-regular possibly due to the knowledge transfer between \textit{stem cells} and \textit{non-stem cells}. Thus, knowledge transferring from \textit{stem cells} to \textit{non-stem cells} becomes much easier when the models are free to use as many heads as they want. In fact, when fine-tuned without the pruning objective, many \textit{stem cell} attention heads transfer their knowledge to \textit{non-stem cell} heads to get specialized. It is a phenomenon frequently observed in many previous works \cite{tenney-etal-2019-bert, liu-etal-2019-linguistic, jawahar-etal-2019-bert} and this work (Section~\ref{sec:after-finetune}) that certain layers achieve best performance of certain tasks. Given that the \textit{stem cells} of BERT are mostly in middle layers (Section~\ref{ssec:pruning-vis}), we believe that the best performing layers or heads in lower or higher layers are the results of transfer learning on \textit{stem cells}. In reality, a \textit{stem cell} also moves from its original area (e.g., bone marrow) to another area (e.g., bone surface) to get specialized.


 
\section{Conclusion}
\label{sec:conclusion}

This study analyzes interference on the 5 core tasks by highlighting naturally talented attention heads, whose importance turns out to be invariant for many downstream tasks. 
The \textit{Stem Cell Hypothesis} states that these talented heads are like \textit{stem cells} that can develop into experts but not all-rounders.
Our hypothesis is validated by several novel parameter-free probes, revealing the interfered representations of \textit{stem cells}.
We will adapt this work to more tasks and languages for broader generality in the future.

\bibliography{references}
\bibliographystyle{acl_natbib}

\cleardoublepage
\appendix
\section{Appendix}
\label{sec:appendix}

\subsection{Corpus Statistics}

Table~\ref{tab:data-pos} and ~\ref{tab:data-ner} describes statistics of the \POS, \NER, \DEP, \CON, \SRL\ datasets used in our experiments.

\begin{table}[h]
\centering
\begin{tabular}{c||c|c}
    & Sentences & Tokens    \\
    \hline \hline
\texttt{TRN} & 75,187    & 1,299,312 \\
\texttt{DEV} & 9,603     & 163,104   \\
\texttt{TST} & 9,479     & 169,579  
\end{tabular}
\caption{\POS, \DEP, \CON, \SRL\ statistics of OntoNotes 5. \texttt{TRN}/\texttt{DEV}/\texttt{TST}: training/development/evaluation set.}
\label{tab:data-pos}
\end{table}

\begin{table}[h]
\centering
\begin{tabular}{c||c|c}
    & Sentences & Tokens    \\
    \hline \hline
\texttt{TRN} & 59,924    & 1,088,503 \\
\texttt{DEV} & 8,528     & 147,724   \\
\texttt{TST} & 8,262     & 152,728   
\end{tabular}
\caption{\NER\ statistics of OntoNotes 5. \texttt{TRN}/\texttt{DEV}/\texttt{TST}: training/development/evaluation set.}
\label{tab:data-ner}
\end{table}

\subsection{Hyper-Parameter Configuration}
\label{ssec:experimental-settings}

The hyper-parameters used in our models are described in Table \ref{tbl:hyper-param}. 

\begin{table}[htbp!]
	\centering\small
	\begin{tabular}{lr}
		\hline
		\multicolumn{2}{l}{\textbf{BERT Encoder}} \\
		name & bert-base-cased \\
		layers tag& 12 \\
		hidden size & 768 \\
		subword dropout & 0.2 \\
		\hline
		\multicolumn{2}{l}{\textbf{Adam Optimizer}} \\
		encoder lr & 5e-5 \\
		decoder lr & 1e-3 \\
		$\epsilon$ & 1e-8 \\
		epochs & 30 \\
		warm up & 10 \\
		\hline
		\multicolumn{2}{l}{\textbf{NER Decoder}} \\
		MLP size & 150 \\
		dropout & 0.5 \\
		\hline	
		\multicolumn{2}{l}{\textbf{DEP Decoder}} \\
		arc MLP size & 500 \\
		rel MLP size & 100 \\
		dropout & 0.33 \\
		\hline	
		\multicolumn{2}{l}{\textbf{CON Decoder}} \\
		span MLP size & 500 \\
		label MLP size & 100 \\
		dropout & 0.33 \\
		\hline	
		\multicolumn{2}{l}{\textbf{SRL Decoder}} \\
		argument ratio & 0.8 \\
        predicate ratio & 0.4 \\
		span width size & 20 \\
		max Span width & 30 \\
		label MLP size & 100 \\
		dropout & 0.2 \\
		\hline												
	\end{tabular}
	\caption{Hyper-parameters settings. }
	\label{tbl:hyper-param}
\end{table}

\subsection{Extra BERT Probing Results}
\label{ssec:probing-results}

\subsubsection{Dependency Parsing}

\begin{figure}[htbp!]
\centering
\includegraphics[width=\columnwidth]{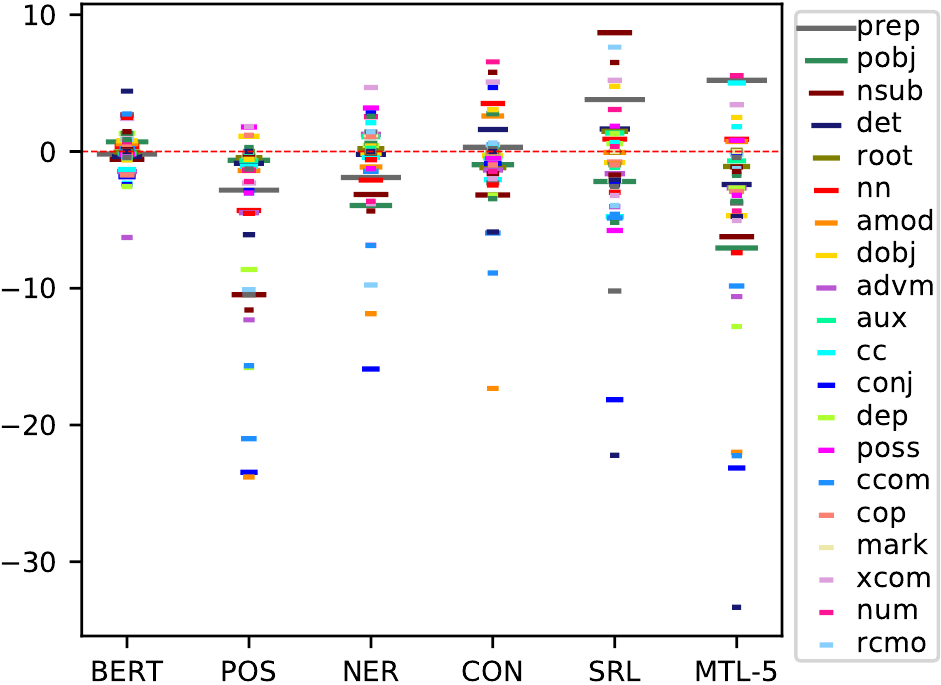}
\caption{\DEP\ probing results comparison.}
\label{fig:diff-dep}
\end{figure}

\noindent Compared to STL, a similar trending with \POS\ and \NER\ can be observed that MTL improves only certain relations as shown in Figure \ref{fig:diff-dep}. Among the 4 tasks, \POS\ improves the least tags in terms of both overall accuracy and probing accuracy. It improves \texttt{dobj} (direct object) and \texttt{expl} (expletive) possibly due to that its decoder needs to assign a verb tag to the \texttt{ROOT} verb and \texttt{EX} (existential there) to ``there'', enhancing the representations of these two. Regarding \NER, it mainly improves modifiers that modify nouns which comprises named entities. In the case of \CON, modifiers and complement arguments are improved, most of which usually reside in \texttt{NP} (Noun Phrase) or \texttt{VP} (Verb Phrase) phrases, placing upper bounds on the distance of dependencies. As regards \SRL, it improves subjects and clausal relations. 

\subsubsection{Semantic Role Labeling}

\begin{figure}[htbp!]
\centering
\includegraphics[width=\columnwidth]{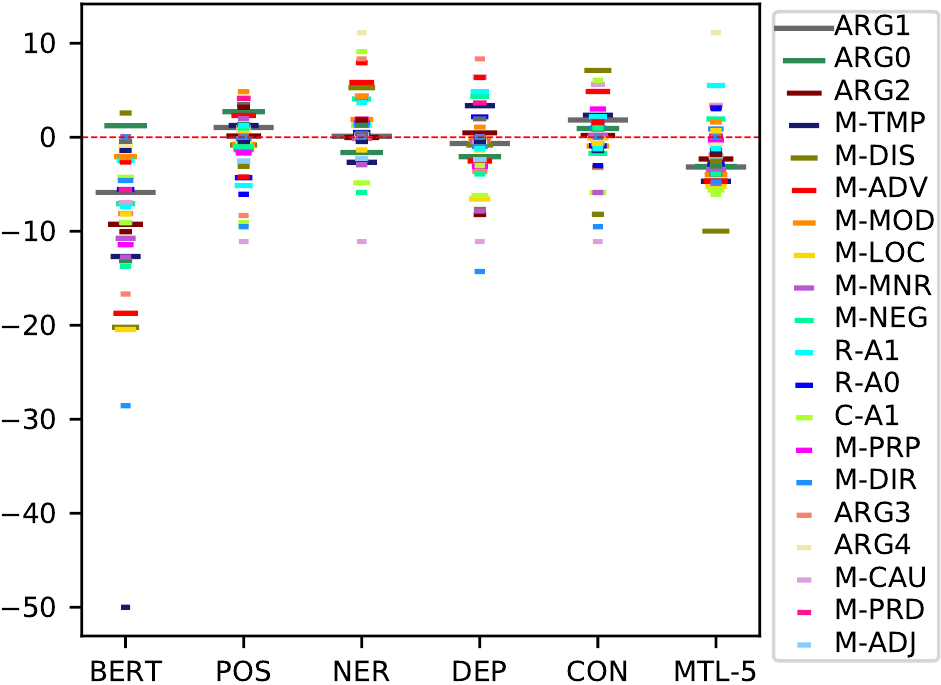}
\caption{\SRL\ probing results comparison.}
\label{fig:diff-srl}
\end{figure}

\noindent In comparison to STL illustrated in Figure \ref{fig:diff-srl}, \POS\ and \CON\ mainly improves \texttt{ARG0-3} (agent, patient, instrument, benefactive, attribute and starting point) and some modifiers including \texttt{ARGM-TMP} (temporal), \texttt{ARGM-CAU} (cause), \texttt{ARGM-PRD} (secondary predication), \texttt{ARGM-EXT} (extent), \texttt{ARGM-PNC} (purpose) and \texttt{ARGM-REC} (reciprocals). Both \POS\ and \CON\ reveal syntactic functions which appear to coordinate attentions on semantic roles in a similar way. Regarding \NER\ and \DEP, they  improve arguments that include referent or pronouns (\texttt{R-ARG1}, \texttt{R-ARG0}) and modifiers (\texttt{ARGM-NEG} which is negation, \texttt{ARGM-CAU}, \texttt{ARGM-PRD}, \texttt{ARGM-EXT}), possibly due to the biaffine decoders they employ analogously enhance the heads.

\subsubsection{Detailed Probing Results}
\label{ssec:probing-results-detail}

With STL as the baseline, probing results for each task are recorded in Table~\ref{tbl:prob-pos} to \ref{tbl:prob-srl} respectively. In each table, the first column shows the labels sorted by their frequencies in the test set, the second column shows the mean probing results over 3 runs for STL which is the baseline, other columns show the probing results for BERT and MTL models with their differences against the STL baseline (in parentheses, \dG{green} indicates improvements and \dR{red} indicates decreases). BERT probing scores are usually much lower than STL, which is expected as it is not fine-tuned on the task data. MTL probing scores are also generally lower than STL, implying interference with other joint tasks.

\begin{table*}[htbp]
\centering\resizebox{\textwidth}{!}{
\begin{tabular}{l||l|l|l|l|l|l|l}
      & \textbf{STL}   & \textbf{BERT}           & \textbf{NER}           & \textbf{DEP}           & \textbf{CON}           & \textbf{SRL}           & \textbf{MTL-5}         \\
      \hline\hline
\texttt{NN} & \textbf{91.44} & 44.11\negdiff{-47.33} & 78.22\negdiff{-13.22} & 75.61\negdiff{-15.83} & 75.88\negdiff{-15.56} & 76.80\negdiff{-14.64} & 54.75\negdiff{-36.69} \\ 
\texttt{IN} & \textbf{89.79} & 53.45\negdiff{-36.34} & 88.46\negdiff{-1.33} & 82.43\negdiff{-7.36} & 71.27\negdiff{-18.52} & 65.22\negdiff{-24.57} & 53.86\negdiff{-35.93} \\ 
\texttt{DT} & \textbf{93.93} & 65.15\negdiff{-28.78} & 73.36\negdiff{-20.57} & 74.13\negdiff{-19.80} & 81.62\negdiff{-12.31} & 75.69\negdiff{-18.24} & 64.11\negdiff{-29.82} \\ 
\texttt{NNP} & \textbf{90.21} & 67.16\negdiff{-23.05} & 81.29\negdiff{-8.92} & 71.09\negdiff{-19.12} & 73.17\negdiff{-17.04} & 80.42\negdiff{-9.79} & 62.11\negdiff{-28.10} \\ 
\texttt{JJ} & \textbf{83.59} & 37.96\negdiff{-45.63} & 71.73\negdiff{-11.86} & 71.10\negdiff{-12.49} & 70.09\negdiff{-13.50} & 77.91\negdiff{-5.68} & 63.35\negdiff{-20.24} \\ 
\texttt{.} & 95.02 & 91.03\negdiff{-3.99} & 95.36\posdiff{0.34} & 95.82\posdiff{0.80} & 95.91\posdiff{0.89} & \textbf{95.99}\posdiff{0.97} & 94.61\negdiff{-0.41} \\ 
\texttt{NNS} & \textbf{91.75} & 66.19\negdiff{-25.56} & 80.43\negdiff{-11.32} & 76.77\negdiff{-14.98} & 77.16\negdiff{-14.59} & 84.17\negdiff{-7.58} & 63.66\negdiff{-28.09} \\ 
\texttt{PRP} & \textbf{93.76} & 65.69\negdiff{-28.07} & 91.17\negdiff{-2.59} & 82.31\negdiff{-11.45} & 84.87\negdiff{-8.89} & 87.68\negdiff{-6.08} & 74.13\negdiff{-19.63} \\ 
\texttt{RB} & \textbf{83.33} & 28.92\negdiff{-54.41} & 75.44\negdiff{-7.89} & 67.30\negdiff{-16.03} & 63.78\negdiff{-19.55} & 67.86\negdiff{-15.47} & 54.27\negdiff{-29.06} \\ 
\texttt{,} & \textbf{95.23} & 86.94\negdiff{-8.29} & 89.07\negdiff{-6.16} & 78.15\negdiff{-17.08} & 70.08\negdiff{-25.15} & 93.50\negdiff{-1.73} & 69.87\negdiff{-25.36} \\ 
\texttt{VB} & \textbf{90.17} & 69.52\negdiff{-20.65} & 84.21\negdiff{-5.96} & 73.87\negdiff{-16.30} & 76.77\negdiff{-13.40} & 80.49\negdiff{-9.68} & 68.76\negdiff{-21.41} \\ 
\texttt{VBD} & \textbf{90.97} & 56.33\negdiff{-34.64} & 85.68\negdiff{-5.29} & 70.51\negdiff{-20.46} & 71.36\negdiff{-19.61} & 70.25\negdiff{-20.72} & 60.42\negdiff{-30.55} \\ 
\texttt{CC} & \textbf{91.40} & 68.26\negdiff{-23.14} & 90.09\negdiff{-1.31} & 71.91\negdiff{-19.49} & 68.02\negdiff{-23.38} & 83.39\negdiff{-8.01} & 67.49\negdiff{-23.91} \\ 
\texttt{VBZ} & \textbf{92.44} & 74.91\negdiff{-17.53} & 90.40\negdiff{-2.04} & 78.07\negdiff{-14.37} & 80.85\negdiff{-11.59} & 77.62\negdiff{-14.82} & 66.56\negdiff{-25.88} \\ 
\texttt{VBP} & \textbf{90.02} & 56.63\negdiff{-33.39} & 84.35\negdiff{-5.67} & 72.21\negdiff{-17.81} & 73.85\negdiff{-16.17} & 71.60\negdiff{-18.42} & 60.45\negdiff{-29.57} \\ 
\texttt{VBN} & \textbf{88.81} & 53.78\negdiff{-35.03} & 74.71\negdiff{-14.10} & 67.99\negdiff{-20.82} & 70.24\negdiff{-18.57} & 76.35\negdiff{-12.46} & 55.83\negdiff{-32.98} \\ 
\texttt{CD} & \textbf{93.10} & 61.67\negdiff{-31.43} & 74.84\negdiff{-18.26} & 88.25\negdiff{-4.85} & 89.97\negdiff{-3.13} & 89.95\negdiff{-3.15} & 66.87\negdiff{-26.23} \\ 
\texttt{VBG} & \textbf{92.63} & 65.44\negdiff{-27.19} & 85.51\negdiff{-7.12} & 78.36\negdiff{-14.27} & 82.53\negdiff{-10.10} & 82.19\negdiff{-10.44} & 66.12\negdiff{-26.51} \\ 
\texttt{TO} & \textbf{92.38} & 85.24\negdiff{-7.14} & 87.95\negdiff{-4.43} & 83.07\negdiff{-9.31} & 79.26\negdiff{-13.12} & 78.24\negdiff{-14.14} & 74.63\negdiff{-17.75} \\ 
\texttt{MD} & \textbf{92.70} & 65.72\negdiff{-26.98} & 91.93\negdiff{-0.77} & 90.69\negdiff{-2.01} & 90.29\negdiff{-2.41} & 89.55\negdiff{-3.15} & 89.29\negdiff{-3.41} \\ 
\texttt{PRP\$} & \textbf{92.33} & 76.90\negdiff{-15.43} & 89.37\negdiff{-2.96} & 88.84\negdiff{-3.49} & 86.96\negdiff{-5.37} & 86.48\negdiff{-5.85} & 80.86\negdiff{-11.47} \\ 
\texttt{UH} & \textbf{93.66} & 75.95\negdiff{-17.71} & 87.03\negdiff{-6.63} & 76.42\negdiff{-17.24} & 83.97\negdiff{-9.69} & 85.52\negdiff{-8.14} & 73.79\negdiff{-19.87} \\ 
\texttt{HYPH} & 91.98 & 90.15\negdiff{-1.83} & 92.01\posdiff{0.03} & 89.90\negdiff{-2.08} & 88.20\negdiff{-3.78} & \textbf{92.47}\posdiff{0.49} & 91.23\negdiff{-0.75} \\ 
\texttt{POS} & \textbf{98.85} & 93.01\negdiff{-5.84} & 90.80\negdiff{-8.05} & 98.21\negdiff{-0.64} & 97.86\negdiff{-0.99} & 98.37\negdiff{-0.48} & 92.02\negdiff{-6.83} \\ 
\texttt{'} & \textbf{89.05} & 76.84\negdiff{-12.21} & 74.25\negdiff{-14.80} & 77.14\negdiff{-11.91} & 71.12\negdiff{-17.93} & 85.48\negdiff{-3.57} & 70.51\negdiff{-18.54} \\ 
\texttt{``} & \textbf{89.54} & 76.98\negdiff{-12.56} & 79.12\negdiff{-10.42} & 81.55\negdiff{-7.99} & 68.74\negdiff{-20.80} & 87.44\negdiff{-2.10} & 69.67\negdiff{-19.87} \\ 
\texttt{WDT} & \textbf{87.26} & 68.74\negdiff{-18.52} & 85.27\negdiff{-1.99} & 82.54\negdiff{-4.72} & 83.23\negdiff{-4.03} & 80.71\negdiff{-6.55} & 80.34\negdiff{-6.92} \\ 
\texttt{WP} & \textbf{90.19} & 76.18\negdiff{-14.01} & 86.99\negdiff{-3.20} & 81.88\negdiff{-8.31} & 83.79\negdiff{-6.40} & 77.35\negdiff{-12.84} & 75.89\negdiff{-14.30} \\ 
\texttt{WRB} & \textbf{89.15} & 72.35\negdiff{-16.80} & 87.38\negdiff{-1.77} & 86.68\negdiff{-2.47} & 83.99\negdiff{-5.16} & 79.75\negdiff{-9.40} & 78.44\negdiff{-10.71} \\ 
\texttt{RP} & 82.13 & 74.33\negdiff{-7.80} & 72.20\negdiff{-9.93} & 79.06\negdiff{-3.07} & \textbf{82.93}\posdiff{0.80} & 76.13\negdiff{-6.00} & 71.68\negdiff{-10.45} \\ 
\texttt{:} & 84.58 & 66.48\negdiff{-18.10} & 76.22\negdiff{-8.36} & \textbf{85.89}\posdiff{1.31} & 80.03\negdiff{-4.55} & 85.33\posdiff{0.75} & 84.02\negdiff{-0.56} \\ 
\texttt{JJR} & \textbf{87.18} & 56.62\negdiff{-30.56} & 71.87\negdiff{-15.31} & 73.65\negdiff{-13.53} & 72.29\negdiff{-14.89} & 72.22\negdiff{-14.96} & 58.62\negdiff{-28.56} \\ 
\texttt{NNPS} & \textbf{84.39} & 72.11\negdiff{-12.28} & 75.45\negdiff{-8.94} & 71.02\negdiff{-13.37} & 79.74\negdiff{-4.65} & 79.81\negdiff{-4.58} & 79.96\negdiff{-4.43} \\ 
\texttt{EX} & \textbf{95.33} & 89.25\negdiff{-6.08} & 92.18\negdiff{-3.15} & 90.55\negdiff{-4.78} & 90.56\negdiff{-4.77} & 90.77\negdiff{-4.56} & 88.49\negdiff{-6.84} \\ 
\texttt{JJS} & \textbf{83.41} & 58.95\negdiff{-24.46} & 77.58\negdiff{-5.83} & 80.64\negdiff{-2.77} & 76.57\negdiff{-6.84} & 79.33\negdiff{-4.08} & 69.87\negdiff{-13.54} \\ 
\texttt{RBR} & \textbf{87.29} & 67.71\negdiff{-19.58} & 66.07\negdiff{-21.22} & 66.07\negdiff{-21.22} & 69.96\negdiff{-17.33} & 65.17\negdiff{-22.12} & 58.00\negdiff{-29.29} \\ 
\texttt{-LRB-} & \textbf{91.54} & 83.76\negdiff{-7.78} & 89.68\negdiff{-1.86} & 89.51\negdiff{-2.03} & 88.83\negdiff{-2.71} & 88.66\negdiff{-2.88} & 85.96\negdiff{-5.58} \\ 
\texttt{-RRB-} & \textbf{92.18} & 83.67\negdiff{-8.51} & 88.78\negdiff{-3.40} & 88.78\negdiff{-3.40} & 87.07\negdiff{-5.11} & 89.97\negdiff{-2.21} & 86.22\negdiff{-5.96} \\ 
\texttt{\$} & \textbf{97.30} & 95.38\negdiff{-1.92} & 95.38\negdiff{-1.92} & 94.99\negdiff{-2.31} & 94.80\negdiff{-2.50} & 96.92\negdiff{-0.38} & 94.99\negdiff{-2.31} \\ 
\texttt{PDT} & \textbf{87.15} & 84.94\negdiff{-2.21} & 83.54\negdiff{-3.61} & 82.73\negdiff{-4.42} & 81.12\negdiff{-6.03} & 84.74\negdiff{-2.41} & 82.53\negdiff{-4.62} \\ 
\texttt{RBS} & \textbf{86.31} & 72.32\negdiff{-13.99} & 72.62\negdiff{-13.69} & 77.68\negdiff{-8.63} & 71.73\negdiff{-14.58} & 77.98\negdiff{-8.33} & 74.41\negdiff{-11.90} \\ 
\texttt{FW} & 72.51 & 64.95\negdiff{-7.56} & 65.98\negdiff{-6.53} & 69.07\negdiff{-3.44} & \textbf{76.97}\posdiff{4.46} & 68.04\negdiff{-4.47} & 70.79\negdiff{-1.72} \\ 
\texttt{NFP} & 90.56 & 86.67\negdiff{-3.89} & 88.89\negdiff{-1.67} & \textbf{93.33}\posdiff{2.77} & 90.00\negdiff{-0.56} & 92.22\posdiff{1.66} & 86.67\negdiff{-3.89} \\ 
\texttt{WP\$} & 92.99 & 89.47\negdiff{-3.52} & \textbf{95.62}\posdiff{2.63} & 94.74\posdiff{1.75} & 92.11\negdiff{-0.88} & 94.74\posdiff{1.75} & 92.98\negdiff{-0.01} \\ 
\texttt{XX} & 83.34 & 81.25\negdiff{-2.09} & 82.29\negdiff{-1.05} & \textbf{86.46}\posdiff{3.12} & 83.33\negdiff{-0.01} & 82.29\negdiff{-1.05} & 81.25\negdiff{-2.09} \\ 
\texttt{SYM} & 91.11 & 86.67\negdiff{-4.44} & 88.89\negdiff{-2.22} & \textbf{94.45}\posdiff{3.34} & 88.89\negdiff{-2.22} & 93.33\posdiff{2.22} & 91.11\negdiff{0.00} \\ 
\texttt{ADD} & \textbf{100.00} & \textbf{100.00}\negdiff{0.00} & \textbf{100.00}\negdiff{0.00} & \textbf{100.00}\negdiff{0.00} & \textbf{100.00}\negdiff{0.00} & \textbf{100.00}\negdiff{0.00} & \textbf{100.00}\negdiff{0.00} \\ 
\texttt{LS} & \textbf{100.00} & \textbf{100.00}\negdiff{0.00} & \textbf{100.00}\negdiff{0.00} & \textbf{100.00}\negdiff{0.00} & \textbf{100.00}\negdiff{0.00} & \textbf{100.00}\negdiff{0.00} & \textbf{100.00}\negdiff{0.00} \\ 
\texttt{AFX} & \textbf{100.00} & \textbf{100.00}\negdiff{0.00} & \textbf{100.00}\negdiff{0.00} & \textbf{100.00}\negdiff{0.00} & \textbf{100.00}\negdiff{0.00} & \textbf{100.00}\negdiff{0.00} & \textbf{100.00}\negdiff{0.00} \\ 
\end{tabular}
}
\caption{Probing results for POS. STL: the basline, BERT:pre-trained BERT before fine-tuning.}
\label{tbl:prob-pos}
\end{table*}


\begin{table*}[htbp]
\centering\resizebox{\textwidth}{!}{
\begin{tabular}{l||l|l|l|l|l|l|l}
      & \textbf{STL}   & \textbf{BERT}           & \textbf{POS}           & \textbf{DEP}           & \textbf{CON}           & \textbf{SRL}           & \textbf{MTL-5}         \\
      \hline\hline
\texttt{GPE} & \textbf{91.40} & 76.43\negdiff{-14.97} & 82.02\negdiff{-9.38} & 74.70\negdiff{-16.70} & 79.97\negdiff{-11.43} & 84.97\negdiff{-6.43} & 79.73\negdiff{-11.67} \\ 
\texttt{PERSON} & 85.90 & 85.51\negdiff{-0.39} & 85.38\negdiff{-0.52} & \textbf{86.67}\posdiff{0.77} & 83.65\negdiff{-2.25} & 83.96\negdiff{-1.94} & 83.60\negdiff{-2.30} \\ 
\texttt{ORG} & 73.81 & 67.80\negdiff{-6.01} & \textbf{74.95}\posdiff{1.14} & 69.01\negdiff{-4.80} & 68.04\negdiff{-5.77} & 72.09\negdiff{-1.72} & 69.99\negdiff{-3.82} \\ 
\texttt{DATE} & \textbf{87.39} & 63.17\negdiff{-24.22} & 62.51\negdiff{-24.88} & 78.49\negdiff{-8.90} & 73.53\negdiff{-13.86} & 82.25\negdiff{-5.14} & 83.37\negdiff{-4.02} \\ 
\texttt{CARDINAL} & 76.33 & 66.63\negdiff{-9.70} & 66.41\negdiff{-9.92} & 64.81\negdiff{-11.52} & \textbf{79.22}\posdiff{2.89} & 73.05\negdiff{-3.28} & 67.06\negdiff{-9.27} \\ 
\texttt{NORP} & \textbf{93.34} & 82.28\negdiff{-11.06} & 89.57\negdiff{-3.77} & 88.47\negdiff{-4.87} & 82.72\negdiff{-10.62} & 89.58\negdiff{-3.76} & 90.37\negdiff{-2.97} \\ 
\texttt{PERCENT} & 99.81 & 99.71\negdiff{-0.10} & 98.95\negdiff{-0.86} & 99.71\negdiff{-0.10} & 99.43\negdiff{-0.38} & \textbf{99.90}\posdiff{0.09} & 99.71\negdiff{-0.10} \\ 
\texttt{MONEY} & \textbf{98.30} & 97.77\negdiff{-0.53} & 96.60\negdiff{-1.70} & 97.98\negdiff{-0.32} & 96.60\negdiff{-1.70} & 97.13\negdiff{-1.17} & 96.07\negdiff{-2.23} \\ 
\texttt{TIME} & \textbf{79.25} & 77.36\negdiff{-1.89} & 78.14\negdiff{-1.11} & 72.48\negdiff{-6.77} & 78.77\negdiff{-0.48} & 77.51\negdiff{-1.74} & 77.05\negdiff{-2.20} \\ 
\texttt{ORDINAL} & 90.94 & 89.74\negdiff{-1.20} & 90.26\negdiff{-0.68} & \textbf{92.65}\posdiff{1.71} & 90.43\negdiff{-0.51} & 88.03\negdiff{-2.91} & 84.61\negdiff{-6.33} \\ 
\texttt{LOC} & 74.86 & 69.83\negdiff{-5.03} & 76.35\posdiff{1.49} & 72.81\negdiff{-2.05} & \textbf{82.87}\posdiff{8.01} & 74.68\negdiff{-0.18} & 76.91\posdiff{2.05} \\ 
\texttt{WORK\_OF\_ART} & 78.11 & 76.51\negdiff{-1.60} & 74.50\negdiff{-3.61} & 76.31\negdiff{-1.80} & 74.50\negdiff{-3.61} & \textbf{79.52}\posdiff{1.41} & 67.87\negdiff{-10.24} \\ 
\texttt{FAC} & 82.96 & 77.78\negdiff{-5.18} & 86.42\posdiff{3.46} & 81.73\negdiff{-1.23} & 74.07\negdiff{-8.89} & 85.43\posdiff{2.47} & \textbf{86.67}\posdiff{3.71} \\ 
\texttt{QUANTITY} & 95.55 & 83.81\negdiff{-11.74} & \textbf{95.56}\posdiff{0.01} & 92.70\negdiff{-2.85} & 87.62\negdiff{-7.93} & 94.92\negdiff{-0.63} & 89.84\negdiff{-5.71} \\ 
\texttt{PRODUCT} & 85.09 & 81.58\negdiff{-3.51} & 86.84\posdiff{1.75} & \textbf{88.60}\posdiff{3.51} & 76.75\negdiff{-8.34} & 84.65\negdiff{-0.44} & 78.95\negdiff{-6.14} \\ 
\texttt{EVENT} & \textbf{77.25} & 68.25\negdiff{-9.00} & 74.60\negdiff{-2.65} & 71.43\negdiff{-5.82} & 66.14\negdiff{-11.11} & 71.43\negdiff{-5.82} & 70.37\negdiff{-6.88} \\ 
\texttt{LAW} & 91.67 & \textbf{92.50}\posdiff{0.83} & 87.50\negdiff{-4.17} & 90.83\negdiff{-0.84} & 86.67\negdiff{-5.00} & 90.00\negdiff{-1.67} & 87.50\negdiff{-4.17} \\ 
\texttt{LANGUAGE} & 98.48 & 95.45\negdiff{-3.03} & \textbf{100.00}\posdiff{1.52} & 98.48\negdiff{0.00} & 98.48\negdiff{0.00} & 98.48\negdiff{0.00} & 98.48\negdiff{0.00} \\ 
\end{tabular}
}
\caption{Probing results for NER. STL: the basline, BERT:pre-trained BERT before fine-tuning.}
\label{tbl:prob-ner}
\end{table*}


\begin{table*}[htbp]
\centering\resizebox{\textwidth}{!}{
\begin{tabular}{l||l|l|l|l|l|l|l}
      & \textbf{STL}   & \textbf{BERT}           & \textbf{POS}           & \textbf{NER}           & \textbf{CON}           & \textbf{SRL}           & \textbf{MTL-5}         \\
      \hline\hline
\texttt{prep} & 72.80 & 72.62\negdiff{-0.18} & 69.98\negdiff{-2.82} & 70.90\negdiff{-1.90} & 73.11\posdiff{0.31} & 76.60\posdiff{3.80} & \textbf{78.02}\posdiff{5.22} \\ 
\texttt{pobj} & 95.64 & \textbf{96.35}\posdiff{0.71} & 95.01\negdiff{-0.63} & 91.70\negdiff{-3.94} & 94.67\negdiff{-0.97} & 93.45\negdiff{-2.19} & 88.59\negdiff{-7.05} \\ 
\texttt{nsubj} & 76.92 & 76.36\negdiff{-0.56} & 66.45\negdiff{-10.47} & 73.78\negdiff{-3.14} & 73.75\negdiff{-3.17} & \textbf{85.61}\posdiff{8.69} & 70.69\negdiff{-6.23} \\ 
\texttt{det} & 94.07 & 93.70\negdiff{-0.37} & 93.22\negdiff{-0.85} & 93.87\negdiff{-0.20} & 95.68\posdiff{1.61} & \textbf{95.72}\posdiff{1.65} & 91.66\negdiff{-2.41} \\ 
\texttt{root} & 96.28 & 96.25\negdiff{-0.03} & 95.83\negdiff{-0.45} & 96.49\posdiff{0.21} & 95.10\negdiff{-1.18} & \textbf{97.77}\posdiff{1.49} & 95.18\negdiff{-1.10} \\ 
\texttt{nn} & 89.08 & 89.47\posdiff{0.39} & 84.77\negdiff{-4.31} & 86.99\negdiff{-2.09} & \textbf{92.59}\posdiff{3.51} & 89.99\posdiff{0.91} & 89.98\posdiff{0.90} \\ 
\texttt{amod} & 92.76 & 93.33\posdiff{0.57} & 91.38\negdiff{-1.38} & 91.63\negdiff{-1.13} & \textbf{95.36}\posdiff{2.60} & 92.70\negdiff{-0.06} & 93.50\posdiff{0.74} \\ 
\texttt{dobj} & 93.76 & 94.55\posdiff{0.79} & \textbf{94.88}\posdiff{1.12} & 93.60\negdiff{-0.16} & 93.46\negdiff{-0.30} & 92.95\negdiff{-0.81} & 89.07\negdiff{-4.69} \\ 
\texttt{advmod} & 70.58 & 69.12\negdiff{-1.46} & 66.11\negdiff{-4.47} & \textbf{71.85}\posdiff{1.27} & 69.23\negdiff{-1.35} & 68.96\negdiff{-1.62} & 67.92\negdiff{-2.66} \\ 
\texttt{aux} & 84.70 & 84.53\negdiff{-0.17} & 84.03\negdiff{-0.67} & 85.78\posdiff{1.08} & 84.22\negdiff{-0.48} & \textbf{86.06}\posdiff{1.36} & 84.02\negdiff{-0.68} \\ 
\texttt{cc} & 59.17 & 57.86\negdiff{-1.31} & 58.10\negdiff{-1.07} & 58.73\negdiff{-0.44} & 57.11\negdiff{-2.06} & 54.40\negdiff{-4.77} & \textbf{64.18}\posdiff{5.01} \\ 
\texttt{conj} & \textbf{66.66} & 64.82\negdiff{-1.84} & 43.20\negdiff{-23.46} & 50.76\negdiff{-15.90} & 65.78\negdiff{-0.88} & 48.50\negdiff{-18.16} & 43.52\negdiff{-23.14} \\ 
\texttt{dep} & 41.44 & \textbf{42.77}\posdiff{1.33} & 32.82\negdiff{-8.62} & 41.88\posdiff{0.44} & 41.22\negdiff{-0.22} & 42.01\posdiff{0.57} & 38.78\negdiff{-2.66} \\ 
\texttt{poss} & 78.20 & 77.96\negdiff{-0.24} & 79.99\posdiff{1.79} & \textbf{81.39}\posdiff{3.19} & 77.72\negdiff{-0.48} & 72.42\negdiff{-5.78} & 79.02\posdiff{0.82} \\ 
\texttt{ccomp} & 71.24 & \textbf{71.47}\posdiff{0.23} & 50.24\negdiff{-21.00} & 69.80\negdiff{-1.44} & 65.29\negdiff{-5.95} & 66.38\negdiff{-4.86} & 61.40\negdiff{-9.84} \\ 
\texttt{cop} & 88.80 & 87.09\negdiff{-1.71} & 87.47\negdiff{-1.33} & \textbf{89.91}\posdiff{1.11} & 88.74\negdiff{-0.06} & 88.04\negdiff{-0.76} & 85.88\negdiff{-2.92} \\ 
\texttt{mark} & 89.78 & 90.88\posdiff{1.10} & 89.44\negdiff{-0.34} & \textbf{91.01}\posdiff{1.23} & 90.29\posdiff{0.51} & 87.65\negdiff{-2.13} & 89.75\negdiff{-0.03} \\ 
\texttt{xcomp} & 68.11 & 70.53\posdiff{2.42} & 65.81\negdiff{-2.30} & 72.80\posdiff{4.69} & 73.20\posdiff{5.09} & \textbf{73.33}\posdiff{5.22} & 71.54\posdiff{3.43} \\ 
\texttt{num} & 84.94 & 87.61\posdiff{2.67} & 83.58\negdiff{-1.36} & 87.48\posdiff{2.54} & \textbf{91.50}\posdiff{6.56} & 88.02\posdiff{3.08} & 90.49\posdiff{5.55} \\ 
\texttt{rcmod} & 51.72 & 52.62\posdiff{0.90} & 41.62\negdiff{-10.10} & 41.96\negdiff{-9.76} & 52.22\posdiff{0.50} & \textbf{59.36}\posdiff{7.64} & 51.52\negdiff{-0.20} \\ 
\texttt{advcl} & 52.26 & \textbf{54.93}\posdiff{2.67} & 41.78\negdiff{-10.48} & 53.70\posdiff{1.44} & 49.97\negdiff{-2.29} & 42.06\negdiff{-10.20} & 48.50\negdiff{-3.76} \\ 
\texttt{neg} & 82.06 & 83.40\posdiff{1.34} & 81.66\negdiff{-0.40} & 82.67\posdiff{0.61} & \textbf{84.82}\posdiff{2.76} & 81.09\negdiff{-0.97} & 78.43\negdiff{-3.63} \\ 
\texttt{auxpass} & 97.37 & \textbf{97.51}\posdiff{0.14} & 97.18\negdiff{-0.19} & 97.26\negdiff{-0.11} & 95.74\negdiff{-1.63} & 95.66\negdiff{-1.71} & 96.24\negdiff{-1.13} \\ 
\texttt{nsubjpass} & 79.34 & \textbf{83.76}\posdiff{4.42} & 73.26\negdiff{-6.08} & 79.43\posdiff{0.09} & 73.45\negdiff{-5.89} & 77.24\negdiff{-2.10} & 74.59\negdiff{-4.75} \\ 
\texttt{possessive} & 99.23 & 99.23\negdiff{0.00} & 99.23\negdiff{0.00} & 99.20\negdiff{-0.03} & 99.26\posdiff{0.03} & 99.20\negdiff{-0.03} & \textbf{99.33}\posdiff{0.10} \\ 
\texttt{pcomp} & 87.55 & \textbf{90.06}\posdiff{2.51} & 83.03\negdiff{-4.52} & 86.95\negdiff{-0.60} & 85.11\negdiff{-2.44} & 84.57\negdiff{-2.98} & 80.15\negdiff{-7.40} \\ 
\texttt{discourse} & \textbf{74.04} & 73.80\negdiff{-0.24} & 50.24\negdiff{-23.80} & 62.18\negdiff{-11.86} & 56.72\negdiff{-17.32} & 73.14\negdiff{-0.90} & 52.05\negdiff{-21.99} \\ 
\texttt{partmod} & 60.08 & 59.47\negdiff{-0.61} & 59.52\negdiff{-0.56} & 60.49\posdiff{0.41} & 63.14\posdiff{3.06} & \textbf{64.85}\posdiff{4.77} & 62.58\posdiff{2.50} \\ 
\texttt{appos} & 54.16 & 47.88\negdiff{-6.28} & 41.85\negdiff{-12.31} & 47.30\negdiff{-6.86} & \textbf{54.65}\posdiff{0.49} & 50.12\negdiff{-4.04} & 43.55\negdiff{-10.61} \\ 
\texttt{prt} & 96.44 & 95.99\negdiff{-0.45} & 96.39\negdiff{-0.05} & 96.69\posdiff{0.25} & \textbf{96.83}\posdiff{0.39} & 95.30\negdiff{-1.14} & 96.09\negdiff{-0.35} \\ 
\texttt{number} & 80.75 & 78.43\negdiff{-2.32} & 77.71\negdiff{-3.04} & \textbf{82.86}\posdiff{2.11} & 81.12\posdiff{0.37} & 81.41\posdiff{0.66} & 82.57\posdiff{1.82} \\ 
\texttt{quantmod} & 75.49 & 73.06\negdiff{-2.43} & 72.65\negdiff{-2.84} & 78.32\posdiff{2.83} & \textbf{80.18}\posdiff{4.69} & 73.06\negdiff{-2.43} & 75.41\negdiff{-0.08} \\ 
\texttt{parataxis} & 46.48 & 43.92\negdiff{-2.56} & 30.69\negdiff{-15.79} & 45.24\negdiff{-1.24} & 43.30\negdiff{-3.18} & \textbf{47.79}\posdiff{1.31} & 33.69\negdiff{-12.79} \\ 
\texttt{infmod} & 71.38 & 71.20\negdiff{-0.18} & 68.35\negdiff{-3.03} & 70.13\negdiff{-1.25} & 70.49\negdiff{-0.89} & \textbf{73.24}\posdiff{1.86} & 68.18\negdiff{-3.20} \\ 
\texttt{tmod} & 84.89 & \textbf{87.64}\posdiff{2.75} & 69.23\negdiff{-15.66} & 78.02\negdiff{-6.87} & 76.01\negdiff{-8.88} & 80.31\negdiff{-4.58} & 62.64\negdiff{-22.25} \\ 
\texttt{expl} & 85.73 & 86.27\posdiff{0.54} & \textbf{86.93}\posdiff{1.20} & 86.82\posdiff{1.09} & 84.75\negdiff{-0.98} & 84.75\negdiff{-0.98} & 84.10\negdiff{-1.63} \\ 
\texttt{mwe} & \textbf{100.00} & \textbf{100.00}\negdiff{0.00} & \textbf{100.00}\negdiff{0.00} & \textbf{100.00}\negdiff{0.00} & \textbf{100.00}\negdiff{0.00} & \textbf{100.00}\negdiff{0.00} & \textbf{100.00}\negdiff{0.00} \\ 
\texttt{npadvmod} & 86.24 & 87.16\posdiff{0.92} & \textbf{88.07}\posdiff{1.83} & 82.42\negdiff{-3.82} & 84.25\negdiff{-1.99} & 83.03\negdiff{-3.21} & 81.19\negdiff{-5.05} \\ 
\texttt{iobj} & 93.30 & \textbf{94.02}\posdiff{0.72} & 91.12\negdiff{-2.18} & 89.67\negdiff{-3.63} & 91.85\negdiff{-1.45} & 93.66\posdiff{0.36} & 88.95\negdiff{-4.35} \\ 
\texttt{predet} & 91.72 & 91.30\negdiff{-0.42} & 91.93\posdiff{0.21} & \textbf{93.17}\posdiff{1.45} & 92.34\posdiff{0.62} & 87.78\negdiff{-3.94} & 90.47\negdiff{-1.25} \\ 
\texttt{acomp} & 89.60 & 89.17\negdiff{-0.43} & 88.33\negdiff{-1.27} & 89.60\negdiff{0.00} & \textbf{89.81}\posdiff{0.21} & 87.05\negdiff{-2.55} & 89.17\negdiff{-0.43} \\ 
\texttt{csubj} & 56.90 & 57.76\posdiff{0.86} & 57.19\posdiff{0.29} & \textbf{59.48}\posdiff{2.58} & 53.45\negdiff{-3.45} & 51.72\negdiff{-5.18} & 55.17\negdiff{-1.73} \\ 
\texttt{preconj} & 76.81 & 78.26\posdiff{1.45} & 65.22\negdiff{-11.59} & 72.46\negdiff{-4.35} & 82.61\posdiff{5.80} & \textbf{83.33}\posdiff{6.52} & 75.36\negdiff{-1.45} \\ 
\texttt{csubjpass} & \textbf{100.00} & \textbf{100.00}\negdiff{0.00} & \textbf{100.00}\negdiff{0.00} & \textbf{100.00}\negdiff{0.00} & \textbf{100.00}\negdiff{0.00} & 77.78\negdiff{-22.22} & 66.67\negdiff{-33.33} \\ 
\end{tabular}
}
\caption{Probing results for DEP. STL: the basline, BERT:pre-trained BERT before fine-tuning.}
\label{tbl:prob-dep}
\end{table*}


\begin{table*}[htbp]
\centering\resizebox{\textwidth}{!}{
\begin{tabular}{l||l|l|l|l|l|l|l}
      & \textbf{STL}   & \textbf{BERT}           & \textbf{POS}           & \textbf{NER}           & \textbf{DEP}           & \textbf{SRL}           & \textbf{MTL-5}         \\
      \hline\hline
\texttt{NP} & \textbf{85.72} & 55.97\negdiff{-29.75} & 65.53\negdiff{-20.19} & 72.55\negdiff{-13.17} & 67.25\negdiff{-18.47} & 71.80\negdiff{-13.92} & 57.58\negdiff{-28.14} \\ 
\texttt{ADVP} & 68.61 & 68.61\negdiff{0.00} & 70.25\posdiff{1.64} & \textbf{77.86}\posdiff{9.25} & 70.47\posdiff{1.86} & 61.15\negdiff{-7.46} & 66.59\negdiff{-2.02} \\ 
\texttt{ADJP} & 67.83 & 53.25\negdiff{-14.58} & \textbf{69.33}\posdiff{1.50} & 65.40\negdiff{-2.43} & 57.69\negdiff{-10.14} & 64.42\negdiff{-3.41} & 63.43\negdiff{-4.40} \\ 
\texttt{VP} & \textbf{79.29} & 62.83\negdiff{-16.46} & 66.39\negdiff{-12.90} & 70.30\negdiff{-8.99} & 78.07\negdiff{-1.22} & 54.85\negdiff{-24.44} & 69.70\negdiff{-9.59} \\ 
\texttt{NML} & 82.92 & 74.89\negdiff{-8.03} & 81.51\negdiff{-1.41} & 82.17\negdiff{-0.75} & 81.66\negdiff{-1.26} & \textbf{84.24}\posdiff{1.32} & 76.02\negdiff{-6.90} \\ 
\texttt{WHNP} & 77.87 & 62.26\negdiff{-15.61} & 75.12\negdiff{-2.75} & 73.27\negdiff{-4.60} & \textbf{78.82}\posdiff{0.95} & 70.63\negdiff{-7.24} & 71.14\negdiff{-6.73} \\ 
\texttt{INTJ} & 76.61 & 72.67\negdiff{-3.94} & \textbf{84.84}\posdiff{8.23} & 78.11\posdiff{1.50} & 68.79\negdiff{-7.82} & 73.13\negdiff{-3.48} & 74.11\negdiff{-2.50} \\ 
\texttt{QP} & \textbf{94.55} & 90.45\negdiff{-4.10} & 93.49\negdiff{-1.06} & 93.73\negdiff{-0.82} & 91.70\negdiff{-2.85} & 91.65\negdiff{-2.90} & 88.28\negdiff{-6.27} \\ 
\texttt{WHADVP} & 78.99 & 69.43\negdiff{-9.56} & 80.16\posdiff{1.17} & 75.84\negdiff{-3.15} & \textbf{84.04}\posdiff{5.05} & 72.63\negdiff{-6.36} & 76.95\negdiff{-2.04} \\ 
\texttt{PRT} & 80.85 & 79.46\negdiff{-1.39} & 80.90\posdiff{0.05} & 81.20\posdiff{0.35} & \textbf{83.63}\posdiff{2.78} & 73.71\negdiff{-7.14} & 76.34\negdiff{-4.51} \\ 
\texttt{PP} & 73.52 & 69.86\negdiff{-3.66} & \textbf{79.00}\posdiff{5.48} & 78.31\posdiff{4.79} & 72.83\negdiff{-0.69} & 69.86\negdiff{-3.66} & 71.00\negdiff{-2.52} \\ 
\texttt{CONJP} & 90.87 & \textbf{91.78}\posdiff{0.91} & 90.41\negdiff{-0.46} & 85.84\negdiff{-5.03} & 88.58\negdiff{-2.29} & 88.13\negdiff{-2.74} & 87.67\negdiff{-3.20} \\ 
\texttt{X} & 59.35 & \textbf{60.98}\posdiff{1.63} & 52.85\negdiff{-6.50} & 53.66\negdiff{-5.69} & 57.73\negdiff{-1.62} & 55.29\negdiff{-4.06} & 56.91\negdiff{-2.44} \\ 
\texttt{WHADJP} & 95.37 & 91.67\negdiff{-3.70} & 93.52\negdiff{-1.85} & \textbf{96.29}\posdiff{0.92} & 94.44\negdiff{-0.93} & 94.44\negdiff{-0.93} & 91.67\negdiff{-3.70} \\ 
\texttt{META} & \textbf{98.61} & 95.83\negdiff{-2.78} & 97.22\negdiff{-1.39} & 95.83\negdiff{-2.78} & 97.22\negdiff{-1.39} & 97.22\negdiff{-1.39} & \textbf{98.61}\negdiff{0.00} \\ 
\texttt{UCP} & \textbf{100.00} & \textbf{100.00}\negdiff{0.00} & 98.25\negdiff{-1.75} & \textbf{100.00}\negdiff{0.00} & 98.25\negdiff{-1.75} & \textbf{100.00}\negdiff{0.00} & \textbf{100.00}\negdiff{0.00} \\ 
\texttt{S} & 96.29 & 94.44\negdiff{-1.85} & 96.29\negdiff{0.00} & 94.44\negdiff{-1.85} & 96.29\negdiff{0.00} & \textbf{98.15}\posdiff{1.86} & 94.44\negdiff{-1.85} \\ 
\texttt{LST} & \textbf{100.00} & \textbf{100.00}\negdiff{0.00} & \textbf{100.00}\negdiff{0.00} & \textbf{100.00}\negdiff{0.00} & \textbf{100.00}\negdiff{0.00} & \textbf{100.00}\negdiff{0.00} & \textbf{100.00}\negdiff{0.00} \\ 
\texttt{FRAG} & 83.33 & \textbf{100.00}\posdiff{16.67} & \textbf{100.00}\posdiff{16.67} & 94.44\posdiff{11.11} & 94.44\posdiff{11.11} & 88.89\posdiff{5.56} & \textbf{100.00}\posdiff{16.67} \\ 
\texttt{SBAR} & \textbf{100.00} & \textbf{100.00}\negdiff{0.00} & \textbf{100.00}\negdiff{0.00} & \textbf{100.00}\negdiff{0.00} & \textbf{100.00}\negdiff{0.00} & \textbf{100.00}\negdiff{0.00} & \textbf{100.00}\negdiff{0.00} \\ 
\texttt{SQ} & \textbf{100.00} & \textbf{100.00}\negdiff{0.00} & \textbf{100.00}\negdiff{0.00} & \textbf{100.00}\negdiff{0.00} & \textbf{100.00}\negdiff{0.00} & \textbf{100.00}\negdiff{0.00} & \textbf{100.00}\negdiff{0.00} \\ 
\texttt{WHPP} & \textbf{100.00} & \textbf{100.00}\negdiff{0.00} & \textbf{100.00}\negdiff{0.00} & \textbf{100.00}\negdiff{0.00} & \textbf{100.00}\negdiff{0.00} & \textbf{100.00}\negdiff{0.00} & \textbf{100.00}\negdiff{0.00} \\ 
\texttt{TOP} & \textbf{100.00} & \textbf{100.00}\negdiff{0.00} & \textbf{100.00}\negdiff{0.00} & \textbf{100.00}\negdiff{0.00} & \textbf{100.00}\negdiff{0.00} & \textbf{100.00}\negdiff{0.00} & \textbf{100.00}\negdiff{0.00} \\ 
\end{tabular}
}
\caption{Probing results for CON. STL: the basline, BERT:pre-trained BERT before fine-tuning.}
\label{tbl:prob-con}
\end{table*}


\begin{table*}[htbp]
\centering\resizebox{\textwidth}{!}{
\begin{tabular}{l||l|l|l|l|l|l|l}
      & \textbf{STL}   & \textbf{BERT}           & \textbf{POS}           & \textbf{NER}           & \textbf{DEP}           & \textbf{CON}           & \textbf{MTL-5}         \\
      \hline\hline
\texttt{ARG1} & 77.75 & 71.88\negdiff{-5.87} & 78.78\posdiff{1.03} & 77.86\posdiff{0.11} & 77.08\negdiff{-0.67} & \textbf{79.58}\posdiff{1.83} & 74.58\negdiff{-3.17} \\ 
\texttt{ARG0} & 73.24 & 74.46\posdiff{1.22} & \textbf{75.95}\posdiff{2.71} & 71.61\negdiff{-1.63} & 71.17\negdiff{-2.07} & 74.16\posdiff{0.92} & 70.13\negdiff{-3.11} \\ 
\texttt{ARG2} & 89.69 & 80.43\negdiff{-9.26} & 89.81\posdiff{0.12} & 89.65\negdiff{-0.04} & \textbf{90.15}\posdiff{0.46} & 89.86\posdiff{0.17} & 87.36\negdiff{-2.33} \\ 
\texttt{ARGM-TMP} & 73.24 & 60.55\negdiff{-12.69} & 74.48\posdiff{1.24} & 70.57\negdiff{-2.67} & \textbf{76.58}\posdiff{3.34} & 75.59\posdiff{2.35} & 68.55\negdiff{-4.69} \\ 
\texttt{ARGM-DIS} & 70.40 & 50.18\negdiff{-20.22} & 69.56\negdiff{-0.84} & 75.67\posdiff{5.27} & 69.58\negdiff{-0.82} & \textbf{77.51}\posdiff{7.11} & 60.41\negdiff{-9.99} \\ 
\texttt{ARGM-ADV} & 59.01 & 40.28\negdiff{-18.73} & 61.31\posdiff{2.30} & \textbf{64.83}\posdiff{5.82} & 56.48\negdiff{-2.53} & 63.88\posdiff{4.87} & 54.36\negdiff{-4.65} \\ 
\texttt{ARGM-MOD} & 82.53 & 80.46\negdiff{-2.07} & 81.77\negdiff{-0.76} & \textbf{84.41}\posdiff{1.88} & 82.29\negdiff{-0.24} & 81.59\negdiff{-0.94} & 78.54\negdiff{-3.99} \\ 
\texttt{ARGM-LOC} & 79.45 & 58.99\negdiff{-20.46} & 78.24\negdiff{-1.21} & \textbf{80.68}\posdiff{1.23} & 72.88\negdiff{-6.57} & 79.35\negdiff{-0.10} & 74.22\negdiff{-5.23} \\ 
\texttt{ARGM-MNR} & 80.67 & 69.91\negdiff{-10.76} & 79.33\negdiff{-1.34} & 80.52\negdiff{-0.15} & 80.20\negdiff{-0.47} & \textbf{80.89}\posdiff{0.22} & 77.23\negdiff{-3.44} \\ 
\texttt{ARGM-NEG} & 86.47 & 79.39\negdiff{-7.08} & 85.51\negdiff{-0.96} & 90.54\posdiff{4.07} & \textbf{90.81}\posdiff{4.34} & 84.75\negdiff{-1.72} & 88.40\posdiff{1.93} \\ 
\texttt{R-ARG1} & 69.92 & 67.43\negdiff{-2.49} & 64.77\negdiff{-5.15} & 71.28\posdiff{1.36} & 74.75\posdiff{4.83} & 72.09\posdiff{2.17} & \textbf{75.41}\posdiff{5.49} \\ 
\texttt{R-ARG0} & 81.20 & 75.64\negdiff{-5.56} & 76.89\negdiff{-4.31} & 81.71\posdiff{0.51} & \textbf{83.36}\posdiff{2.16} & 80.30\negdiff{-0.90} & 78.36\negdiff{-2.84} \\ 
\texttt{C-ARG1} & \textbf{51.72} & 47.46\negdiff{-4.26} & 51.25\negdiff{-0.47} & 46.86\negdiff{-4.86} & 45.51\negdiff{-6.21} & 45.84\negdiff{-5.88} & 46.11\negdiff{-5.61} \\ 
\texttt{ARGM-PRP} & 80.48 & 69.05\negdiff{-11.43} & 78.81\negdiff{-1.67} & 82.06\posdiff{1.58} & 77.38\negdiff{-3.10} & \textbf{83.49}\posdiff{3.01} & 80.16\negdiff{-0.32} \\ 
\texttt{ARGM-DIR} & 93.89 & 89.29\negdiff{-4.60} & 93.73\negdiff{-0.16} & \textbf{95.16}\posdiff{1.27} & 92.86\negdiff{-1.03} & 94.92\posdiff{1.03} & 94.76\posdiff{0.87} \\ 
\texttt{ARG3} & 91.00 & 82.89\negdiff{-8.11} & \textbf{91.98}\posdiff{0.98} & 90.91\negdiff{-0.09} & 87.34\negdiff{-3.66} & 90.91\negdiff{-0.09} & 89.04\negdiff{-1.96} \\ 
\texttt{ARG4} & 94.78 & 93.96\negdiff{-0.82} & \textbf{95.51}\posdiff{0.73} & 95.05\posdiff{0.27} & 94.32\negdiff{-0.46} & 94.32\negdiff{-0.46} & 93.96\negdiff{-0.82} \\ 
\texttt{ARGM-CAU} & 62.34 & 55.37\negdiff{-6.97} & 63.37\posdiff{1.03} & 66.76\posdiff{4.42} & \textbf{68.74}\posdiff{6.40} & 67.98\posdiff{5.64} & 65.73\posdiff{3.39} \\ 
\texttt{ARGM-PRD} & 57.72 & 52.14\negdiff{-5.58} & \textbf{61.87}\posdiff{4.15} & 59.66\posdiff{1.94} & 61.35\posdiff{3.63} & 59.40\posdiff{1.68} & 57.85\posdiff{0.13} \\ 
\texttt{ARGM-ADJ} & \textbf{85.07} & 76.00\negdiff{-9.07} & 82.53\negdiff{-2.54} & 82.80\negdiff{-2.27} & 82.67\negdiff{-2.40} & 84.40\negdiff{-0.67} & 82.93\negdiff{-2.14} \\ 
\texttt{ARGM-EXT} & 87.37 & 86.93\negdiff{-0.44} & \textbf{90.85}\posdiff{3.48} & 87.80\posdiff{0.43} & 89.32\posdiff{1.95} & 88.89\posdiff{1.52} & 83.44\negdiff{-3.93} \\ 
\texttt{ARGM-PNC} & 82.89 & 69.74\negdiff{-13.15} & \textbf{84.21}\posdiff{1.32} & \textbf{84.21}\posdiff{1.32} & 82.89\negdiff{0.00} & 83.77\posdiff{0.88} & 79.82\negdiff{-3.07} \\ 
\texttt{ARGM-GOL} & 86.76 & 76.71\negdiff{-10.05} & \textbf{89.95}\posdiff{3.19} & 88.58\posdiff{1.82} & 78.54\negdiff{-8.22} & 85.84\negdiff{-0.92} & 84.93\negdiff{-1.83} \\ 
\texttt{ARGM-LVB} & \textbf{98.59} & 97.18\negdiff{-1.41} & 98.12\negdiff{-0.47} & 98.12\negdiff{-0.47} & 98.12\negdiff{-0.47} & 97.18\negdiff{-1.41} & 97.18\negdiff{-1.41} \\ 
\texttt{R-ARGM-LOC} & 82.05 & 84.62\posdiff{2.57} & 78.97\negdiff{-3.08} & \textbf{86.16}\posdiff{4.11} & 74.36\negdiff{-7.69} & 73.85\negdiff{-8.20} & 79.49\negdiff{-2.56} \\ 
\texttt{R-ARGM-TMP} & 74.08 & 71.43\negdiff{-2.65} & 69.84\negdiff{-4.24} & \textbf{82.01}\posdiff{7.93} & 80.42\posdiff{6.34} & 75.66\posdiff{1.58} & 77.25\posdiff{3.17} \\ 
\texttt{R-ARG2} & 73.12 & 70.97\negdiff{-2.15} & \textbf{77.96}\posdiff{4.84} & 77.42\posdiff{4.30} & 74.19\posdiff{1.07} & 69.89\negdiff{-3.23} & 74.73\posdiff{1.61} \\ 
\texttt{C-ARG2} & 44.90 & 36.73\negdiff{-8.17} & \textbf{45.58}\posdiff{0.68} & 43.54\negdiff{-1.36} & 43.54\negdiff{-1.36} & 44.22\negdiff{-0.68} & \textbf{45.58}\posdiff{0.68} \\ 
\texttt{C-ARG0} & 59.80 & 47.06\negdiff{-12.74} & \textbf{61.76}\posdiff{1.96} & 56.86\negdiff{-2.94} & 51.96\negdiff{-7.84} & 53.92\negdiff{-5.88} & 54.90\negdiff{-4.90} \\ 
\texttt{ARGM-REC} & 84.31 & 70.59\negdiff{-13.72} & \textbf{85.29}\posdiff{0.98} & 78.43\negdiff{-5.88} & 80.39\negdiff{-3.92} & \textbf{85.29}\posdiff{0.98} & 80.39\negdiff{-3.92} \\ 
\texttt{ARGM-COM} & 88.89 & 81.48\negdiff{-7.41} & 90.12\posdiff{1.23} & \textbf{92.59}\posdiff{3.70} & 87.66\negdiff{-1.23} & 87.66\negdiff{-1.23} & 87.66\negdiff{-1.23} \\ 
\texttt{C-ARGM-ADV} & 54.55 & 54.55\negdiff{0.00} & 48.48\negdiff{-6.07} & 54.55\negdiff{0.00} & 54.55\negdiff{0.00} & 51.52\negdiff{-3.03} & \textbf{57.58}\posdiff{3.03} \\ 
\texttt{R-ARGM-MNR} & 72.73 & 63.64\negdiff{-9.09} & 63.64\negdiff{-9.09} & \textbf{81.82}\posdiff{9.09} & 69.70\negdiff{-3.03} & 78.79\posdiff{6.06} & 66.67\negdiff{-6.06} \\ 
\texttt{ARG5} & \textbf{100.00} & \textbf{100.00}\negdiff{0.00} & \textbf{100.00}\negdiff{0.00} & \textbf{100.00}\negdiff{0.00} & \textbf{100.00}\negdiff{0.00} & \textbf{100.00}\negdiff{0.00} & \textbf{100.00}\negdiff{0.00} \\ 
\texttt{C-ARGM-TMP} & \textbf{71.43} & 42.86\negdiff{-28.57} & 61.90\negdiff{-9.53} & \textbf{71.43}\negdiff{0.00} & 57.14\negdiff{-14.29} & 61.90\negdiff{-9.53} & 66.67\negdiff{-4.76} \\ 
\texttt{R-ARGM-CAU} & 91.67 & 75.00\negdiff{-16.67} & 83.33\negdiff{-8.34} & \textbf{100.00}\posdiff{8.33} & \textbf{100.00}\posdiff{8.33} & 91.67\negdiff{0.00} & 91.67\negdiff{0.00} \\ 
\texttt{C-ARGM-CAU} & 66.67 & 66.67\negdiff{0.00} & 66.67\negdiff{0.00} & \textbf{77.78}\posdiff{11.11} & 66.67\negdiff{0.00} & 66.67\negdiff{0.00} & \textbf{77.78}\posdiff{11.11} \\ 
\texttt{C-ARG3} & \textbf{100.00} & \textbf{100.00}\negdiff{0.00} & 88.89\negdiff{-11.11} & 88.89\negdiff{-11.11} & 88.89\negdiff{-11.11} & 88.89\negdiff{-11.11} & \textbf{100.00}\negdiff{0.00} \\ 
\texttt{R-ARG4} & \textbf{100.00} & \textbf{100.00}\negdiff{0.00} & \textbf{100.00}\negdiff{0.00} & \textbf{100.00}\negdiff{0.00} & \textbf{100.00}\negdiff{0.00} & \textbf{100.00}\negdiff{0.00} & \textbf{100.00}\negdiff{0.00} \\ 
\texttt{C-ARGM-EXT} & \textbf{100.00} & \textbf{100.00}\negdiff{0.00} & \textbf{100.00}\negdiff{0.00} & \textbf{100.00}\negdiff{0.00} & \textbf{100.00}\negdiff{0.00} & \textbf{100.00}\negdiff{0.00} & \textbf{100.00}\negdiff{0.00} \\ 
\texttt{R-ARGM-ADV} & \textbf{100.00} & \textbf{100.00}\negdiff{0.00} & \textbf{100.00}\negdiff{0.00} & \textbf{100.00}\negdiff{0.00} & \textbf{100.00}\negdiff{0.00} & \textbf{100.00}\negdiff{0.00} & \textbf{100.00}\negdiff{0.00} \\ 
\texttt{C-ARGM-LOC} & \textbf{100.00} & \textbf{100.00}\negdiff{0.00} & \textbf{100.00}\negdiff{0.00} & \textbf{100.00}\negdiff{0.00} & \textbf{100.00}\negdiff{0.00} & \textbf{100.00}\negdiff{0.00} & \textbf{100.00}\negdiff{0.00} \\ 
\texttt{C-ARGM-MNR} & \textbf{100.00} & \textbf{100.00}\negdiff{0.00} & \textbf{100.00}\negdiff{0.00} & \textbf{100.00}\negdiff{0.00} & \textbf{100.00}\negdiff{0.00} & \textbf{100.00}\negdiff{0.00} & \textbf{100.00}\negdiff{0.00} \\ 
\texttt{C-ARGM-MOD} & \textbf{100.00} & 50.00\negdiff{-50.00} & \textbf{100.00}\negdiff{0.00} & \textbf{100.00}\negdiff{0.00} & \textbf{100.00}\negdiff{0.00} & \textbf{100.00}\negdiff{0.00} & \textbf{100.00}\negdiff{0.00} \\ 
\texttt{R-ARGM-PRP} & \textbf{100.00} & \textbf{100.00}\negdiff{0.00} & \textbf{100.00}\negdiff{0.00} & \textbf{100.00}\negdiff{0.00} & \textbf{100.00}\negdiff{0.00} & \textbf{100.00}\negdiff{0.00} & \textbf{100.00}\negdiff{0.00} \\ 
\texttt{ARGM-PRX} & \textbf{100.00} & \textbf{100.00}\negdiff{0.00} & \textbf{100.00}\negdiff{0.00} & \textbf{100.00}\negdiff{0.00} & \textbf{100.00}\negdiff{0.00} & \textbf{100.00}\negdiff{0.00} & \textbf{100.00}\negdiff{0.00} \\ 
\texttt{ARGA} & \textbf{100.00} & \textbf{100.00}\negdiff{0.00} & \textbf{100.00}\negdiff{0.00} & \textbf{100.00}\negdiff{0.00} & \textbf{100.00}\negdiff{0.00} & \textbf{100.00}\negdiff{0.00} & \textbf{100.00}\negdiff{0.00} \\ 
\texttt{R-ARGM-DIR} & \textbf{100.00} & \textbf{100.00}\negdiff{0.00} & \textbf{100.00}\negdiff{0.00} & \textbf{100.00}\negdiff{0.00} & \textbf{100.00}\negdiff{0.00} & \textbf{100.00}\negdiff{0.00} & \textbf{100.00}\negdiff{0.00} \\ 
\texttt{R-ARGM-PRD} & \textbf{100.00} & \textbf{100.00}\negdiff{0.00} & \textbf{100.00}\negdiff{0.00} & \textbf{100.00}\negdiff{0.00} & \textbf{100.00}\negdiff{0.00} & \textbf{100.00}\negdiff{0.00} & \textbf{100.00}\negdiff{0.00} \\ 
\texttt{C-ARGM-PRD} & \textbf{100.00} & \textbf{100.00}\negdiff{0.00} & \textbf{100.00}\negdiff{0.00} & \textbf{100.00}\negdiff{0.00} & \textbf{100.00}\negdiff{0.00} & \textbf{100.00}\negdiff{0.00} & \textbf{100.00}\negdiff{0.00} \\ 
\texttt{R-ARGM-EXT} & \textbf{100.00} & \textbf{100.00}\negdiff{0.00} & \textbf{100.00}\negdiff{0.00} & \textbf{100.00}\negdiff{0.00} & \textbf{100.00}\negdiff{0.00} & \textbf{100.00}\negdiff{0.00} & \textbf{100.00}\negdiff{0.00} \\ 
\texttt{C-ARGM-PRP} & \textbf{100.00} & \textbf{100.00}\negdiff{0.00} & \textbf{100.00}\negdiff{0.00} & \textbf{100.00}\negdiff{0.00} & \textbf{100.00}\negdiff{0.00} & \textbf{100.00}\negdiff{0.00} & \textbf{100.00}\negdiff{0.00} \\ 
\texttt{R-ARG3} & \textbf{100.00} & \textbf{100.00}\negdiff{0.00} & \textbf{100.00}\negdiff{0.00} & \textbf{100.00}\negdiff{0.00} & \textbf{100.00}\negdiff{0.00} & \textbf{100.00}\negdiff{0.00} & \textbf{100.00}\negdiff{0.00} \\ 
\texttt{ARGM-PRR} & \textbf{100.00} & \textbf{100.00}\negdiff{0.00} & \textbf{100.00}\negdiff{0.00} & \textbf{100.00}\negdiff{0.00} & \textbf{100.00}\negdiff{0.00} & \textbf{100.00}\negdiff{0.00} & \textbf{100.00}\negdiff{0.00} \\ 
\texttt{ARGM-DSP} & \textbf{100.00} & \textbf{100.00}\negdiff{0.00} & \textbf{100.00}\negdiff{0.00} & \textbf{100.00}\negdiff{0.00} & \textbf{100.00}\negdiff{0.00} & \textbf{100.00}\negdiff{0.00} & \textbf{100.00}\negdiff{0.00} \\
\end{tabular}
}
\caption{Probing results for SRL. STL: the basline, BERT:pre-trained BERT before fine-tuning.}
\label{tbl:prob-srl}
\end{table*}

\clearpage
\subsection{Results for Other Transformers}
\label{ssec:other-transformers}

We also applied our pruning and probing methods on 3 recent transformer encoders, RoBERTa \cite{liu2019roberta}, ELECTRA \cite{clark2020electra} and DeBERTa \cite{he2020deberta}, to further demonstrate the generality of our hypothesis. For all of them, we use the base size version which has 12 layers with 144 attention heads in total.
Although we did not tune hyper-parameters specifically for any of them and re-used the same hyper-parameters of BERT, their results turned out to be as interesting as BERT results.

Their STL and MTL results are shown in Table~\ref{tab:mtl-results-more}. Unsurprisingly, MTL-5 is outperformed by its single-task counterparts for all tasks and for all transformer encoders, raising the dilemma behind transformer-based MTL.

Their pruning results are shown in Table~\ref{tab:pruning-results-more}. Although the results could be better tuned for each transformer encoder, our DP strategy is still able to prune roughly 50\% heads while keeping comparable performance.

\begin{figure*}[ht]
 \centering
 \begin{subfigure}[b]{0.4\columnwidth}
     \centering
     \includegraphics[width=\textwidth]{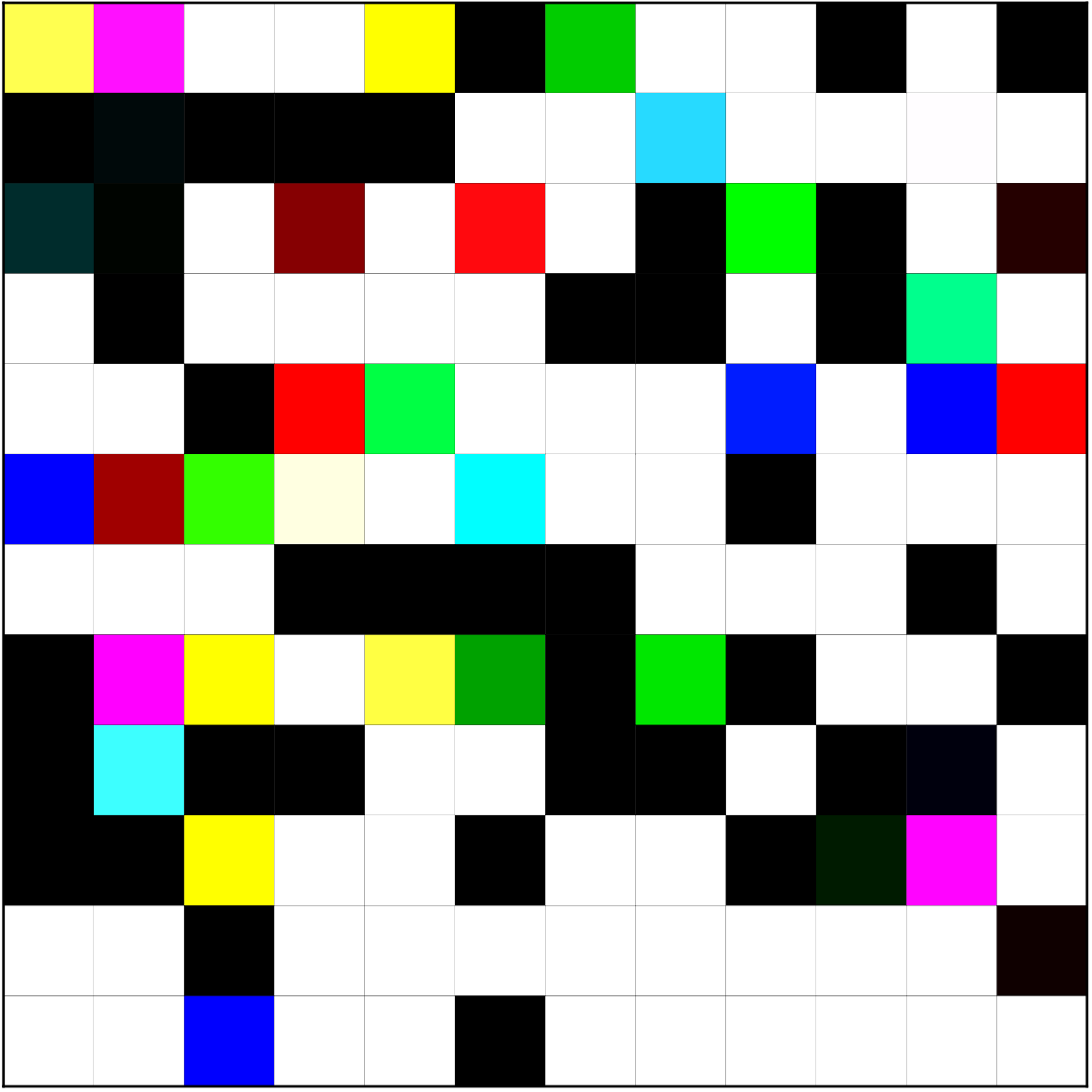}
     \caption{POS}
     \label{fig:prune-pos-roberta}
 \end{subfigure}
 \hfill
 \begin{subfigure}[b]{0.4\columnwidth}
     \centering
     \includegraphics[width=\textwidth]{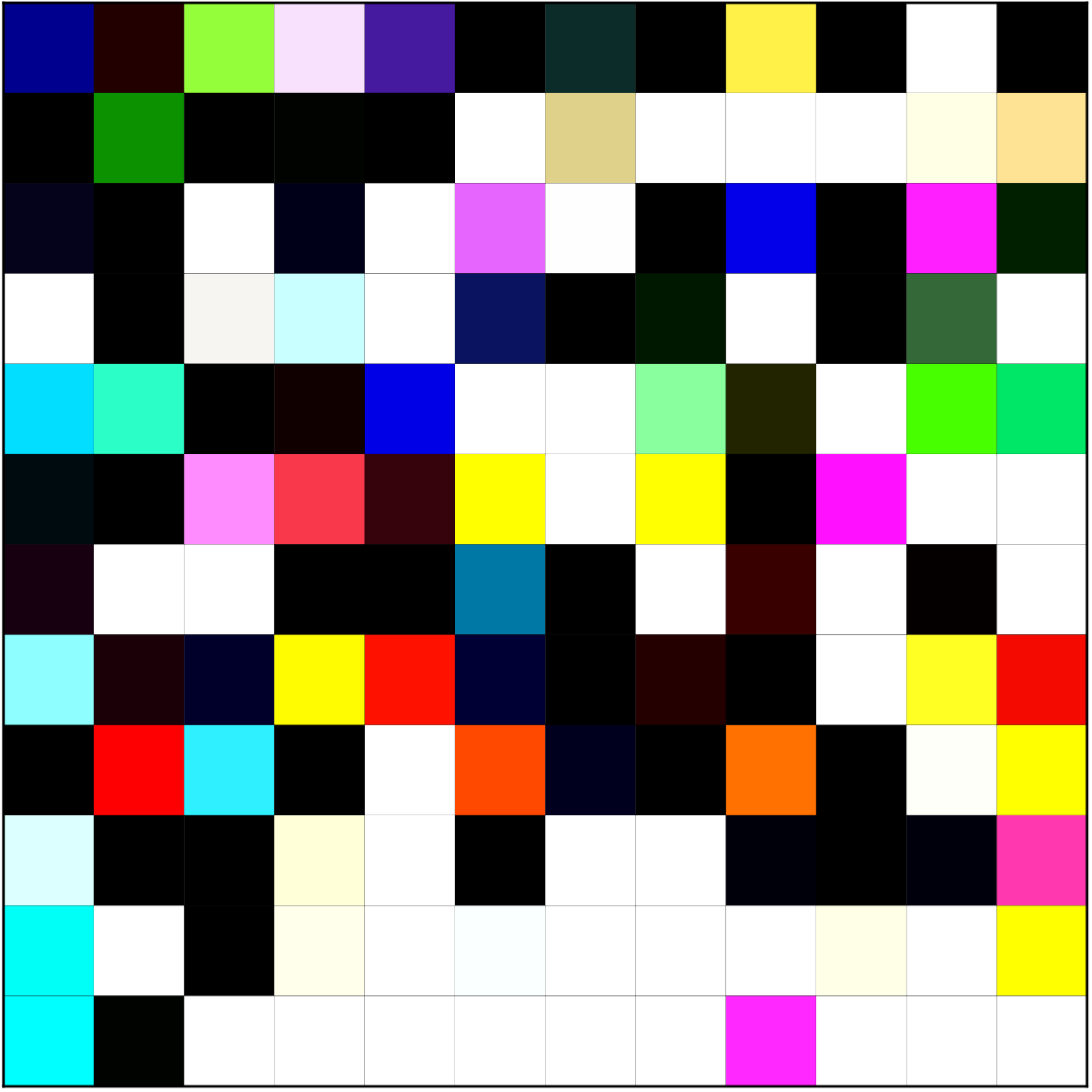}
     \caption{NER}
     \label{fig:prune-ner-roberta}
 \end{subfigure}
 \hfill
 \begin{subfigure}[b]{0.4\columnwidth}
     \centering
     \includegraphics[width=\textwidth]{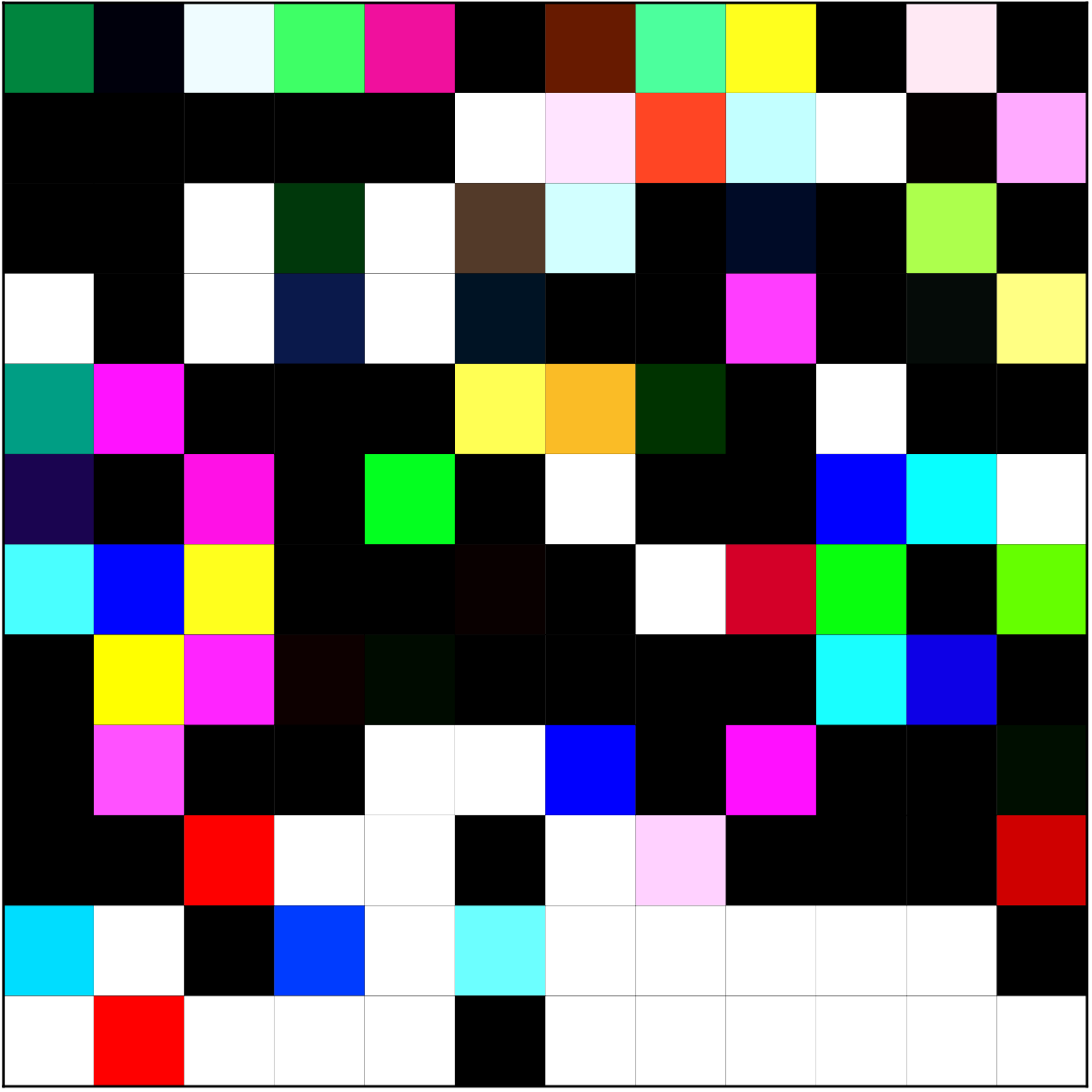}
     \caption{DEP}
     \label{fig:prune-dep-roberta}
 \end{subfigure}
 \hfill
 \begin{subfigure}[b]{0.4\columnwidth}
     \centering
     \includegraphics[width=\textwidth]{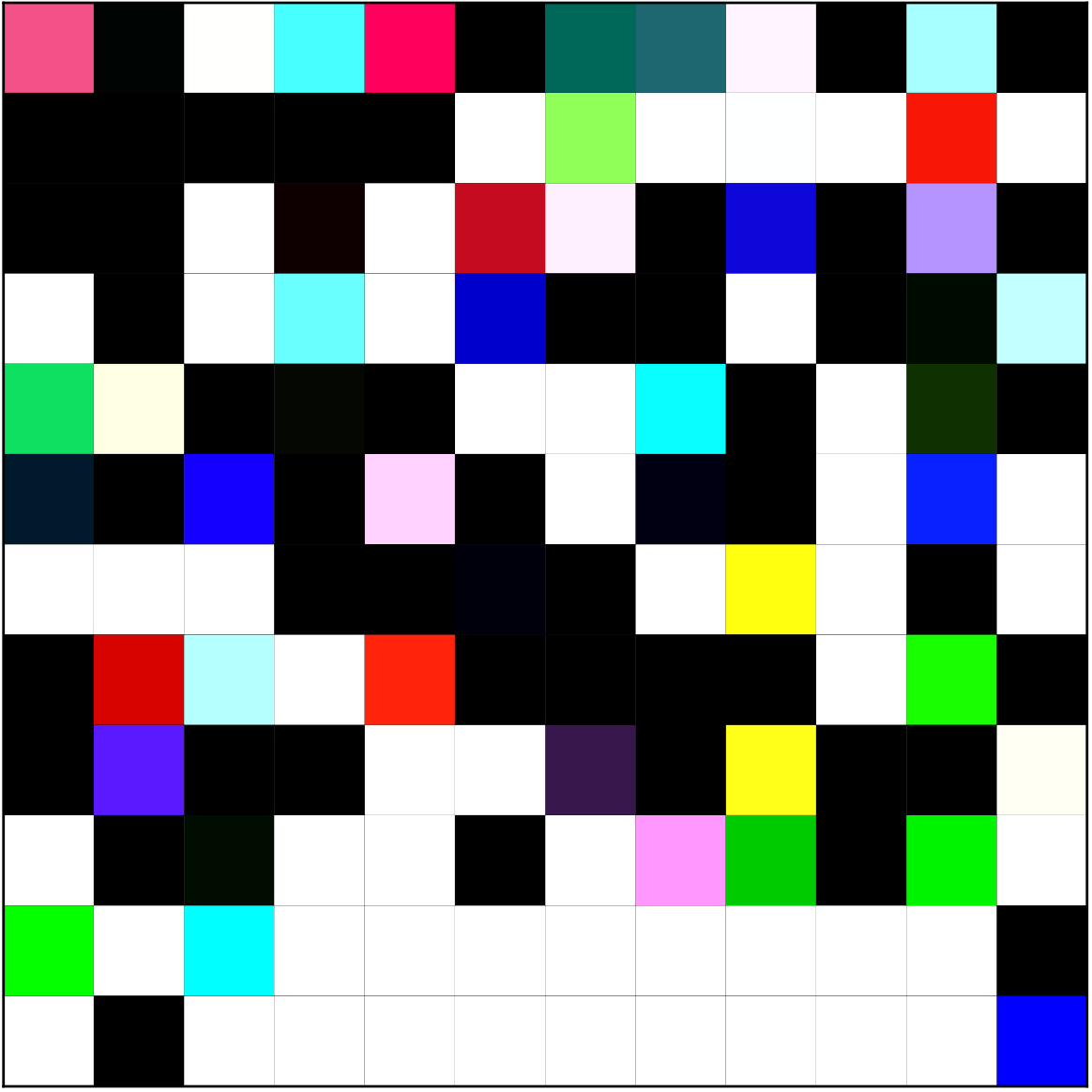}
     \caption{CON}
     \label{fig:prune-con-roberta}
 \end{subfigure}
 \hfill
 \begin{subfigure}[b]{0.4\columnwidth}
     \centering
     \includegraphics[width=\textwidth]{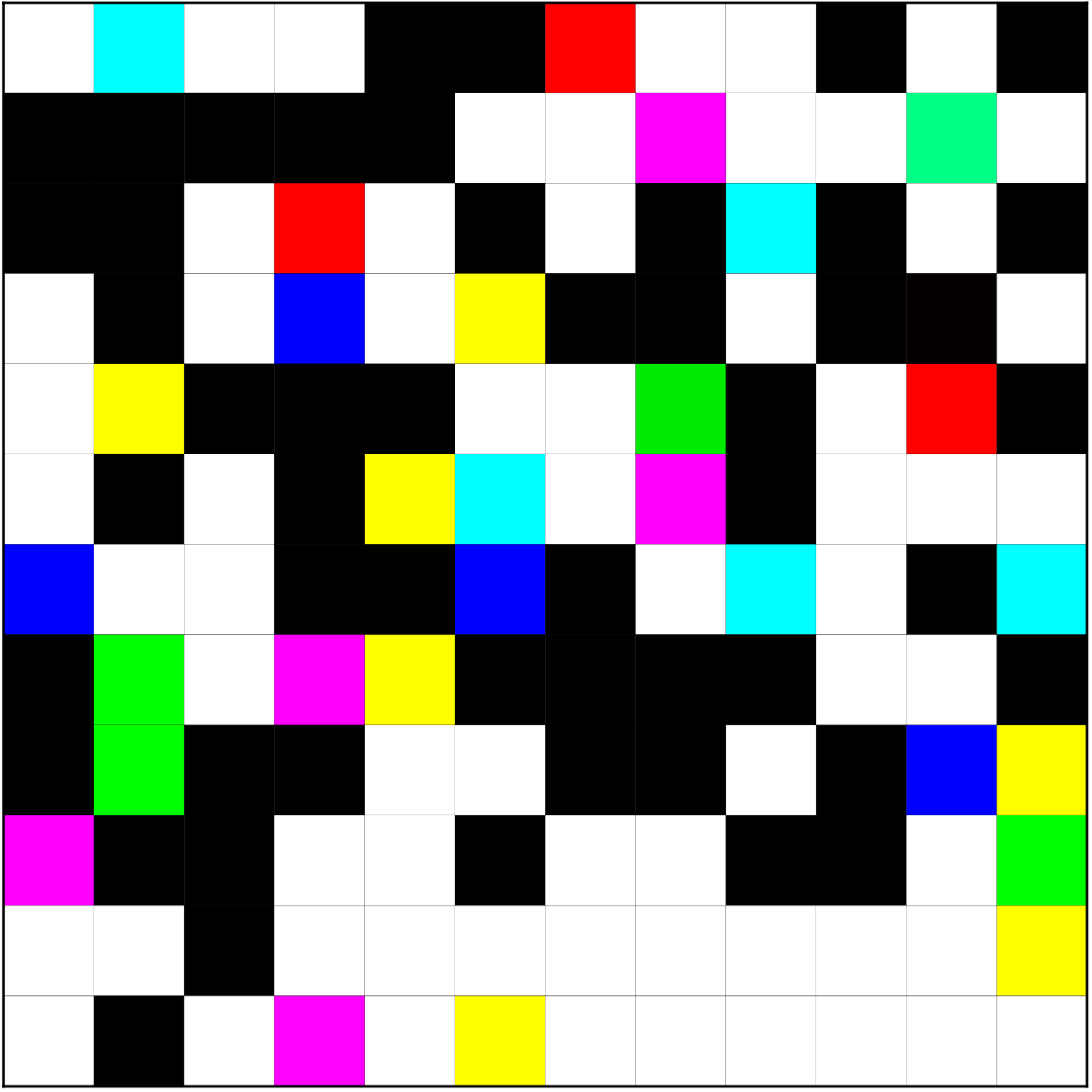}
     \caption{SRL}
     \label{fig:prune-srl-roberta}
 \end{subfigure}

 \begin{subfigure}[b]{\columnwidth}
     \centering
     \includegraphics[width=\textwidth]{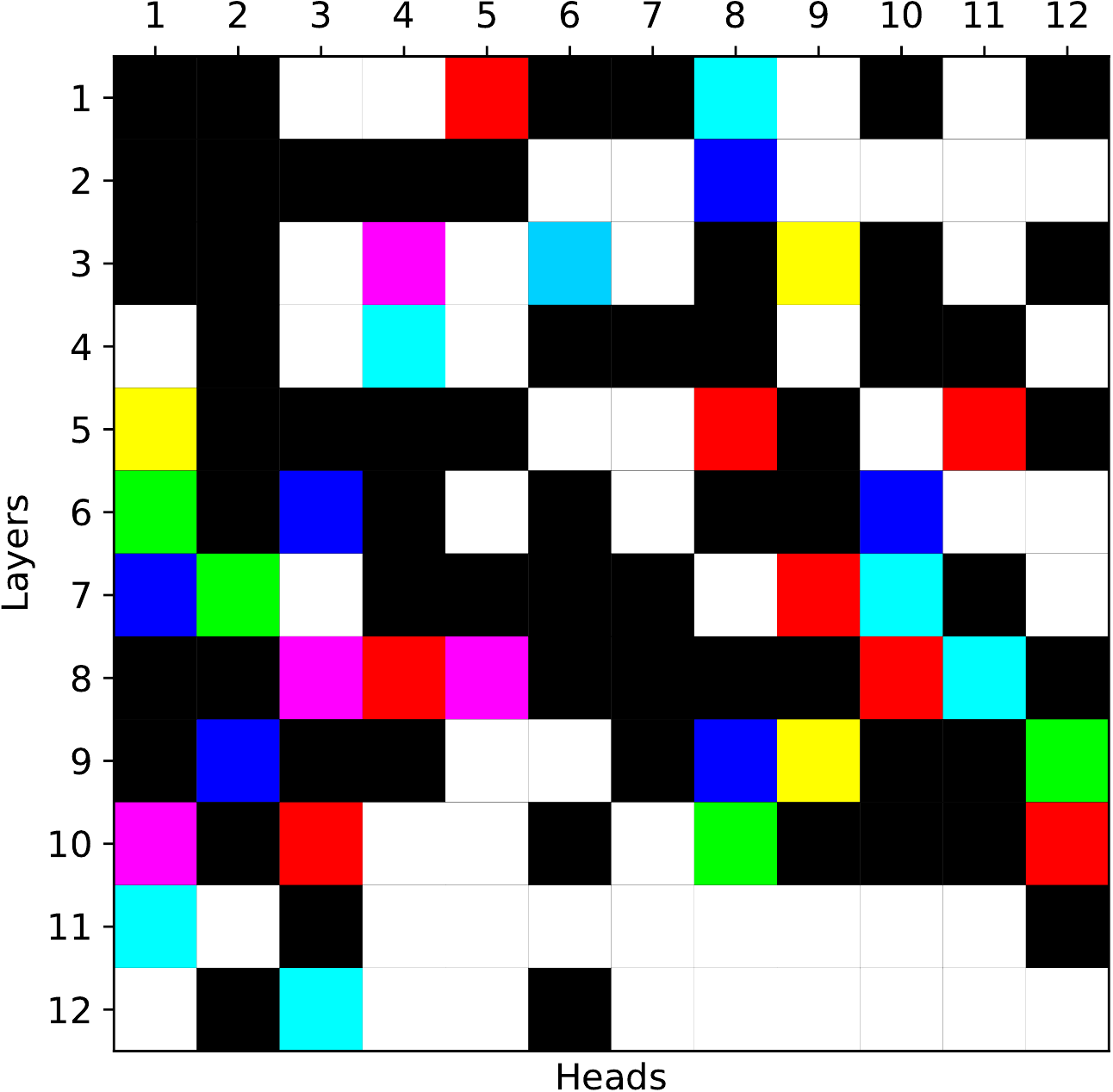}
     \caption{3-run utilization of the MTL-DP model, where each run is encoded in a RGB channel. Darker indicates higher utilization.}
     \label{fig:prune-joint-roberta}
 \end{subfigure}
 \hfill
 \begin{subfigure}[b]{\columnwidth}
     \centering
     \includegraphics[width=\textwidth]{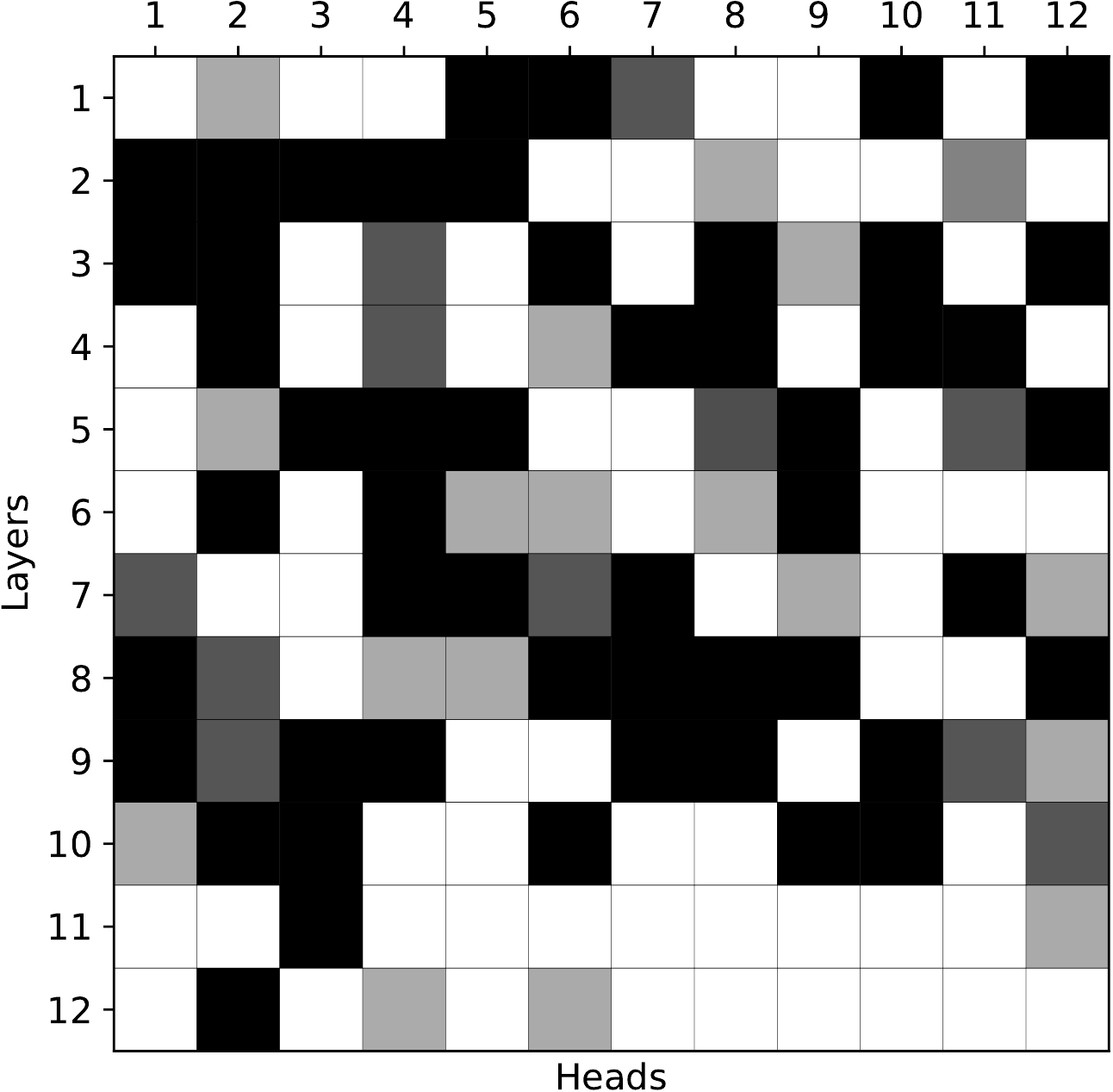}
     \caption{Average head utilization rates among the 5 tasks in 3 runs. Darker cells indicate higher utilization rates.}
     \label{fig:probe-overlay-roberta}
 \end{subfigure}
 \hfill
 
\caption{Head utilization of the RoBERTa \cite{liu2019roberta} STL-DP models (a - e, g) and the MTL-DP model (f).}
\label{fig:overlap-roberta}
\end{figure*}

\begin{figure*}[ht]
 \centering
 \begin{subfigure}[b]{0.4\columnwidth}
     \centering
     \includegraphics[width=\textwidth]{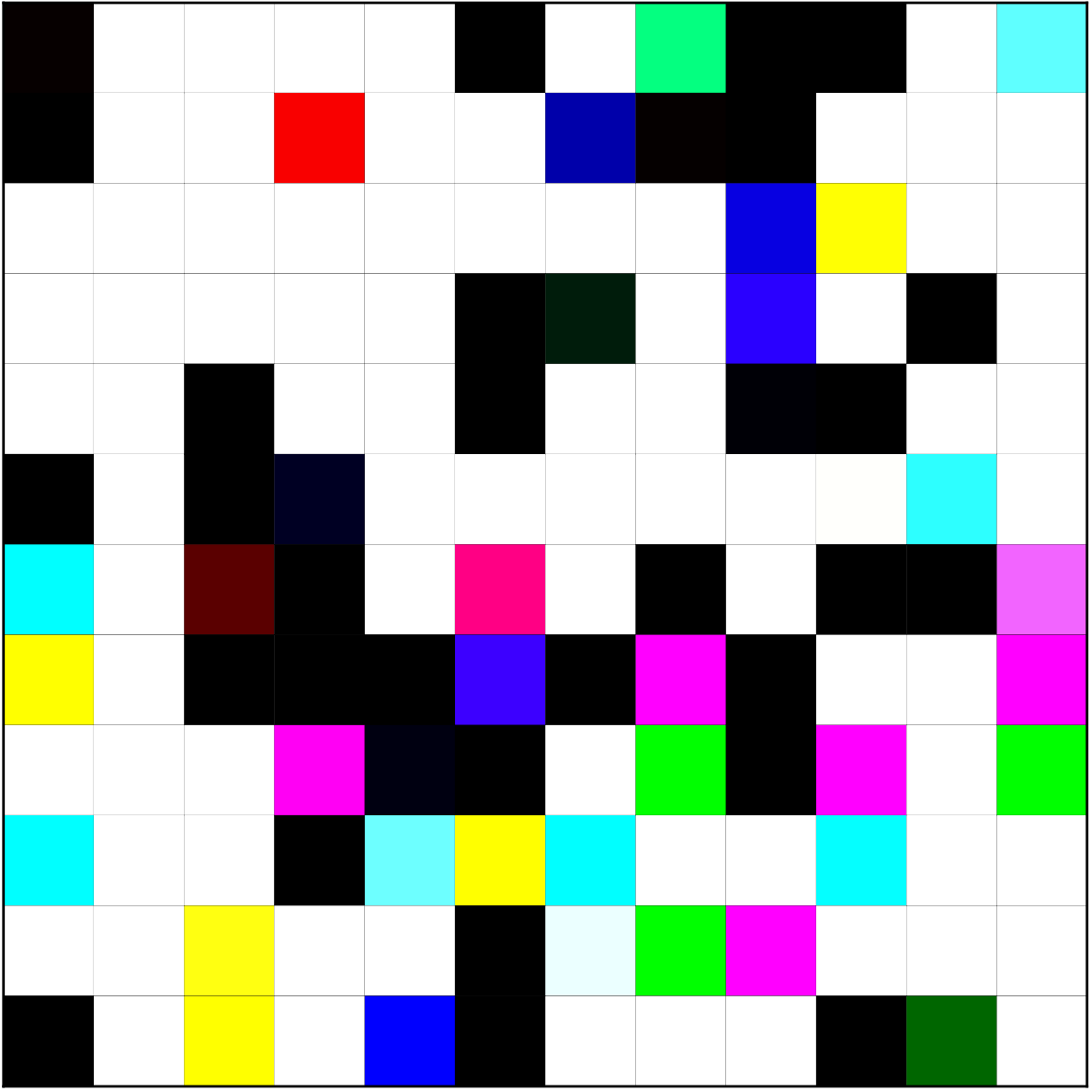}
     \caption{POS}
     \label{fig:prune-pos-electra}
 \end{subfigure}
 \hfill
 \begin{subfigure}[b]{0.4\columnwidth}
     \centering
     \includegraphics[width=\textwidth]{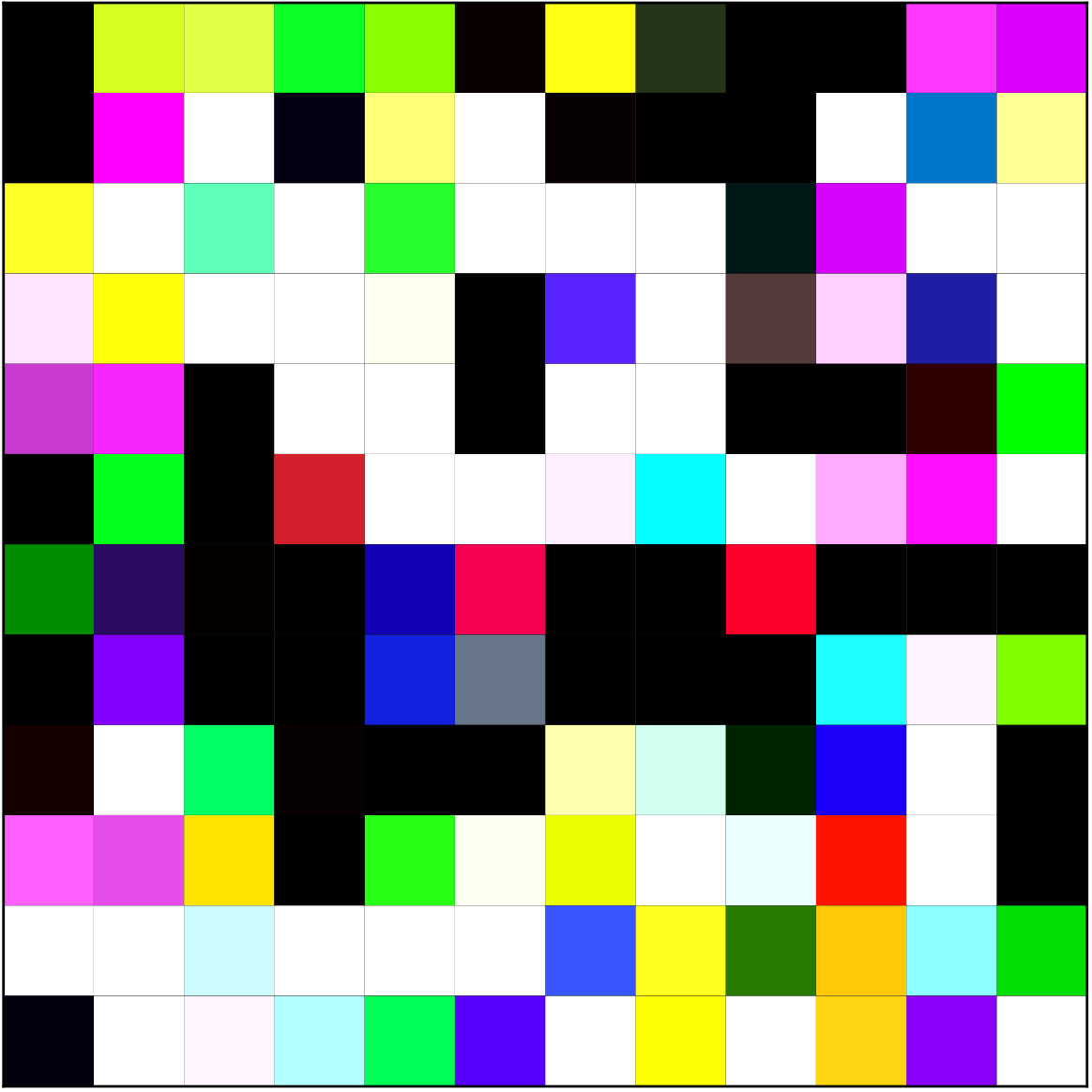}
     \caption{NER}
     \label{fig:prune-ner-electra}
 \end{subfigure}
 \hfill
 \begin{subfigure}[b]{0.4\columnwidth}
     \centering
     \includegraphics[width=\textwidth]{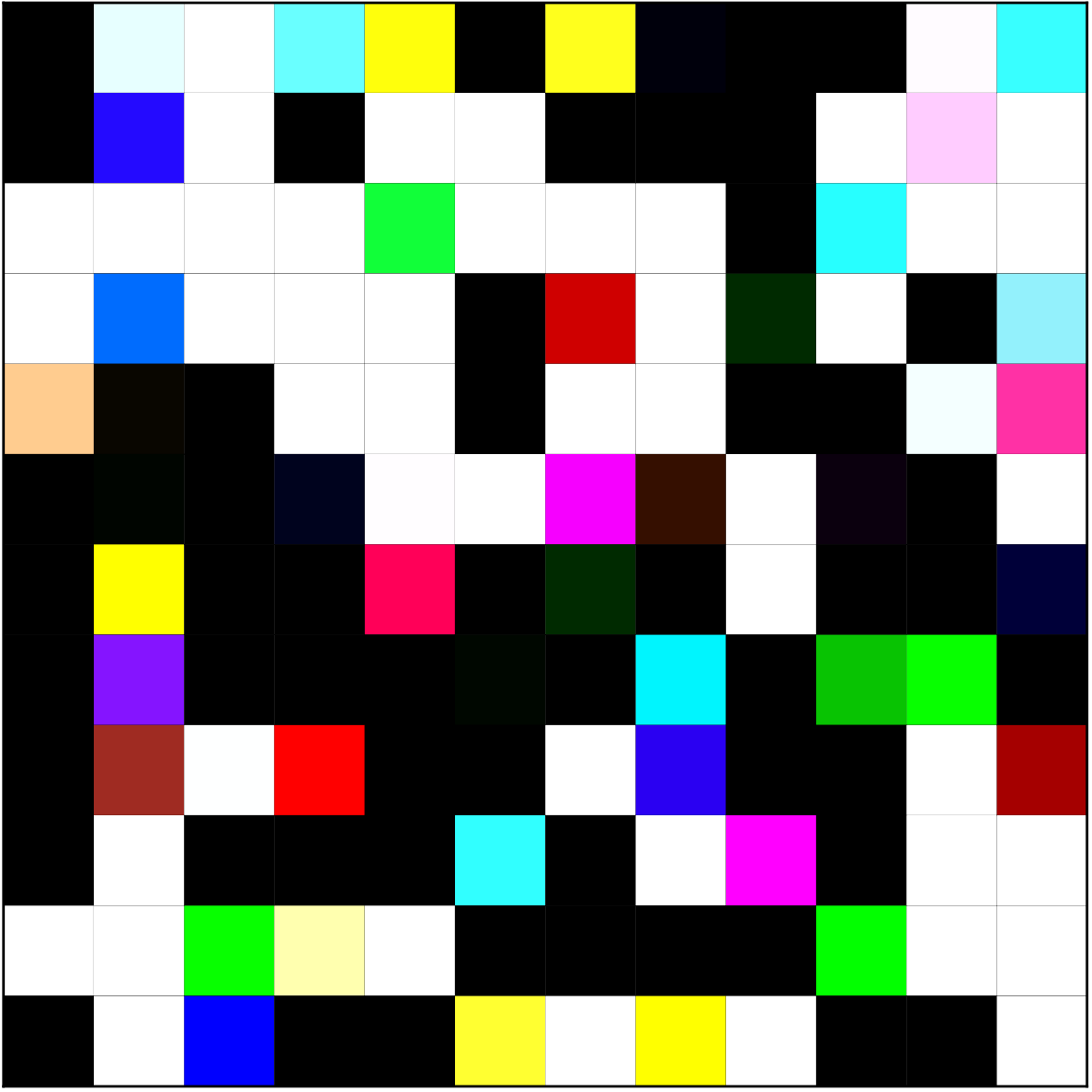}
     \caption{DEP}
     \label{fig:prune-dep-electra}
 \end{subfigure}
 \hfill
 \begin{subfigure}[b]{0.4\columnwidth}
     \centering
     \includegraphics[width=\textwidth]{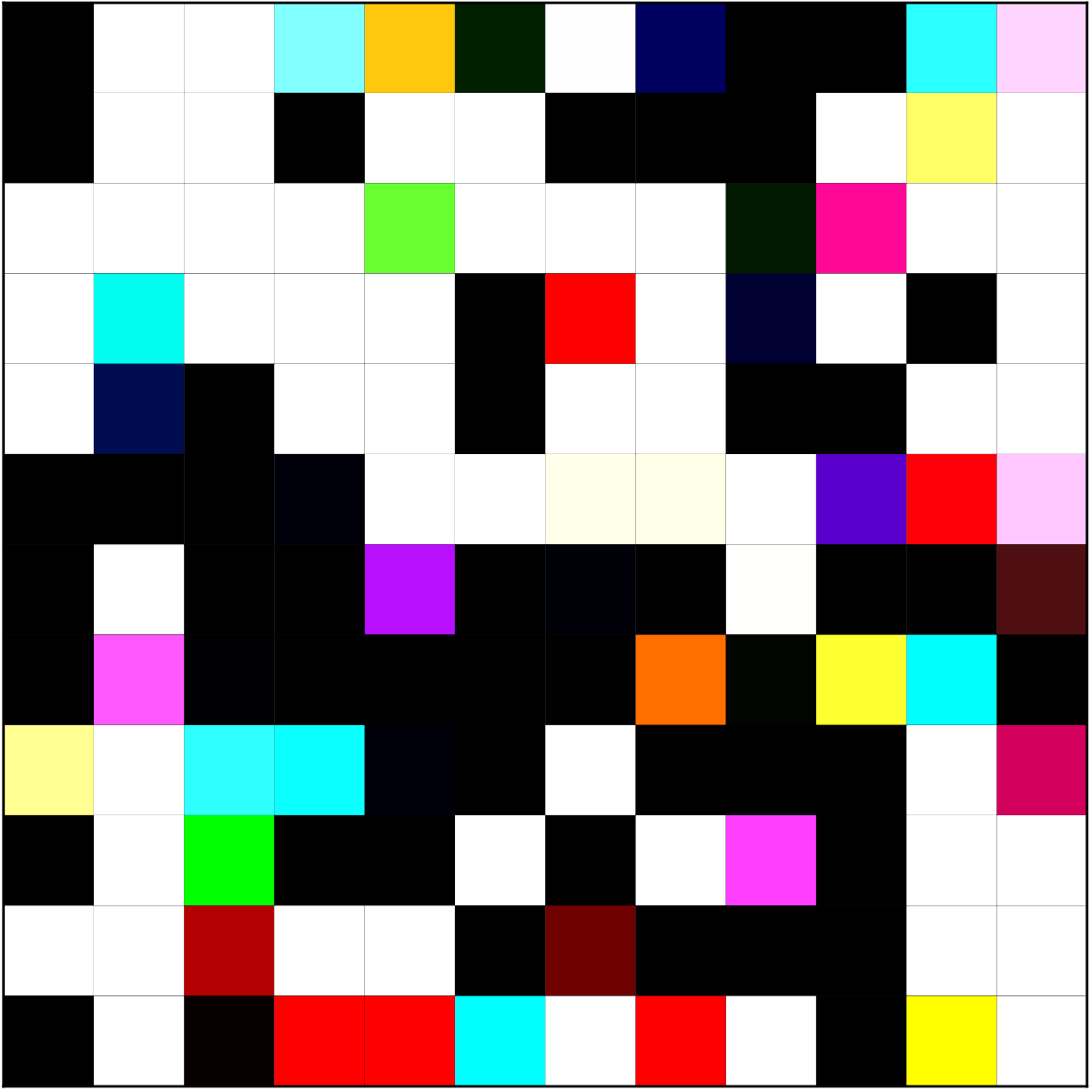}
     \caption{CON}
     \label{fig:prune-con-electra}
 \end{subfigure}
 \hfill
 \begin{subfigure}[b]{0.4\columnwidth}
     \centering
     \includegraphics[width=\textwidth]{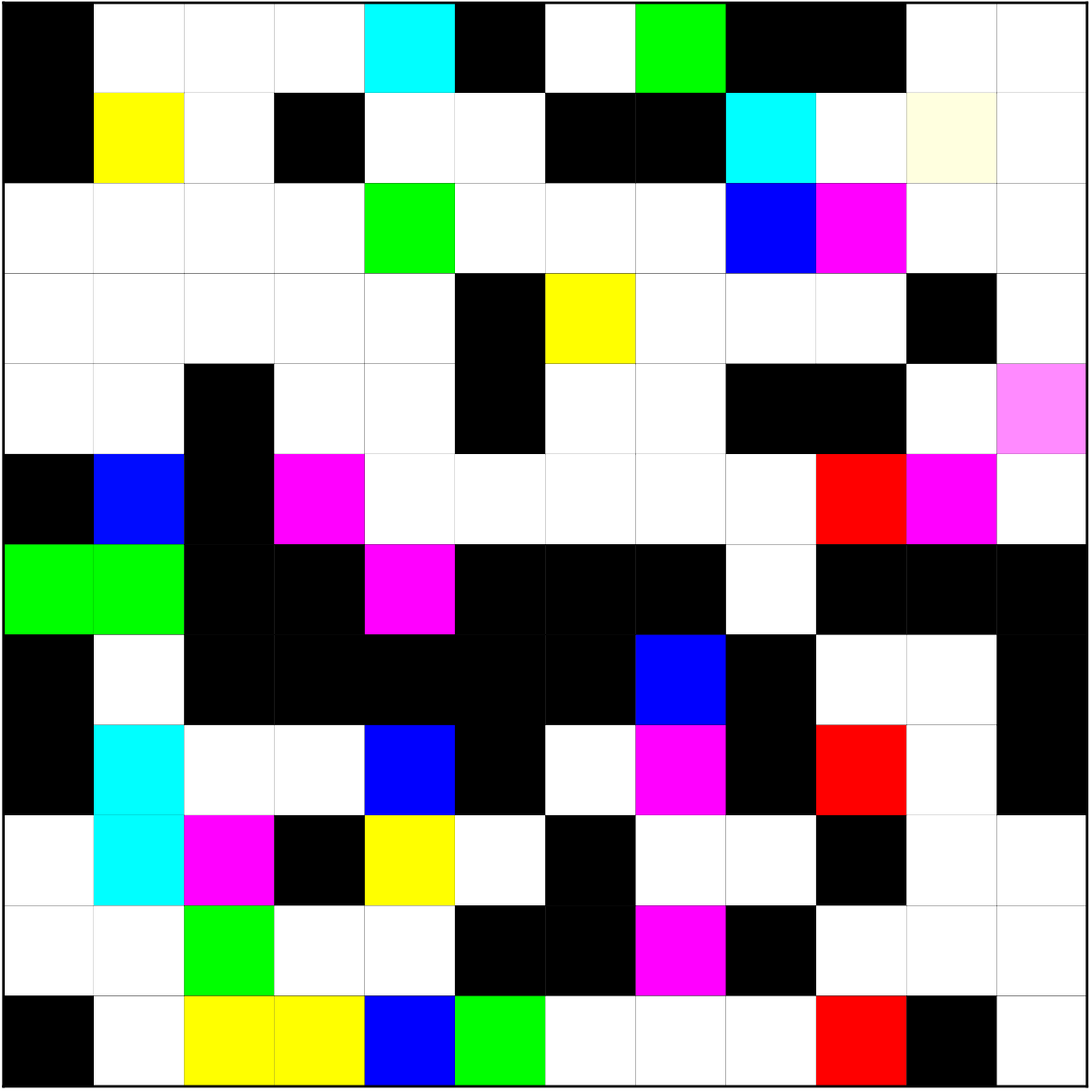}
     \caption{SRL}
     \label{fig:prune-srl-electra}
 \end{subfigure}

 \begin{subfigure}[b]{\columnwidth}
     \centering
     \includegraphics[width=\textwidth]{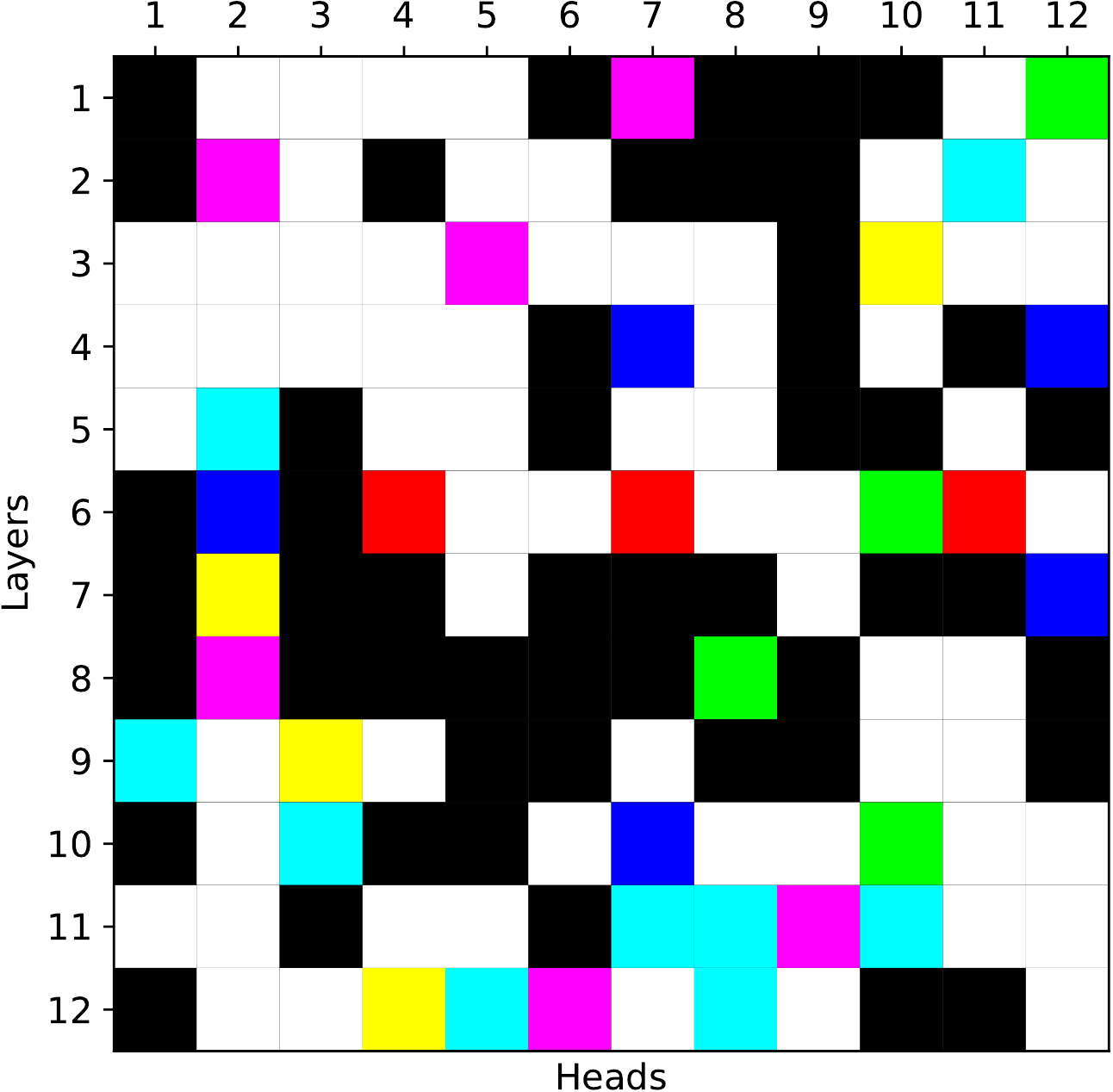}
     \caption{3-run utilization of the MTL-DP model, where each run is encoded in a RGB channel. Darker indicates higher utilization.}
     \label{fig:prune-joint-electra}
 \end{subfigure}
 \hfill
 \begin{subfigure}[b]{\columnwidth}
     \centering
     \includegraphics[width=\textwidth]{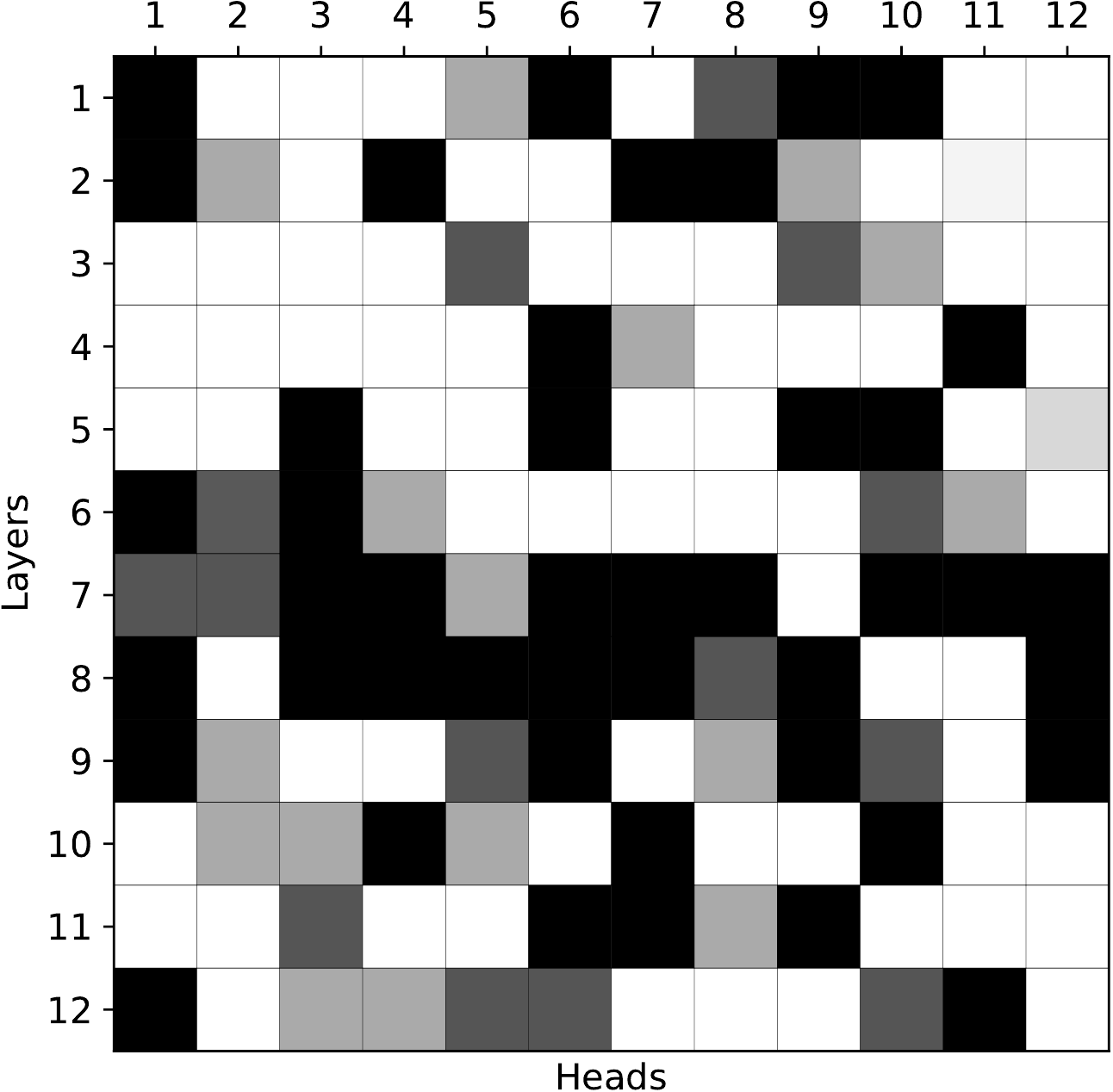}
     \caption{Average head utilization rates among the 5 tasks in 3 runs. Darker cells indicate higher utilization rates.}
     \label{fig:probe-overlay-electra}
 \end{subfigure}
 \hfill
 
\caption{Head utilization of the ELECTRA \cite{clark2020electra} STL-DP models (a - e, g) and the MTL-DP model (f).}
\label{fig:overlap-electra}
\end{figure*}

\begin{figure*}[ht]
 \centering
 \begin{subfigure}[b]{0.4\columnwidth}
     \centering
     \includegraphics[width=\textwidth]{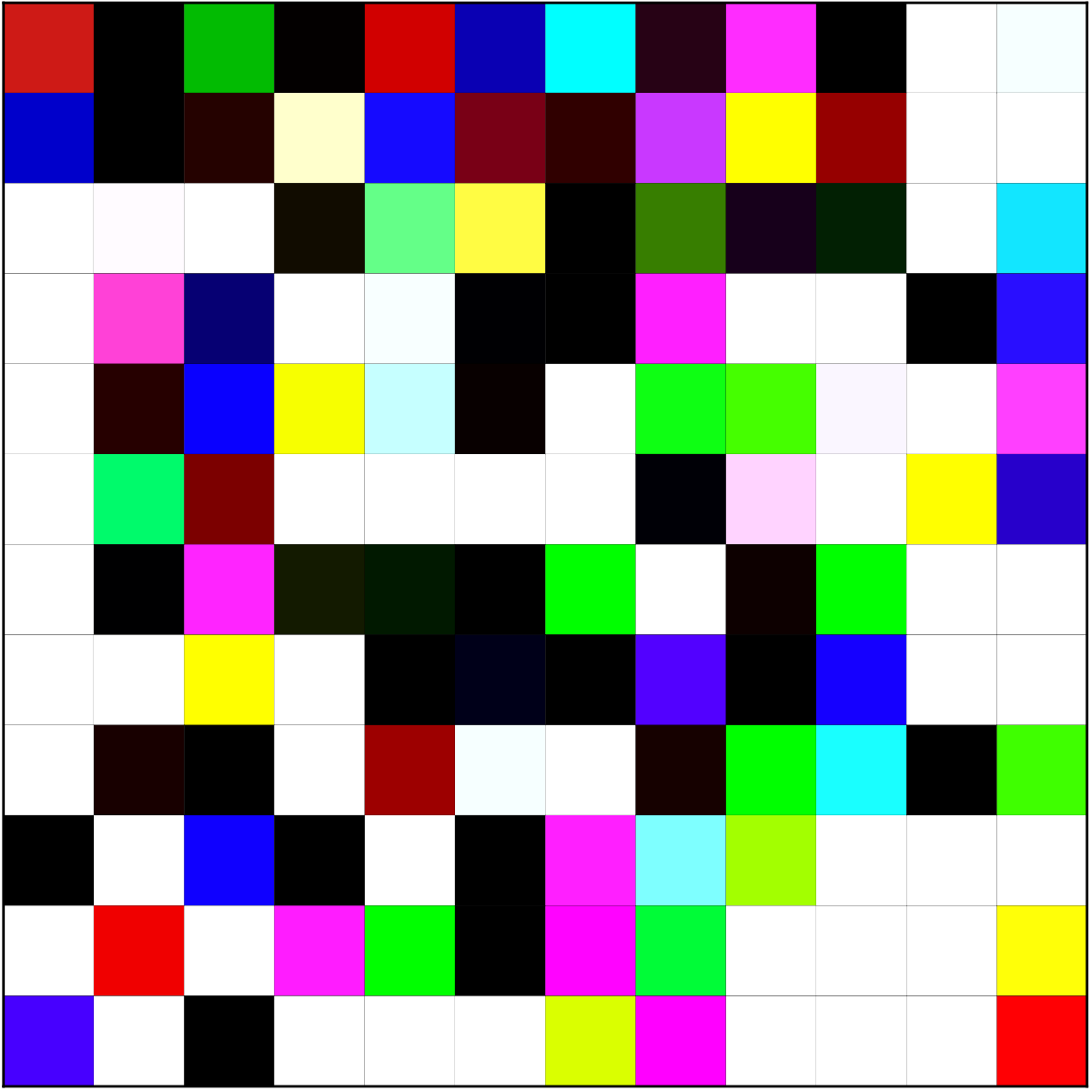}
     \caption{POS}
     \label{fig:prune-pos-deberta}
 \end{subfigure}
 \hfill
 \begin{subfigure}[b]{0.4\columnwidth}
     \centering
     \includegraphics[width=\textwidth]{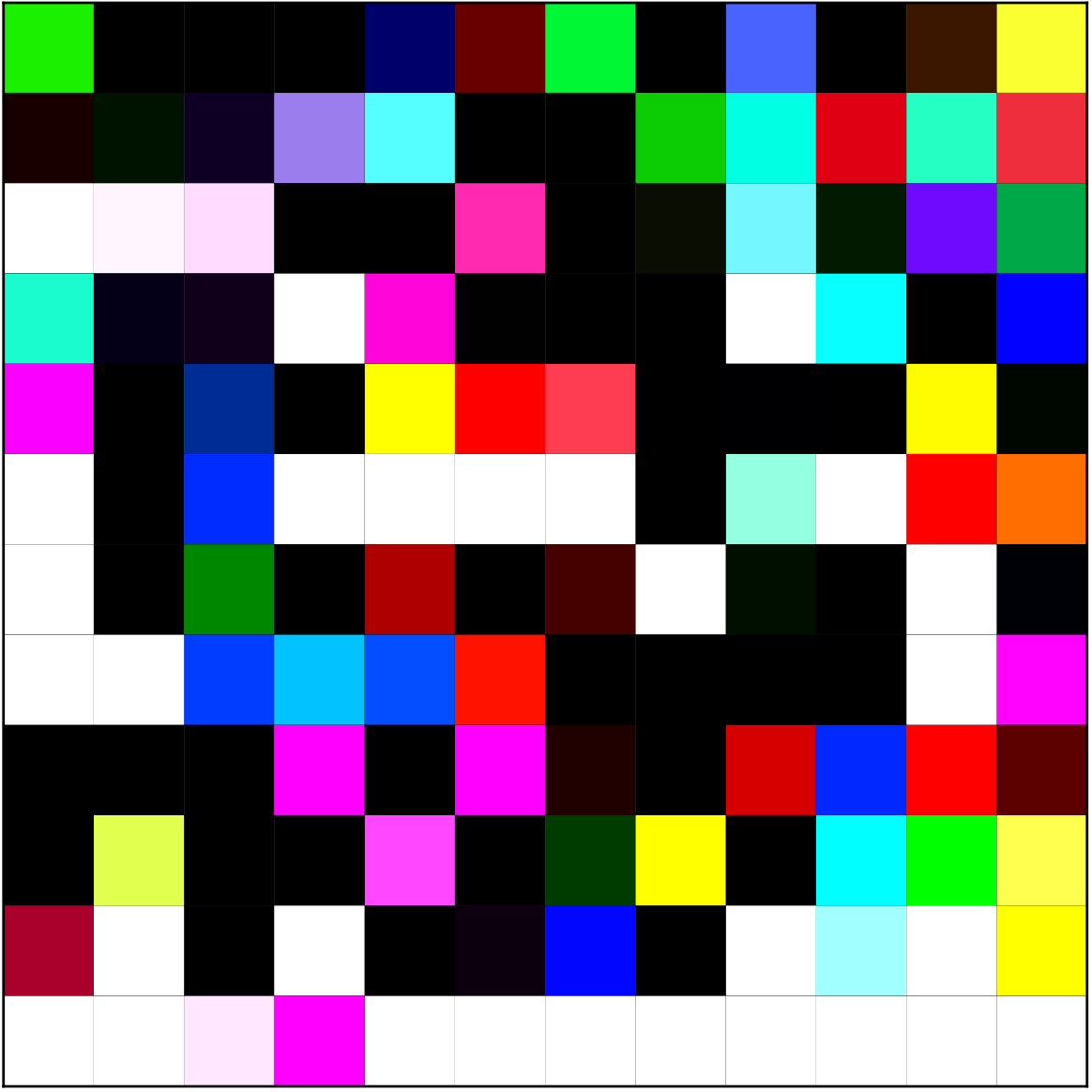}
     \caption{NER}
     \label{fig:prune-ner-deberta}
 \end{subfigure}
 \hfill
 \begin{subfigure}[b]{0.4\columnwidth}
     \centering
     \includegraphics[width=\textwidth]{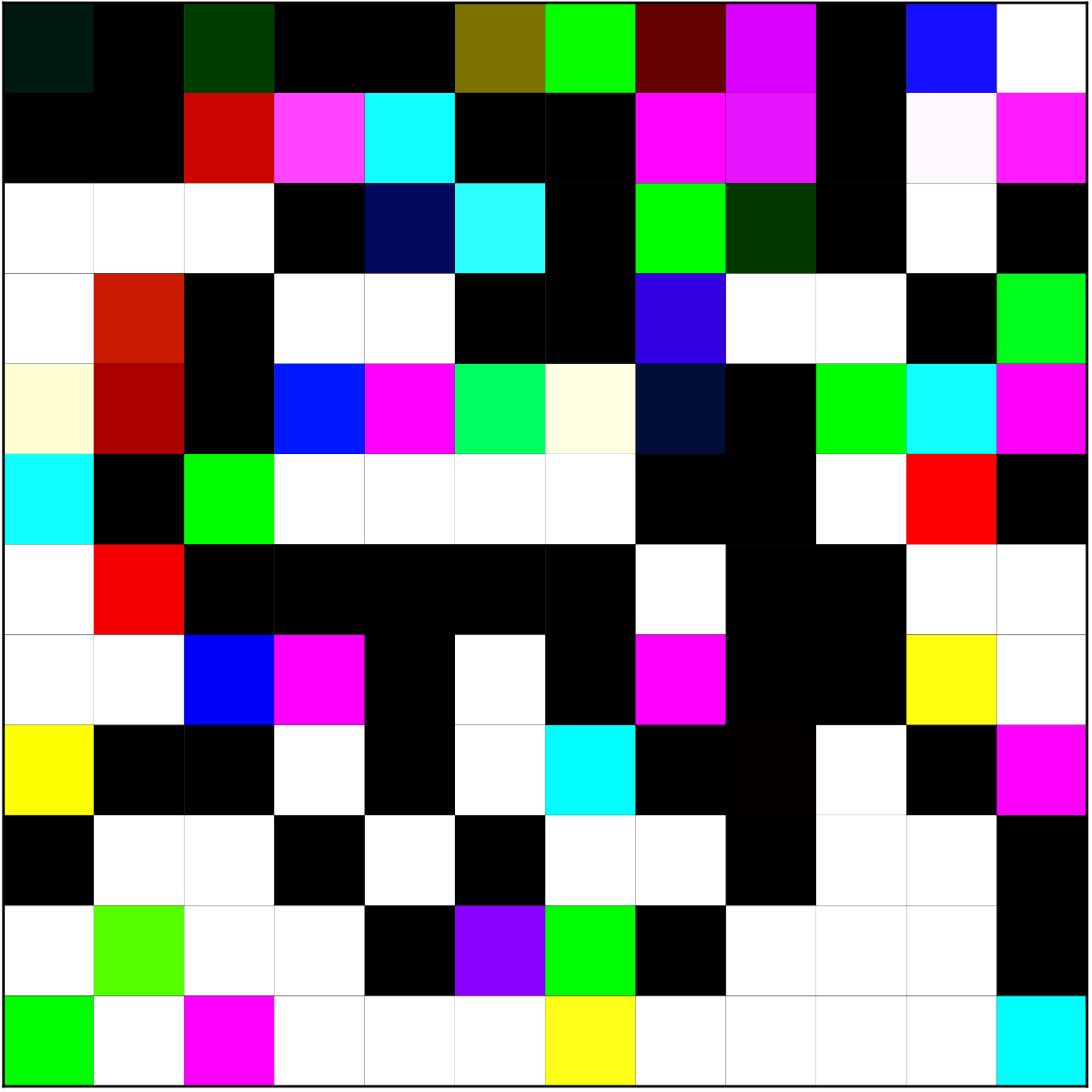}
     \caption{DEP}
     \label{fig:prune-dep-deberta}
 \end{subfigure}
 \hfill
 \begin{subfigure}[b]{0.4\columnwidth}
     \centering
     \includegraphics[width=\textwidth]{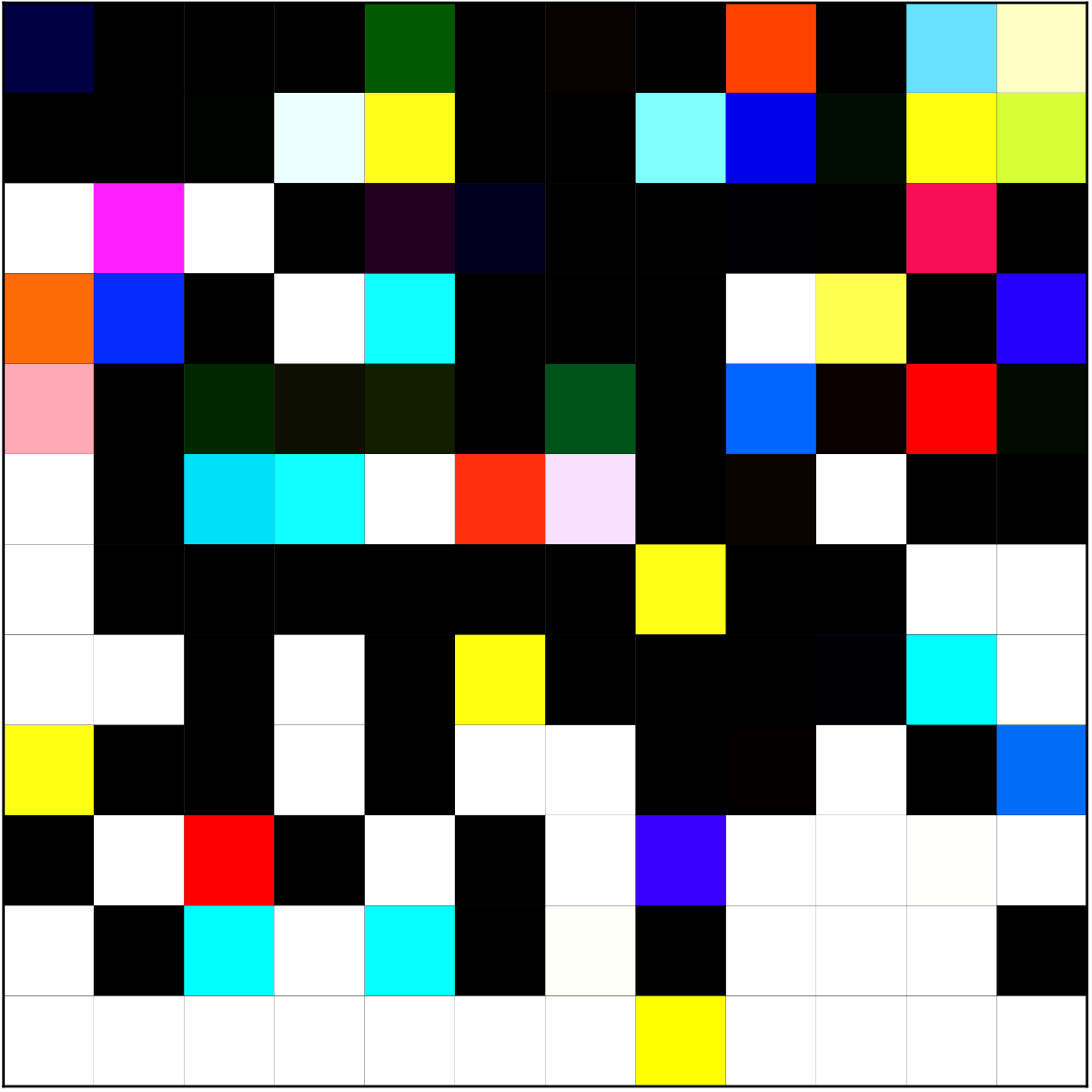}
     \caption{CON}
     \label{fig:prune-con-deberta}
 \end{subfigure}
 \hfill
 \begin{subfigure}[b]{0.4\columnwidth}
     \centering
     \includegraphics[width=\textwidth]{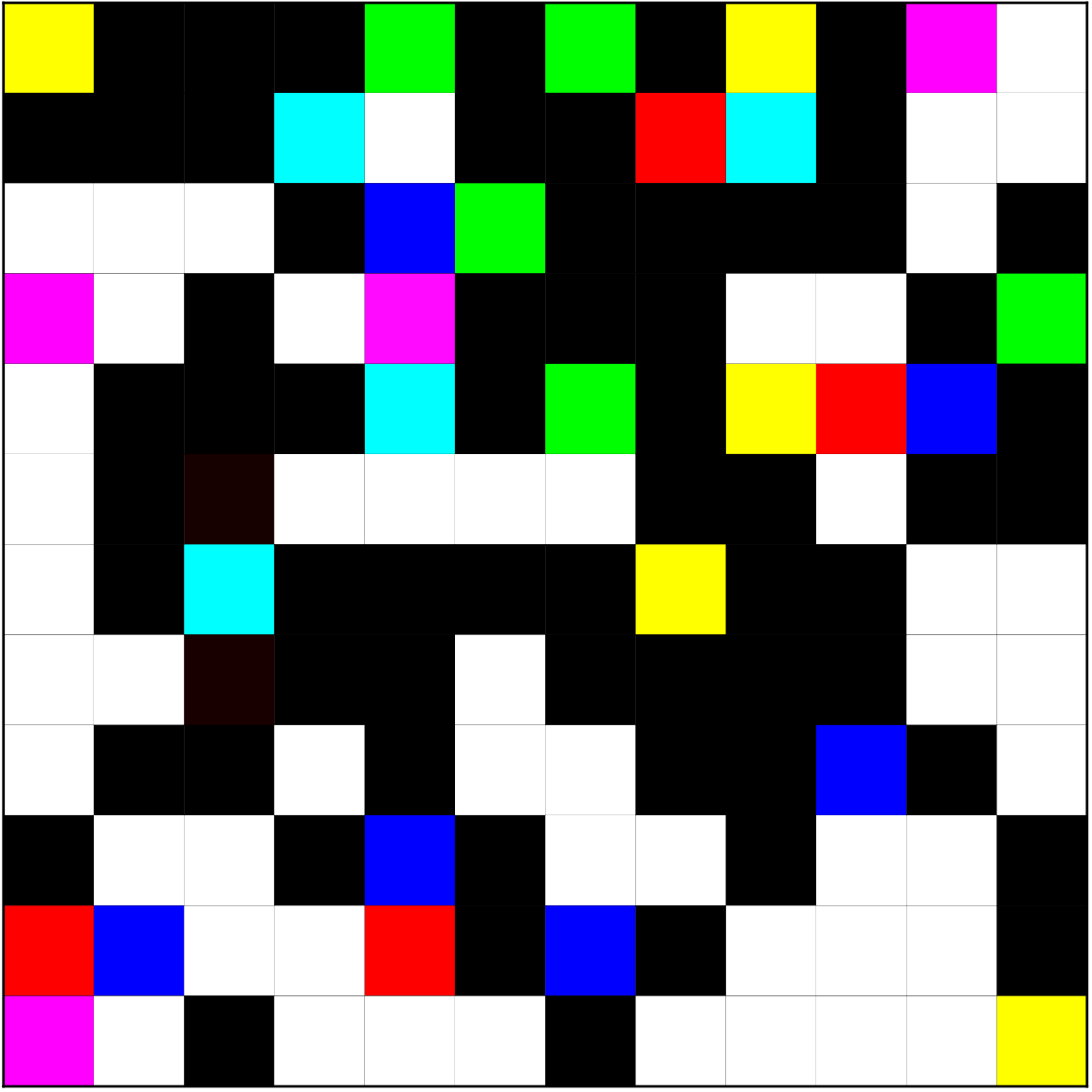}
     \caption{SRL}
     \label{fig:prune-srl-deberta}
 \end{subfigure}

 \begin{subfigure}[b]{\columnwidth}
     \centering
     \includegraphics[width=\textwidth]{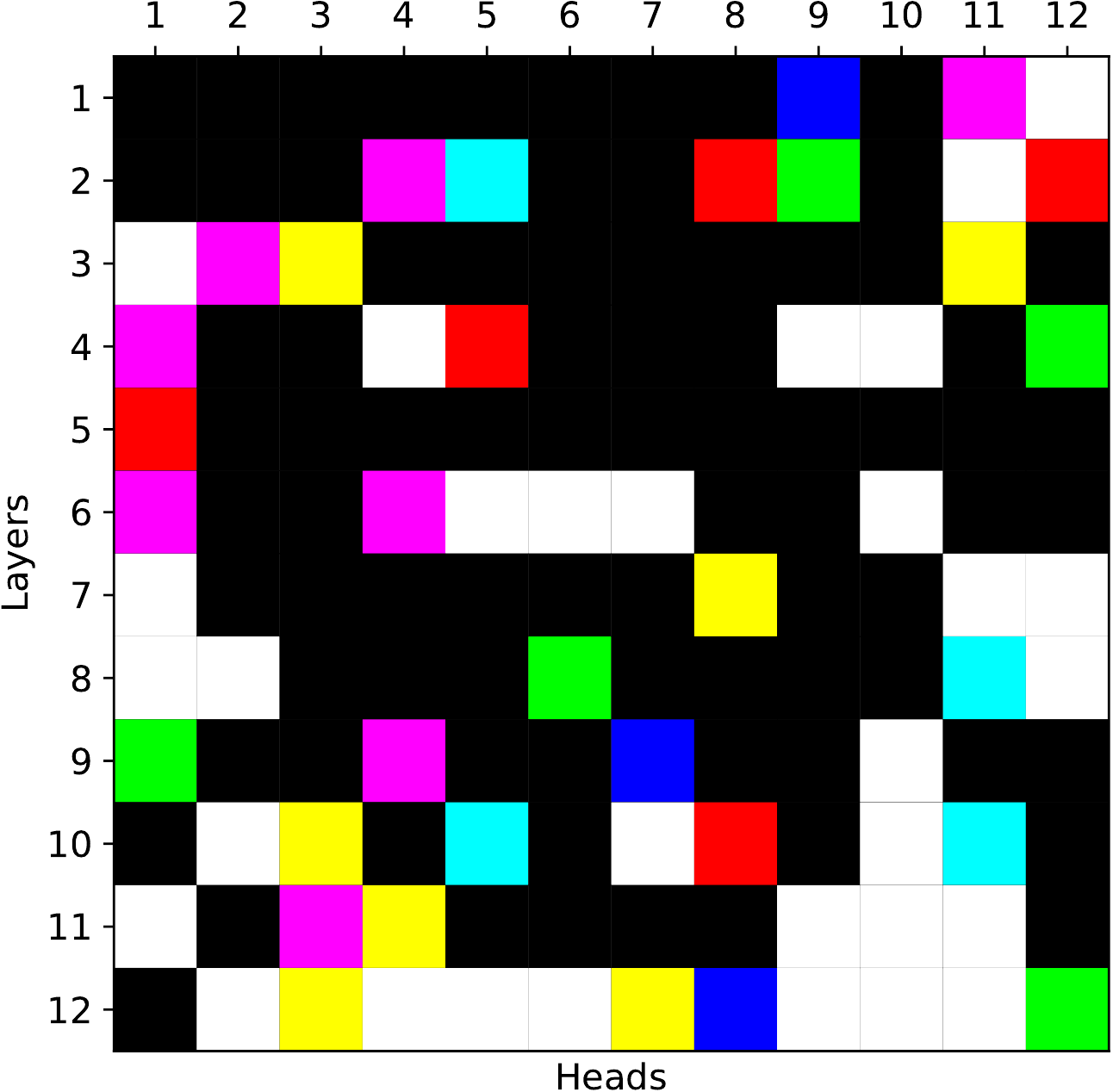}
     \caption{3-run utilization of the MTL-DP model, where each run is encoded in a RGB channel. Darker indicates higher utilization.}
     \label{fig:prune-joint-deberta}
 \end{subfigure}
 \hfill
 \begin{subfigure}[b]{\columnwidth}
     \centering
     \includegraphics[width=\textwidth]{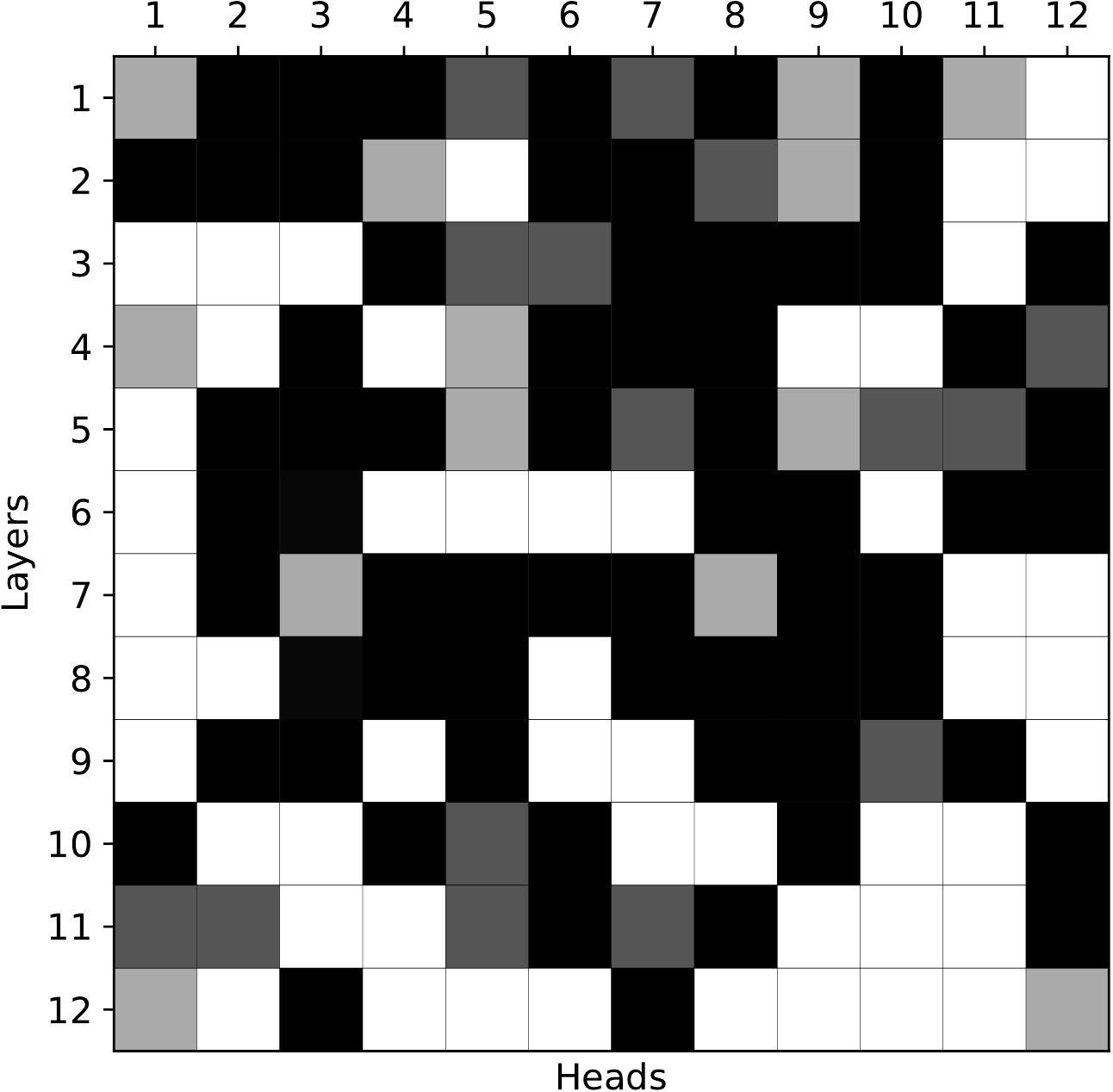}
     \caption{Average head utilization rates among the 5 tasks in 3 runs. Darker cells indicate higher utilization rates.}
     \label{fig:probe-overlay-deberta}
 \end{subfigure}
 \hfill
 
\caption{Head utilization of the DeBERTa \cite{he2020deberta} STL-DP models (a - e, g) and the MTL-DP model (f).}
\label{fig:overlap-deberta}
\end{figure*}

Their visualization of head utilization is illustrated in Figure~\ref{fig:overlap-roberta}, \ref{fig:overlap-electra} and \ref{fig:overlap-deberta}. Similar patterns among each transformer encoder can be observed, supporting our claim that the MTL-DP model re-uses a very similar set of heads used by the STL-DP models.

\begin{figure*}[ht]
 \centering
 \begin{subfigure}[b]{0.67\columnwidth}
     \centering
     \includegraphics[width=\textwidth]{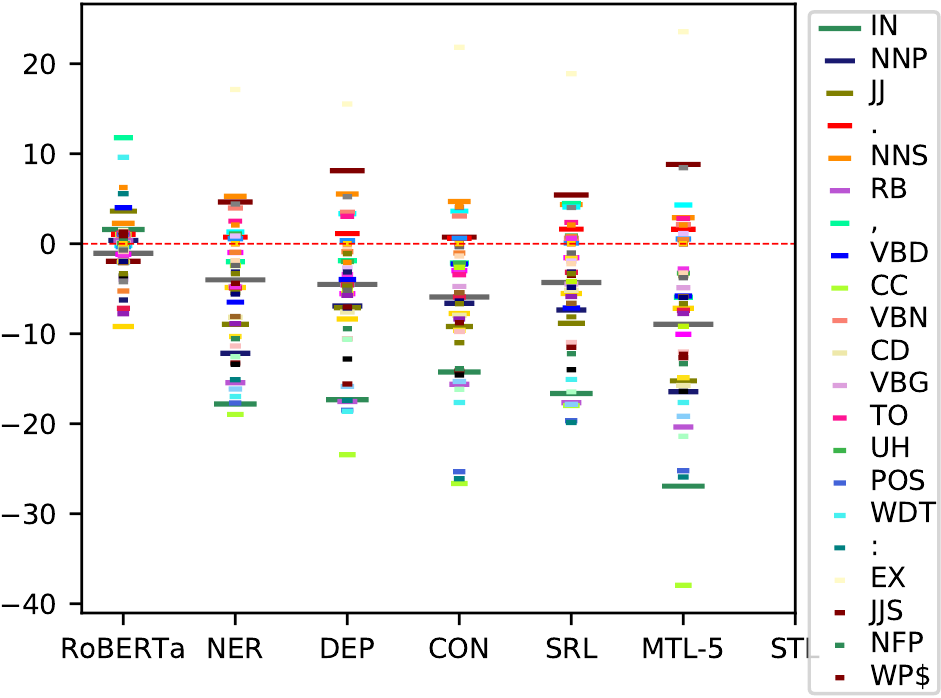}
     \caption{POS}
     \label{fig:probe-pos-roberta}
 \end{subfigure}
 \hfill
 \begin{subfigure}[b]{0.67\columnwidth}
     \centering
     \includegraphics[width=\textwidth]{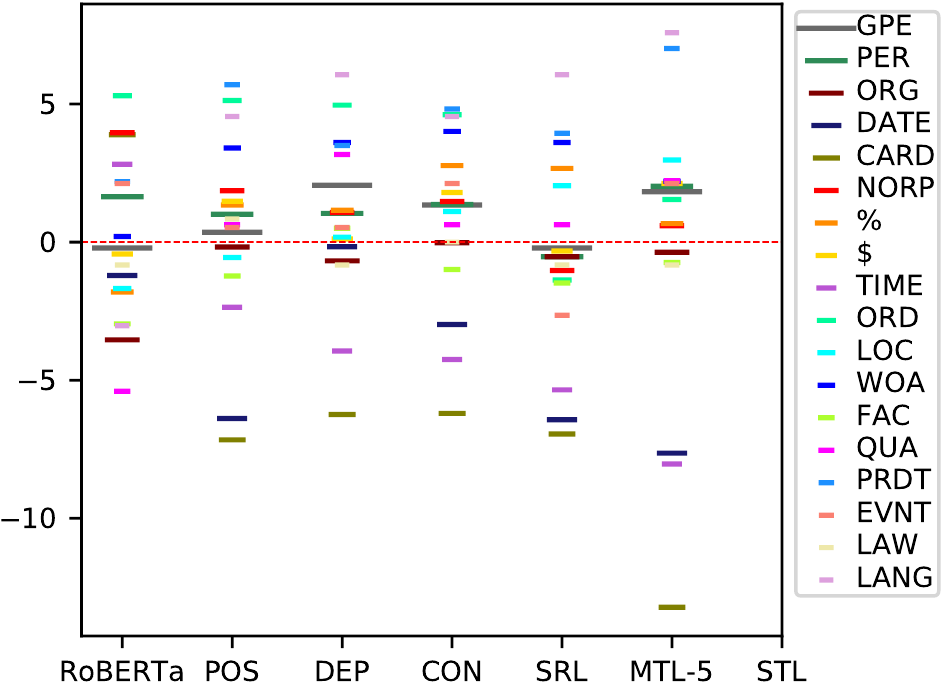}
     \caption{NER}
     \label{fig:probe-ner-roberta}
 \end{subfigure}
 \hfill
 \begin{subfigure}[b]{0.67\columnwidth}
     \centering
     \includegraphics[width=\textwidth]{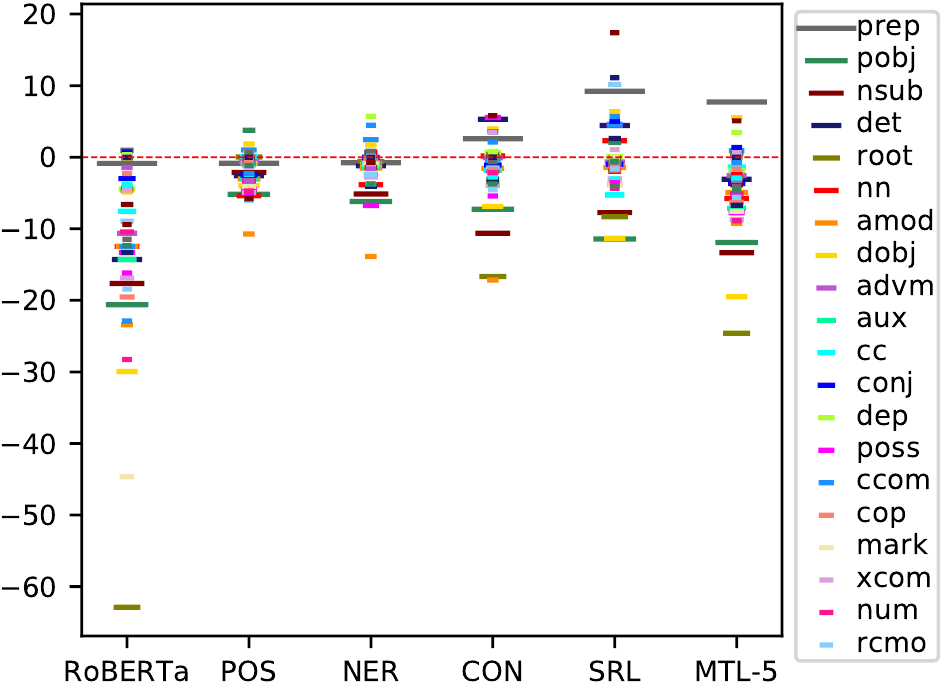}
     \caption{DEP}
     \label{fig:probe-dep-roberta}
 \end{subfigure}
 \hfill
  \begin{subfigure}[b]{0.67\columnwidth}
     \centering
     \includegraphics[width=\textwidth]{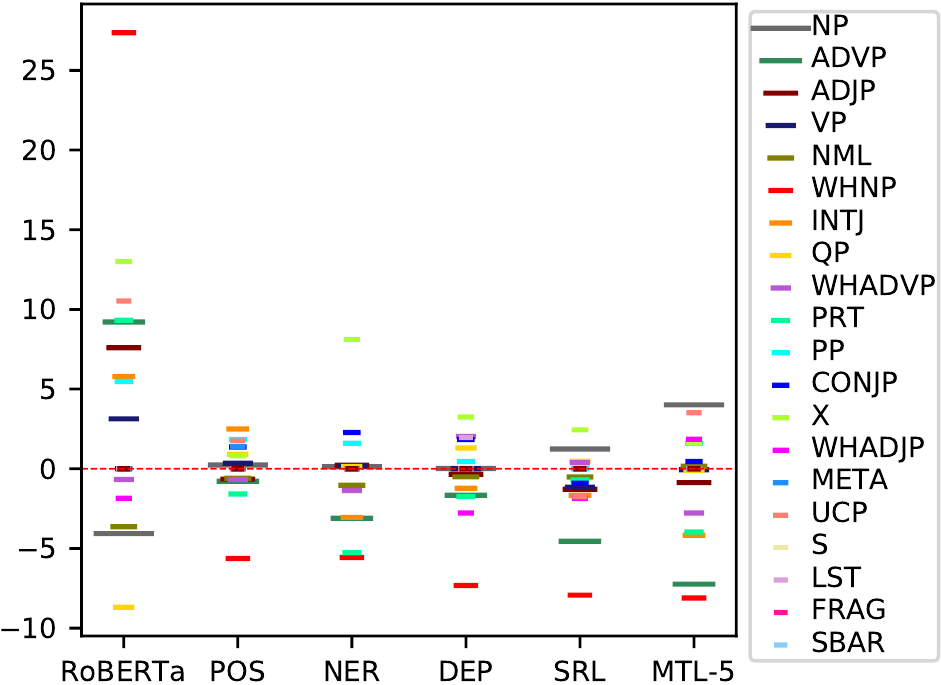}
     \caption{CON}
     \label{fig:probe-con-roberta}
 \end{subfigure}
 \hfill
  \begin{subfigure}[b]{0.67\columnwidth}
     \centering
     \includegraphics[width=\textwidth]{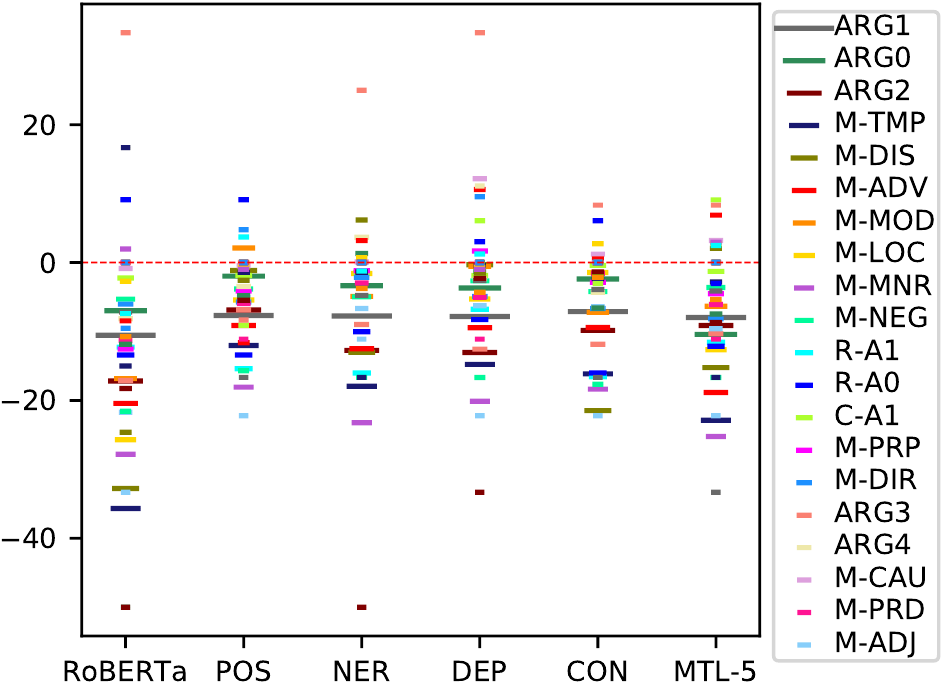}
     \caption{SRL}
     \label{fig:probe-srl-roberta}
 \end{subfigure}
 \hfill
  \begin{subfigure}[b]{0.67\columnwidth}
     \centering
     \includegraphics[width=\textwidth]{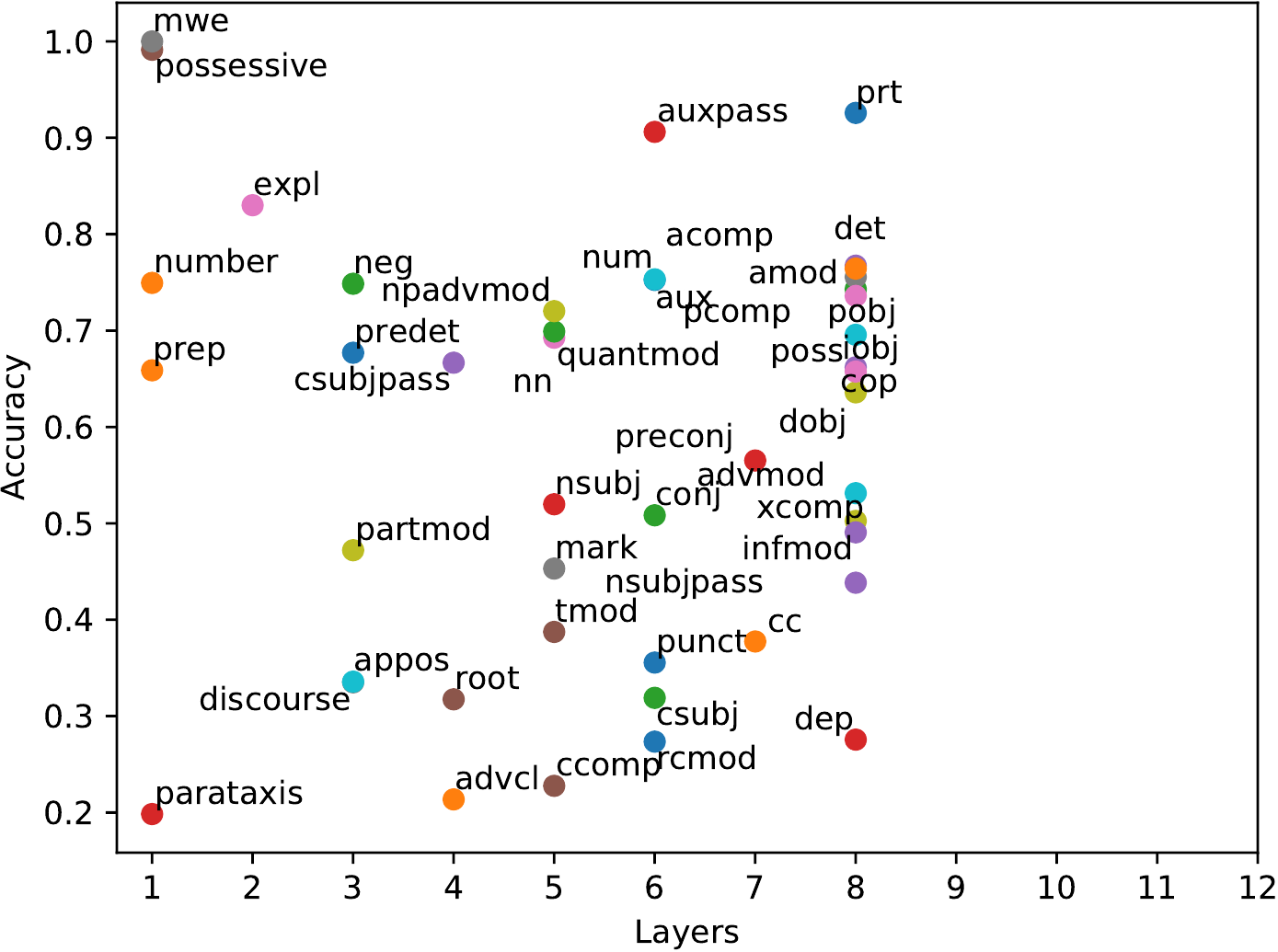}
     \caption{\DEP\ layer analysis.}
     \label{fig:probe-dep-layer-roberta}
 \end{subfigure}
 \hfill
\caption{The RoBERTa \cite{liu2019roberta} probing results comparison (a - e) and layer analysis of pre-trained heads (g).}
\label{fig:probing-roberta}
\end{figure*}

\begin{figure*}[ht]
 \centering
 \begin{subfigure}[b]{0.67\columnwidth}
     \centering
     \includegraphics[width=\textwidth]{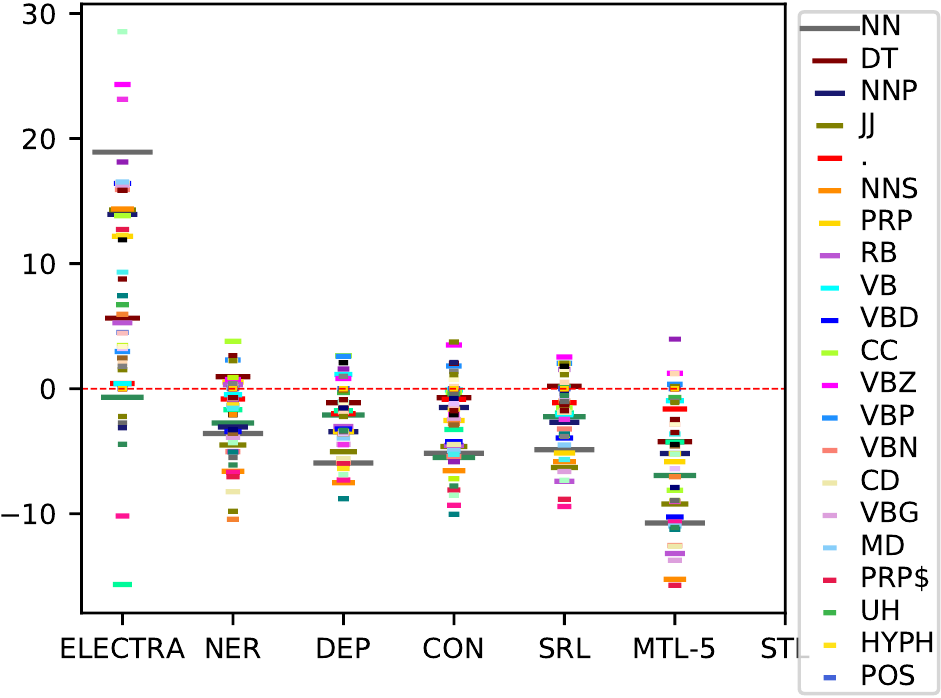}
     \caption{POS}
     \label{fig:probe-pos-electra}
 \end{subfigure}
 \hfill
 \begin{subfigure}[b]{0.67\columnwidth}
     \centering
     \includegraphics[width=\textwidth]{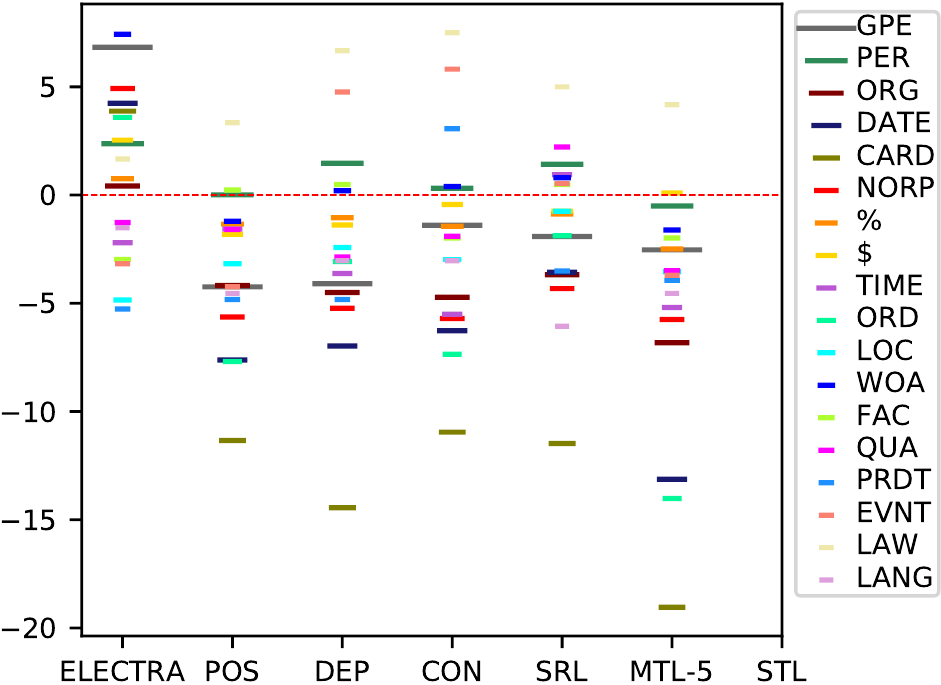}
     \caption{NER}
     \label{fig:probe-ner-electra}
 \end{subfigure}
 \hfill
 \begin{subfigure}[b]{0.67\columnwidth}
     \centering
     \includegraphics[width=\textwidth]{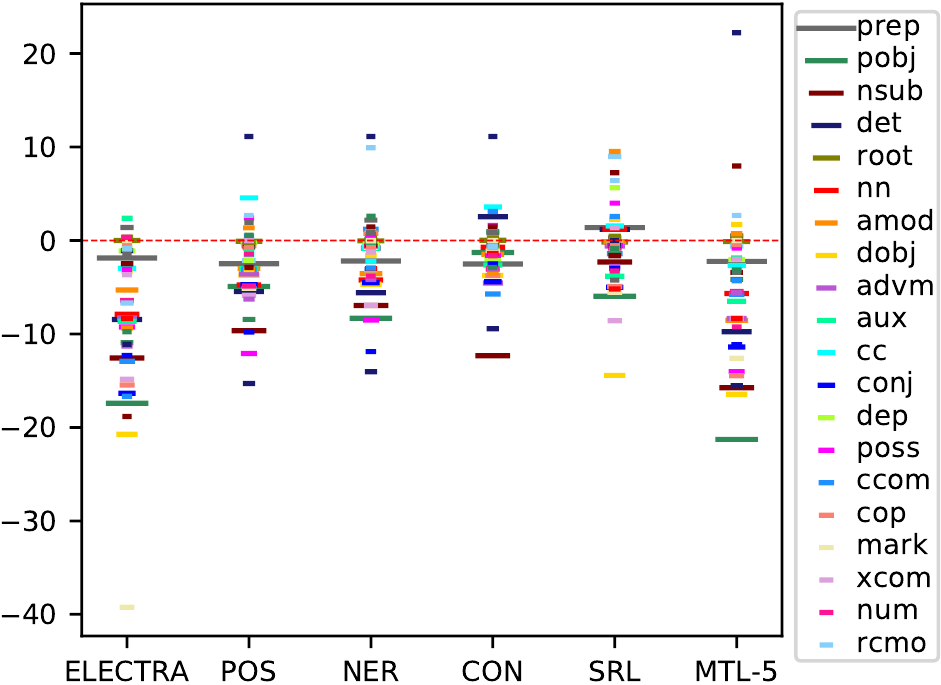}
     \caption{DEP}
     \label{fig:probe-dep-electra}
 \end{subfigure}
 \hfill
  \begin{subfigure}[b]{0.67\columnwidth}
     \centering
     \includegraphics[width=\textwidth]{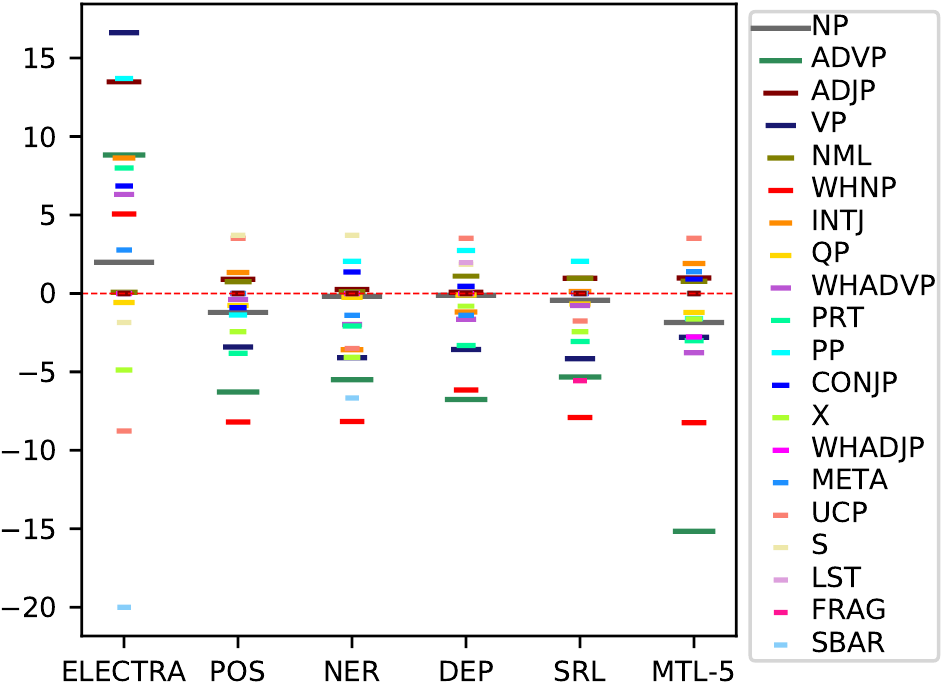}
     \caption{CON}
     \label{fig:probe-con-electra}
 \end{subfigure}
 \hfill
  \begin{subfigure}[b]{0.67\columnwidth}
     \centering
     \includegraphics[width=\textwidth]{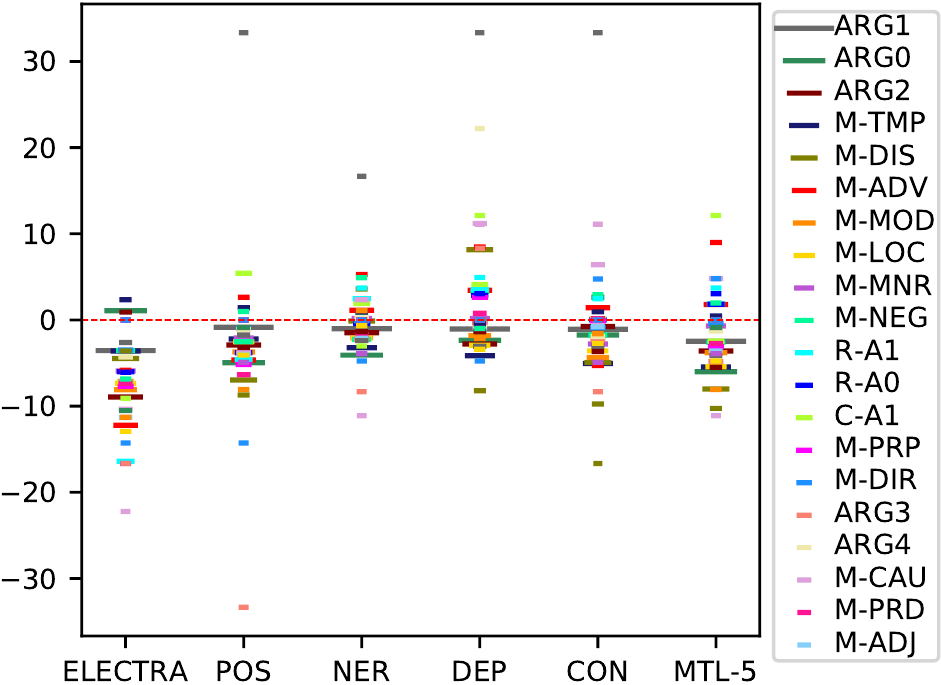}
     \caption{SRL}
     \label{fig:probe-srl-electra}
 \end{subfigure}
 \hfill
  \begin{subfigure}[b]{0.67\columnwidth}
     \centering
     \includegraphics[width=\textwidth]{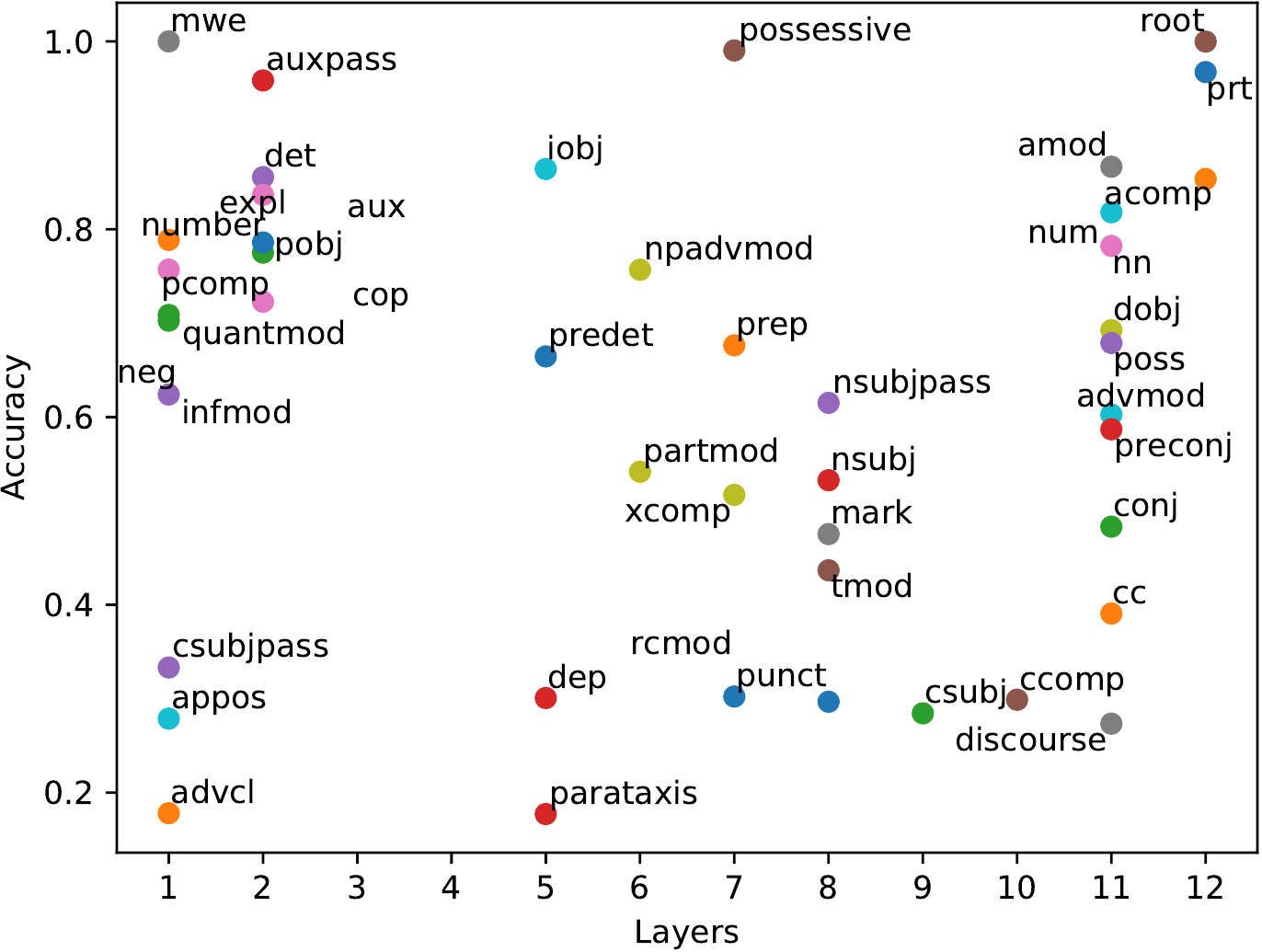}
     \caption{\DEP\ layer analysis.}
     \label{fig:probe-dep-layer-electra}
 \end{subfigure}
 \hfill
\caption{The ELECTRA \cite{clark2020electra} probing results comparison (a - e) and layer analysis of pre-trained heads (g).}
\label{fig:probing-electra}
\end{figure*}

\begin{figure*}[ht]
 \centering
 \begin{subfigure}[b]{0.67\columnwidth}
     \centering
     \includegraphics[width=\textwidth]{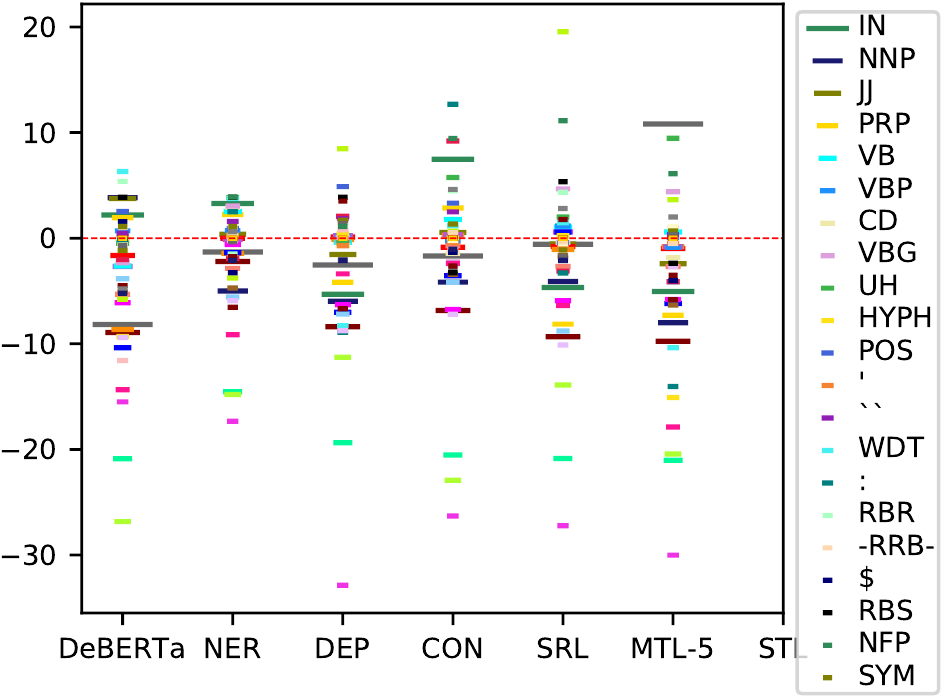}
     \caption{POS}
     \label{fig:probe-pos-deberta}
 \end{subfigure}
 \hfill
 \begin{subfigure}[b]{0.67\columnwidth}
     \centering
     \includegraphics[width=\textwidth]{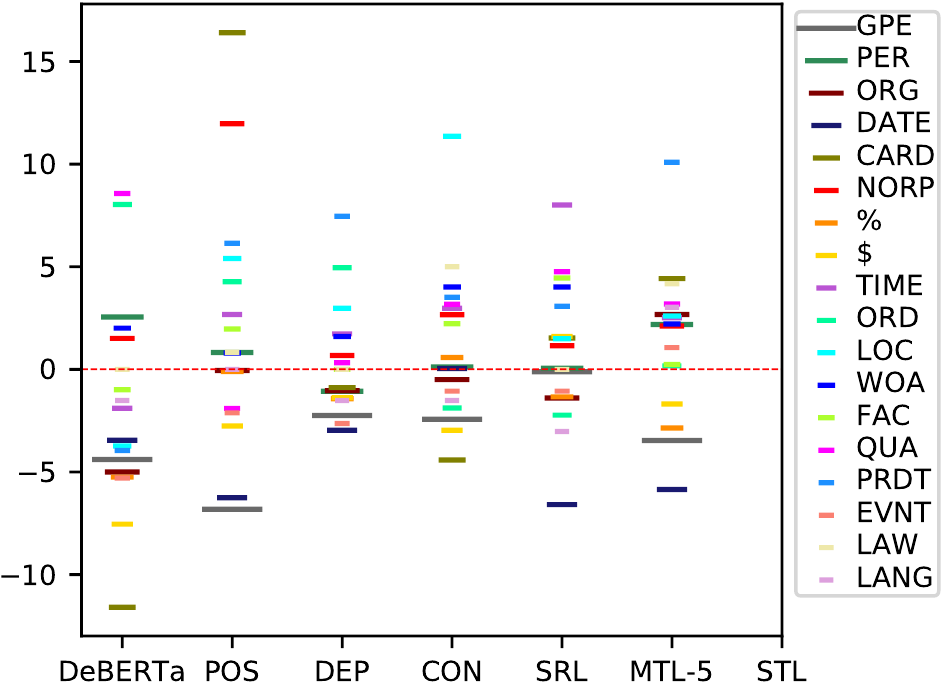}
     \caption{NER}
     \label{fig:probe-ner-deberta}
 \end{subfigure}
 \hfill
 \begin{subfigure}[b]{0.67\columnwidth}
     \centering
     \includegraphics[width=\textwidth]{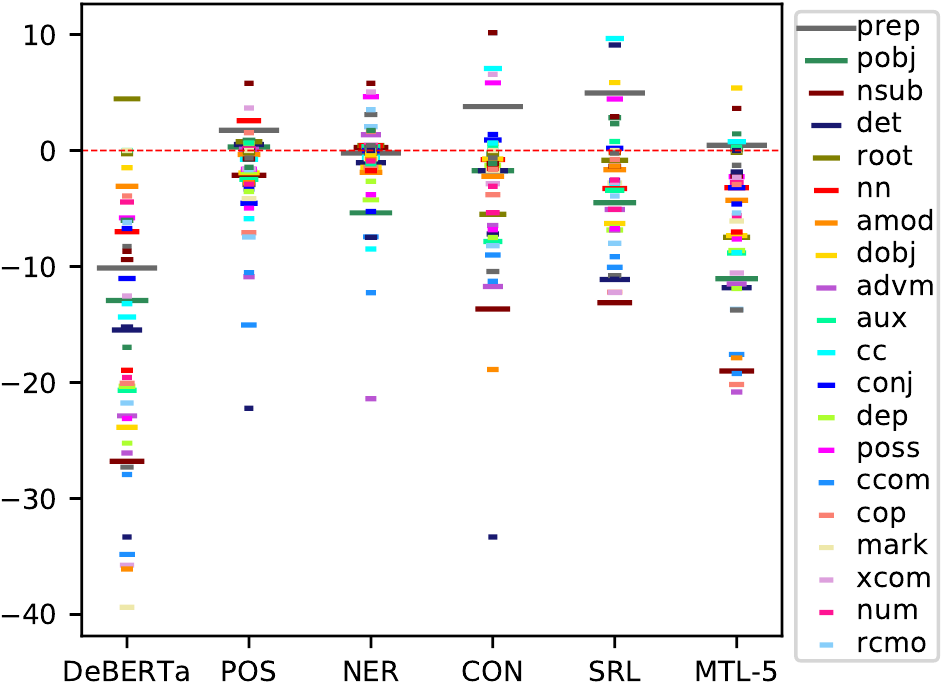}
     \caption{DEP}
     \label{fig:probe-dep-deberta}
 \end{subfigure}
 \hfill
  \begin{subfigure}[b]{0.67\columnwidth}
     \centering
     \includegraphics[width=\textwidth]{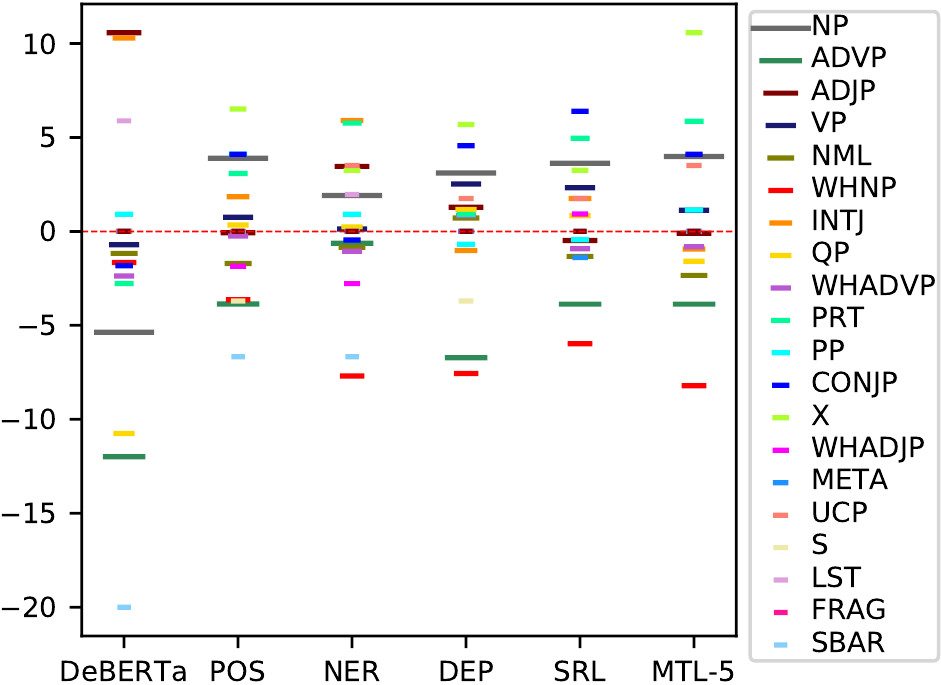}
     \caption{CON}
     \label{fig:probe-con-deberta}
 \end{subfigure}
 \hfill
  \begin{subfigure}[b]{0.67\columnwidth}
     \centering
     \includegraphics[width=\textwidth]{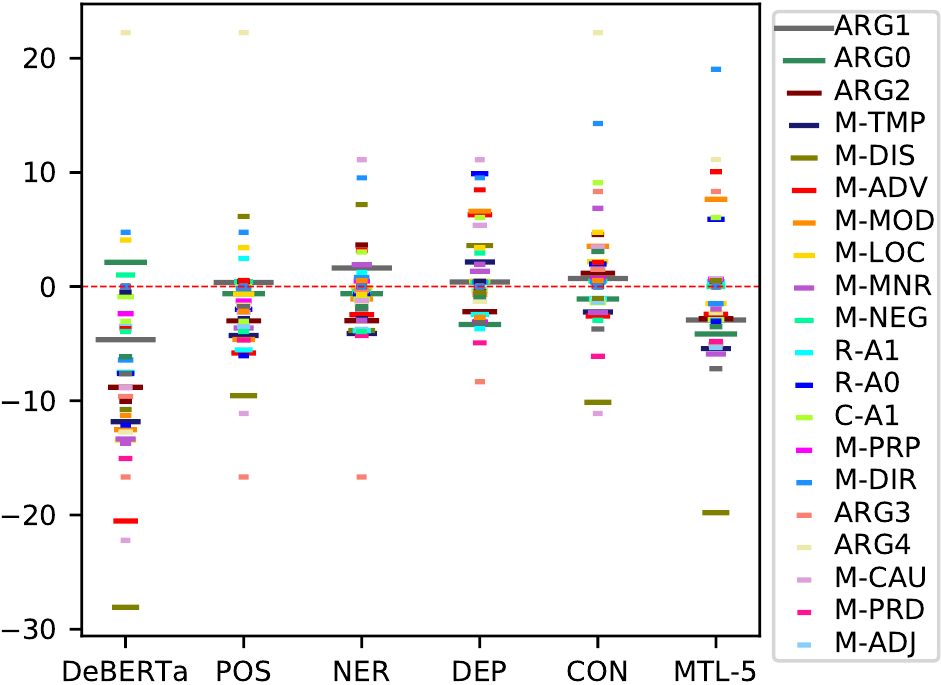}
     \caption{SRL}
     \label{fig:probe-srl-deberta}
 \end{subfigure}
 \hfill
  \begin{subfigure}[b]{0.67\columnwidth}
     \centering
     \includegraphics[width=\textwidth]{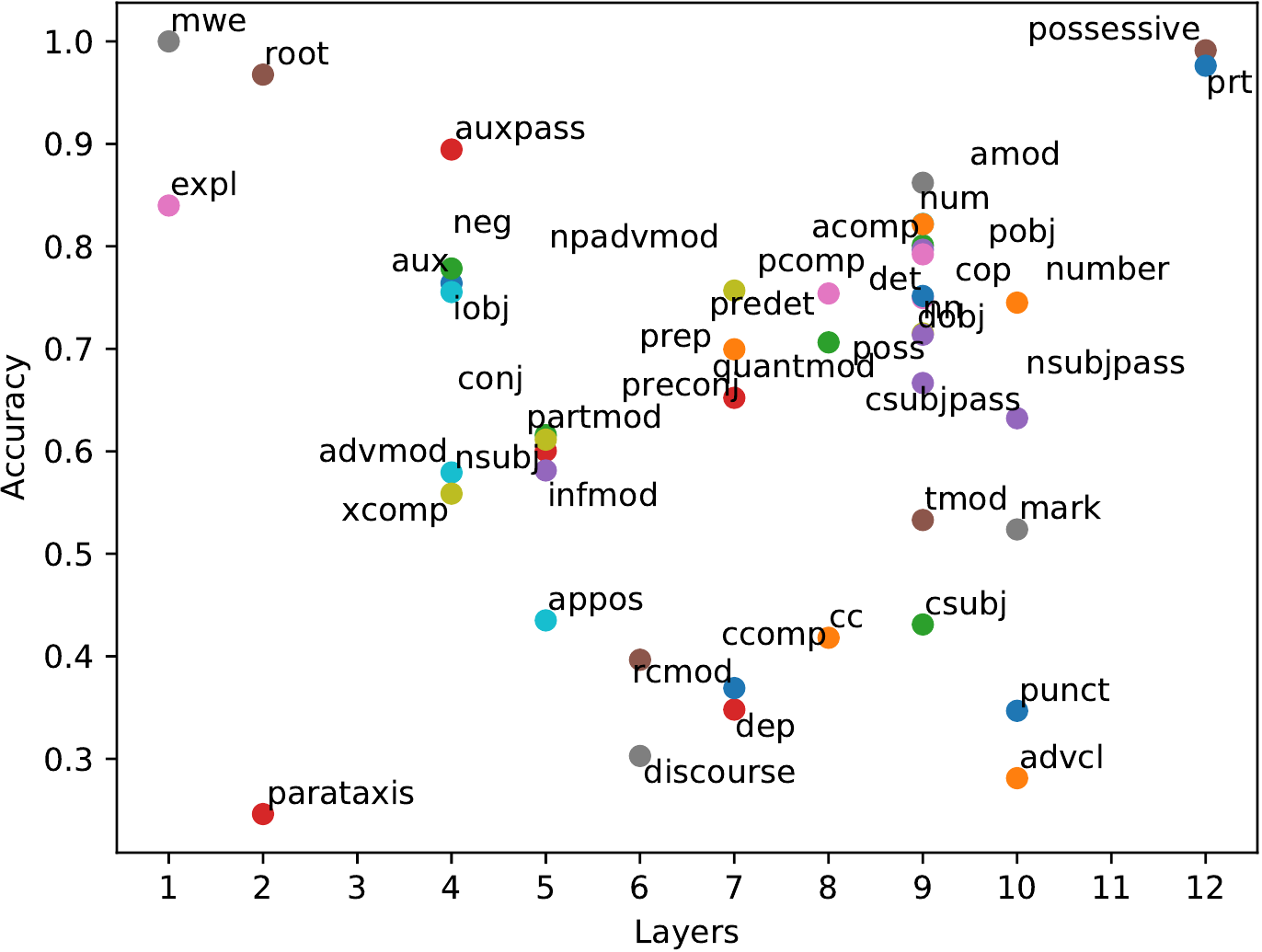}
     \caption{\DEP\ layer analysis.}
     \label{fig:probe-dep-layer-deberta}
 \end{subfigure}
 \hfill
\caption{The DeBERTa \cite{he2020deberta} probing results comparison (a - e) and layer analysis of pre-trained heads (g).}
\label{fig:probing-deberta}
\end{figure*}

Their probing results are illustrated in Figure~\ref{fig:probing-roberta}, \ref{fig:probing-electra} and \ref{fig:probing-deberta}, which also align with our findings. Specifically, the \DEP\; probing results on transformer encoders are already very high even without fine-tuning on actual dependency treebanks (Figure~\ref{fig:probe-dep-layer-roberta}, \ref{fig:probe-dep-layer-electra} and \ref{fig:probe-dep-layer-deberta}), demonstrating the pluripotency of \textit{stem cells}. Other figures show similar patterns that \textit{stem cell} heads get specialized during STL but lose specialities in MTL, supporting our \textit{Stem Cell Hypothesis}.

\begin{table*}[!htbp]
\begin{subtable}{\textwidth}
        \centering\small
\begin{tabular}{c||c|c|c|c|c||c}
    & \textbf{\POS}          & \textbf{\NER}          & \textbf{\DEP}          & \textbf{\CON}          & \textbf{\SRL}          & \bf MTL-5          \\ \hline \hline
\textbf{\POS} & \cellcolor{gray!32} 98.33 $\pm$ 0.01 & 98.30 $\pm$ 0.01 & \textbf{98.34 $\pm$ 0.00} & 98.33 $\pm$ 0.02          & 98.31 $\pm$ 0.02 & 98.29 $\pm$ 0.02 \\
\textbf{\NER} & 89.44 $\pm$ 0.04 & \cellcolor{gray!32}88.93 $\pm$ 0.16 & \textbf{89.84 $\pm$ 0.14} & 89.65 $\pm$ 0.02          & 89.37 $\pm$ 0.19 & 89.66 $\pm$ 0.17 \\
\textbf{\DEP} & 94.48 $\pm$ 0.05 & 94.46 $\pm$ 0.04 & \cellcolor{gray!32}\textbf{94.56 $\pm$ 0.05} & 94.55 $\pm$ 0.02          & 94.44 $\pm$ 0.02 & 94.38 $\pm$ 0.03 \\
\textbf{\CON} & 94.82 $\pm$ 0.01 & 94.76 $\pm$ 0.04 & 94.88 $\pm$ 0.03          & \cellcolor{gray!32}\textbf{94.89 $\pm$ 0.05} & 94.88 $\pm$ 0.01 & 94.75 $\pm$ 0.01 \\
\textbf{\SRL} & 84.04 $\pm$ 0.05 & 83.31 $\pm$ 0.11 & 84.13 $\pm$ 0.08          & \textbf{84.14 $\pm$ 0.02} & \cellcolor{gray!32}83.52 $\pm$ 0.07 & 83.56 $\pm$ 0.10
\end{tabular}
\caption{RoBERTa \cite{liu2019roberta} performance.}
       \label{tab:mtl-results-roberta}
\end{subtable}
\hfill
\begin{subtable}{\textwidth}
        \centering\small
\begin{tabular}{c||c|c|c|c|c||c}
    & \textbf{\POS}          & \textbf{\NER}          & \textbf{\DEP}          & \textbf{\CON}          & \textbf{\SRL}          & \bf MTL-5          \\ \hline \hline
\bf \POS & \cellcolor{gray!32}97.94 $\pm$ 0.04 & \textbf{97.97 $\pm$ 0.02} & 97.94 $\pm$ 0.01 & 97.95 $\pm$ 0.01          & 97.96 $\pm$ 0.05 & 97.93 $\pm$ 0.01 \\
\bf \NER & 88.62 $\pm$ 0.03 & \cellcolor{gray!32}88.22 $\pm$ 0.05          & 88.46 $\pm$ 0.20 & \textbf{88.82 $\pm$ 0.14} & 88.25 $\pm$ 0.11 & 88.38 $\pm$ 0.04 \\
\bf \DEP & 94.52 $\pm$ 0.08 & 94.47 $\pm$ 0.08          & \cellcolor{gray!32}94.61 $\pm$ 0.01 & \textbf{94.65 $\pm$ 0.03} & 94.53 $\pm$ 0.07 & 94.39 $\pm$ 0.04 \\
\bf \CON & 95.01 $\pm$ 0.01 & 95.04 $\pm$ 0.02          & 95.10 $\pm$ 0.06 & \cellcolor{gray!32}\textbf{95.14 $\pm$ 0.03} & 95.08 $\pm$ 0.07 & 94.91 $\pm$ 0.04 \\
\bf \SRL & 84.35 $\pm$ 0.07 & 83.44 $\pm$ 0.05          & 84.55 $\pm$ 0.06 & \textbf{84.61 $\pm$ 0.11} & \cellcolor{gray!32}83.81 $\pm$ 0.11 & 84.09 $\pm$ 0.06
\end{tabular}
\caption{ELECTRA \cite{clark2020electra} performance.}
       \label{tab:mtl-results-electra}
\end{subtable}
\hfill
\begin{subtable}{\textwidth}
        \centering\small
\begin{tabular}{c||c|c|c|c|c||c}
    & \textbf{\POS}          & \textbf{\NER}          & \textbf{\DEP}          & \textbf{\CON}          & \textbf{\SRL}          & \bf MTL-5          \\ \hline \hline
\textbf{\POS} & \cellcolor{gray!32}\textbf{98.33 $\pm$ 0.02} & 98.26 $\pm$ 0.02 & 98.32 $\pm$ 0.01 & 98.32 $\pm$ 0.01 & 98.31 $\pm$ 0.02 & 98.30 $\pm$ 0.02 \\ 
\textbf{\NER} & 89.29 $\pm$ 0.18 & \cellcolor{gray!32}89.02 $\pm$ 0.21 & \textbf{89.67 $\pm$ 0.09} & 89.65 $\pm$ 0.25 & 89.29 $\pm$ 0.36 & 89.28 $\pm$ 0.18 \\ 
\textbf{\DEP} & 94.50 $\pm$ 0.04 & 94.54 $\pm$ 0.02 & \cellcolor{gray!32}\textbf{94.64 $\pm$ 0.03} & 94.61 $\pm$ 0.07 & 94.56 $\pm$ 0.04 & 94.42 $\pm$ 0.03 \\ 
\textbf{\CON} & 94.91 $\pm$ 0.04 & 94.94 $\pm$ 0.02 & 94.96 $\pm$ 0.01 & \cellcolor{gray!32}\textbf{95.06 $\pm$ 0.01} & 94.98 $\pm$ 0.03 & 94.82 $\pm$ 0.03 \\ 
\textbf{\SRL} & 84.26 $\pm$ 0.05 & 83.42 $\pm$ 0.15 & 84.34 $\pm$ 0.04 & \textbf{84.36 $\pm$ 0.09} & \cellcolor{gray!32}83.62 $\pm$ 0.06 & 83.37 $\pm$ 0.01 \\ 
\end{tabular}
\caption{DeBERTa \cite{he2020deberta} performance.}
     \label{tab:mtl-results-deberta}
     \end{subtable}
\caption{Performance of single-task learning (main diagonal highlighted in gray), multi-task learning on all 5 tasks (MTL-5), and multi-task learning on every pair of the tasks (non-diagonal cells; e.g., \DEP'th row in \NER'th column is the \DEP\ result of the joint model between \DEP\ and \NER).}
\label{tab:mtl-results-more}
\end{table*}

\begin{table*}[htbp]
\begin{subtable}[h]{\textwidth}
\centering\small{
\begin{tabular}{c||c|c|c|c||c|c}
    & \multicolumn{4}{c||}{\bf Performance} & \multicolumn{2}{c}{\bf \% of Attention Heads Kept}                   \\ \cline{2-7}
    & \bf STL & \bf STL-DP & \bf MTL & \bf MTL-DP & \bf STL-DP & \bf MTL-DP  \\ \hline\hline
\bf \POS & 98.33 $\pm$ 0.01          & 98.35 $\pm$ 0.02 & 98.29 $\pm$ 0.02          & \textbf{98.38 $\pm$ 0.01} & \textbf{41.90 $\pm$ 0.40} & 53.47 $\pm$ 1.39          \\
\bf \NER & 88.93 $\pm$ 0.16          & 89.08 $\pm$ 0.16 & \textbf{89.66 $\pm$ 0.17} & 89.52 $\pm$ 0.05          & 58.33 $\pm$ 3.48          & \textbf{53.47 $\pm$ 1.39} \\
\bf \DEP & \textbf{94.56 $\pm$ 0.05} & 94.29 $\pm$ 0.01 & 94.38 $\pm$ 0.03          & 94.48 $\pm$ 0.03          & 64.58 $\pm$ 1.21          & \textbf{53.47 $\pm$ 1.39} \\
\bf \CON & \textbf{94.89 $\pm$ 0.05} & 94.59 $\pm$ 0.05 & 94.75 $\pm$ 0.01          & 94.78 $\pm$ 0.06          & 55.33 $\pm$ 2.00          & \textbf{53.47 $\pm$ 1.39} \\
\bf \SRL & 83.52 $\pm$ 0.07          & 83.53 $\pm$ 0.06 & 83.56 $\pm$ 0.10          & \textbf{83.57 $\pm$ 0.14} & \textbf{45.14 $\pm$ 0.70} & 53.47 $\pm$ 1.39    
\end{tabular}
}
\caption{RoBERTa \cite{liu2019roberta} pruning results.}
\label{tab:pruning-results-roberta}
\end{subtable}
\hfill

\begin{subtable}[h]{\textwidth}
\centering\small{
\begin{tabular}{c||c|c|c|c||c|c}
    & \multicolumn{4}{c||}{\bf Performance} & \multicolumn{2}{c}{\bf \% of Attention Heads Kept}                   \\ \cline{2-7}
    & \bf STL & \bf STL-DP & \bf MTL & \bf MTL-DP & \bf STL-DP & \bf MTL-DP  \\ \hline\hline
\bf \POS & 97.94 $\pm$ 0.04          & 98.01 $\pm$ 0.01 & 97.93 $\pm$ 0.01          & \textbf{98.03 $\pm$ 0.02} & \textbf{35.88 $\pm$ 2.12} & 44.67 $\pm$ 2.44          \\
\bf \NER & 88.22 $\pm$ 0.05          & 88.04 $\pm$ 0.24 & \textbf{88.38 $\pm$ 0.04} & 88.11 $\pm$ 0.22          & 58.33 $\pm$ 2.78          & \textbf{44.67 $\pm$ 2.44} \\
\bf \DEP & \textbf{94.61 $\pm$ 0.01} & 94.36 $\pm$ 0.04 & 94.39 $\pm$ 0.04          & 94.40 $\pm$ 0.01          & 58.80 $\pm$ 1.06          & \textbf{44.67 $\pm$ 2.44} \\
\bf \CON & \textbf{95.14 $\pm$ 0.03} & 94.82 $\pm$ 0.02 & 94.91 $\pm$ 0.04          & 94.86 $\pm$ 0.07          & 52.78 $\pm$ 2.50          & \textbf{44.67 $\pm$ 2.44} \\
\bf \SRL & 83.81 $\pm$ 0.11          & 83.71 $\pm$ 0.10 & \textbf{84.09 $\pm$ 0.06} & 84.07 $\pm$ 0.06          & \textbf{41.20 $\pm$ 0.40} & 44.67 $\pm$ 2.44   
\end{tabular}
}
\caption{ELECTRA \cite{clark2020electra} pruning results.}
\label{tab:pruning-results-electra}
\end{subtable}
\hfill

\begin{subtable}[h]{\textwidth}
\centering\small{
\begin{tabular}{c||c|c|c|c||c|c}
    & \multicolumn{4}{c||}{\bf Performance} & \multicolumn{2}{c}{\bf \% of Attention Heads Kept}                   \\ \cline{2-7}
    & \bf STL & \bf STL-DP & \bf MTL & \bf MTL-DP & \bf STL-DP & \bf MTL-DP  \\ \hline\hline
\bf \POS & 98.33 $\pm$ 0.02 & 98.37 $\pm$ 0.01 & 98.30 $\pm$ 0.02 & \textbf{98.39 $\pm$ 0.01} & \textbf{50.93 $\pm$ 3.13} & 66.67 $\pm$ 1.84 \\
\bf \NER & 89.02 $\pm$ 0.21 & 88.72 $\pm$ 0.18 & \textbf{89.28 $\pm$ 0.18} & 89.08 $\pm$ 0.07 & 66.67 $\pm$ 3.67 & 66.67 $\pm$ 1.84 \\
\bf \DEP & \textbf{94.64 $\pm$ 0.03} & 94.42 $\pm$ 0.02 & 94.42 $\pm$ 0.03 & 94.60 $\pm$ 0.01 & \textbf{54.63 $\pm$ 2.00} & 66.67 $\pm$ 1.84 \\
\bf \CON & \textbf{95.06 $\pm$ 0.01} & 94.93 $\pm$ 0.03 & 94.82 $\pm$ 0.03 & 94.94 $\pm$ 0.13 & \textbf{61.11 $\pm$ 1.20} & 66.67 $\pm$ 1.84 \\
\bf \SRL & 83.62 $\pm$ 0.06 & \textbf{83.78 $\pm$ 0.13} & 83.37 $\pm$ 0.01 & 83.37 $\pm$ 0.00 & \textbf{55.09 $\pm$ 0.40} & 66.67 $\pm$ 1.84
\end{tabular}
}
\caption{DeBERTa \cite{he2020deberta} pruning results.}
\label{tab:pruning-results-deberta}
\end{subtable}
\hfill
\caption{Results of single-task learning (STL), STL with static pruning (STL-SP) and multi-task learning on the 5 tasks with/without dynamic pruning (MTL/MTL-DP). The STL Performance column is equivalent to the main diagonal in Table~\ref{tab:mtl-results-more}.}
\label{tab:pruning-results-more}
\end{table*}

\end{document}